\documentclass{article}
\usepackage{arxiv}

\usepackage{amssymb, amsmath}
\usepackage[accsupp]{axessibility}  
\usepackage[ruled,vlined]{algorithm2e}
\usepackage{enumitem}

\usepackage[utf8]{inputenc} 
\usepackage[T1]{fontenc}    
\usepackage{url}            
\usepackage{booktabs}       
\usepackage{amsfonts}       
\usepackage{graphicx}       
\usepackage{nicefrac}       
\usepackage{microtype}      
\usepackage{xcolor}         
\usepackage{cancel}
\usepackage{wrapfig}
\usepackage{multirow}
\usepackage{tikz}
\usetikzlibrary{positioning}
\usepackage{geometry}  
\usepackage{orcidlink}

\usepackage{multirow}
\usepackage{array}
\newtheorem{definition}{Definition}

\newcommand{\taco}[1]{\texttt{TACO-Net}}

\begin{document}


\title{Topo-ADV: Generating Topology-Driven Imperceptible Adversarial Point Clouds}

%

\author{Gayathry Chandramana Krishnan Nampoothiry, Raghuram Venkatapuram, Anirban Ghosh, Ayan Dutta
\thanks{University of North Florida, Jacksonville, USA}}


\maketitle

\begin{abstract}
Deep neural networks for 3D point cloud understanding have achieved remarkable success in object classification and recognition, yet recent work shows that these models remain highly vulnerable to adversarial perturbations. Existing 3D attacks predominantly manipulate geometric properties such as point locations, curvature, or surface structure, implicitly assuming that preserving global shape fidelity preserves semantic content. In this work, we challenge this assumption and introduce the first \emph{topology-driven adversarial attack} for point cloud deep learning. Our key insight is that the homological structure of a 3D object constitutes a previously unexplored vulnerability surface. We propose Topo-ADV, an end-to-end differentiable framework that incorporates persistent homology as an explicit optimization objective, enabling gradient-based manipulation of topological features during adversarial example generation. By embedding persistence diagrams through differentiable topological representations, our method jointly optimizes (i) a topology divergence loss that alters persistence, (ii) a misclassification objective, and (iii) geometric imperceptibility constraints that preserve visual plausibility. Experiments demonstrate that subtle topology-driven perturbations consistently achieve up to 100\% attack success rates on benchmark datasets such as ModelNet40, ShapeNet Part, and ScanObjectNN using PointNet and DGCNN classifiers, while remaining geometrically indistinguishable from the original point clouds, beating state-of-the-art methods on various perceptibility metrics. 
\end{abstract}

\section{Introduction}

Deep neural networks for 3D point cloud understanding have rapidly advanced the state of the art in object classification, segmentation, and recognition; see~\cite{qi2017pointnet, qi2017pointnet++, tang2022contrastive, zhou2018voxelnet, wang2019dynamic}, for instance. Such architectures learn directly from unordered point sets and are now widely deployed across robotics, autonomous driving, AR/VR, and digital manufacturing. Despite this progress, recent work~\cite{xiang2019generating,tsai2020robust, wen2020geometry, huang2022shape, liu2022boosting, zhang20233d, dong2022isometric, lou2024hide, pang2025towards} has revealed that 3D deep learning models remain highly vulnerable to adversarial perturbations. Carefully crafted changes to point coordinates can significantly degrade model performance while remaining visually imperceptible to human observers. Most existing adversarial attacks on point clouds focus on \emph{geometric manipulation}. Prior work perturbs point locations, surface normals, curvature, or sampling density while enforcing constraints such as bounded displacement or surface preservation. These approaches implicitly assume that preserving global geometric appearance preserves the semantic identity of the object. However, this assumption overlooks a fundamental property of 3D shapes: their \emph{geometric topology}. 

\begin{figure}
    \centering
    \includegraphics[scale=0.1]{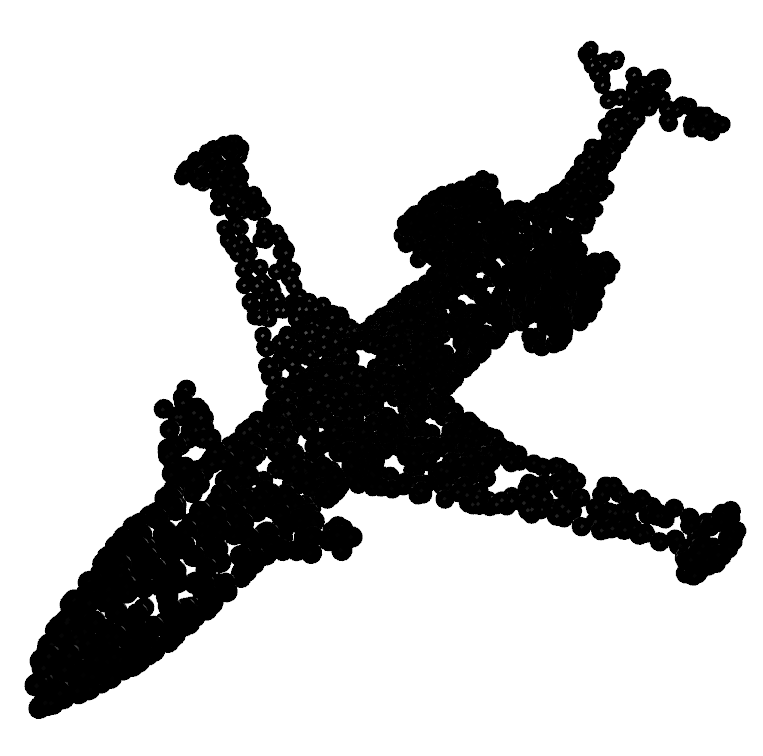}\hspace{-5pt}
        \includegraphics[scale=0.21]{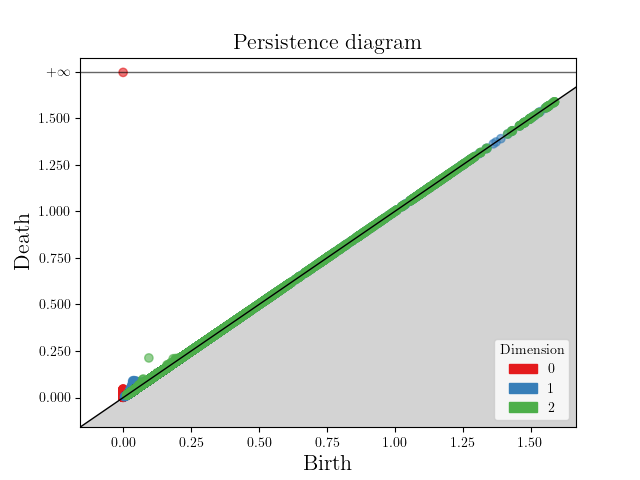}\hspace{-5pt}
            \includegraphics[scale=0.09]{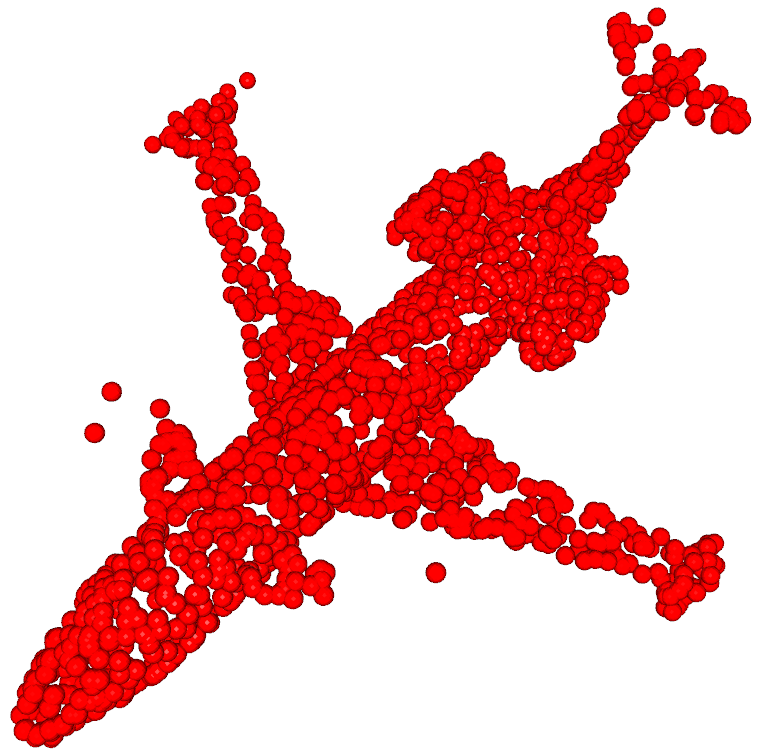}\hspace{-5pt}
                \includegraphics[scale=0.21]
                {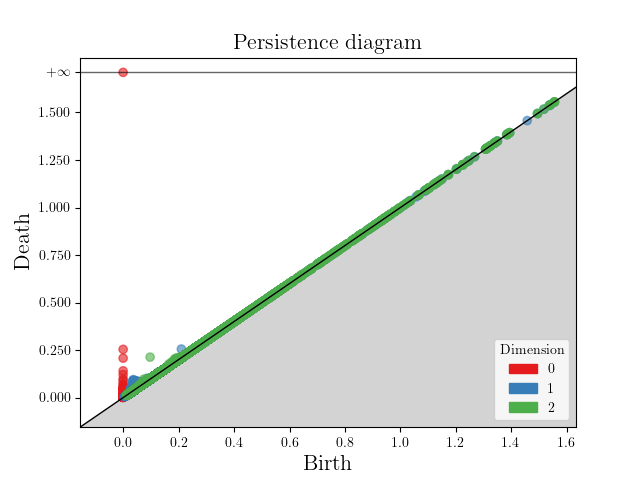}\hspace{-12pt}
    \caption{An \texttt{airplane} point cloud from ModelNet40 and its adversarial counterpart generated by Topo-ADV are shown alongside their corresponding persistence diagrams. The adversarial point cloud causes PointNet to misclassify the airplane as a \texttt{piano}. Note that the persistence diagram changed after we perturbed the original point cloud.}
    \label{fig:example}
\end{figure}

Geometric topology~\cite{chazal2021introduction, dey2022computational}, describes the intrinsic structural properties of a shape, including connected components, loops, tunnels, and cavities. These properties are captured by \textit{persistent homology}, a powerful tool from topological data analysis (TDA) that summarizes the birth and death of topological features across multiple spatial scales. 
This paper introduces a new threat model for 3D deep learning: {\textit{topology-driven adversarial attacks}}. Our central hypothesis is that neural networks trained on point clouds implicitly rely on topological cues, and that deliberately manipulating the homological structure of an object can induce misclassification even when geometric distortions remain imperceptible. This perspective shifts the adversarial paradigm from geometry-only perturbations to \emph{topology-aware optimization}.
Our topology-aware attack, called Topo-ADV, jointly optimizes three objectives: (i) a topology divergence loss that intentionally modifies persistent homology, (ii) a misclassification loss that reduces classifier confidence, and (iii) geometric regularization that preserves visual and structural plausibility. The resulting adversarial point clouds remain close to the original shapes yet exhibit altered topological signatures that significantly degrade classifier performance.

\noindent
\textit{Contributions.} This paper makes the following contributions:
\begin{itemize}
    \item We introduce Topo-ADV, the first \textit{topology-driven adversarial attack} for 3D point cloud deep learning, establishing topology manipulation as a new threat model beyond geometry-based perturbations. Our \textit{joint optimization framework}  simultaneously enforces misclassification, topology manipulation, and geometric imperceptibility constraints.
    
    
    
  \item We evaluate Topo-ADV on popular point cloud datasets, including ModelNet40~\cite{wu20153d}, ShapeNet Part~\cite{chang2015shapenet}, and ScanObjectNN (PB-T50-RS,  its hardest variant)~\cite{uy2019revisiting}, using PointNet~\cite{qi2017pointnet} and DGCNN classifiers~\cite{wang2019dynamic}. Experiments demonstrate that topology-driven perturbations achieve up to 100\% attack success rates while maintaining competitive imperceptibility scores. 

\end{itemize}

\noindent
\textbf{Related Work.} 
Deep neural networks operating directly on point clouds have become the dominant paradigm for 3D shape understanding. PointNet~\cite{qi2017pointnet} introduced a permutation-invariant architecture that learns global features directly from unordered point sets, while PointNet++~\cite{qi2017pointnet++} extended this framework with hierarchical local feature learning. Subsequent works have explored graph-based and convolutional approaches to capture richer geometric relationships, including DGCNN~\cite{wang2019dynamic}, KPConv~\cite{thomas2019kpconv}, and transformer-based models~\cite{guo2021pct,zhao2021point}. Despite their success across classification and segmentation tasks, recent studies have shown that these models remain highly vulnerable to adversarial perturbations.


Adversarial robustness has been extensively studied in 2D vision, and similar vulnerabilities have been identified in 3D deep learning. Early works demonstrated that small perturbations to point coordinates can significantly degrade model performance~\cite{liu2019extending}. Subsequent research explored various geometric attack strategies, including point shifting, point adding, and point dropping~\cite{xiang2019generating,zheng2019pointcloud, zhou2019dup}. Other works introduced surface-constrained and curvature-aware perturbations to maintain geometric plausibility~\cite{hamdi2020advpc}. More recent approaches investigate physically realizable attacks and robustness under sensor noise and transformations~\cite{sun2020adversarial}.  Despite these advances, existing 3D adversarial attacks remain fundamentally \emph{geometry-driven}. They primarily manipulate Euclidean coordinates or local surface properties while implicitly preserving global structural properties of the object. In contrast, our work explores adversarial manipulation of \emph{topological structure}, revealing a previously unexplored vulnerability surface in 3D deep learning.


Topological Data Analysis (TDA), particularly persistent homology, provides a principled framework for capturing multi-scale structural properties of data~\cite{edelsbrunner2002topological, chazal2021introduction}. Persistent homology has been successfully applied to various machine learning tasks, including shape analysis, representation learning, and robustness analysis~\cite{hofer2017deep, carriere2017sliced, adams2017persistence}. To integrate topology into deep learning pipelines, several differentiable persistence representations have been proposed, such as persistence images~\cite{adams2017persistence}, persistence landscapes~\cite{bubenik2015statistical}, and learnable embeddings including PersLay~\cite{carriere2020perslay}. These approaches enable gradient-based optimization using topological descriptors and have been applied to tasks including classification, segmentation, and representation learning.

Recent work has also explored topology-aware regularization and robustness in neural networks, demonstrating that persistent homology can improve generalization and stability~\cite{hofer2019connectivity, chen2019topological}. However, these methods primarily use topology as a \emph{regularizer} or analysis tool rather than an optimization objective for adversarial manipulation.


A small but growing body of work investigates the interaction between topology and adversarial robustness. Prior studies have analyzed how adversarial perturbations affect topological signatures of data~\cite{hu2019topology}, and others have proposed topology-based regularization to improve robustness~\cite{hofer2019connectivity}. However, to the best of our knowledge, no existing work directly optimizes persistent homology as an adversarial objective for 3D point clouds. 


\section{Preliminaries}
\label{sec:prelims}


Let $P := \{p_i\}_{i=1}^N \subset \mathbb{R}^3$ denote a clean point cloud sampled from a 3D object, where $N$ is the number of points. Given a pretrained classifier $f(\cdot)$ and ground-truth label $y$, our goal is to generate an adversarial point cloud
    $P_{\text{adv}} = P + \delta$,
where $\delta = \{\delta_i\}_{i=1}^N$ denotes per-point perturbations constrained by a bounded budget: $\|\delta_i\|_2 \le \epsilon,  \forall i$, such that $f(P) \neq f(P_\text{adv})$. 

In what follows, we give a brief overview of the concepts used from topological data analysis in Topo-ADV. 
For a comprehensive overview, refer to~\cite{dey2022computational,edelsbrunner2002topological,edelsbrunner2010computational,rabadan2019topological,chazal2021introduction}. 

A topological space is a pair $(X,\mathcal{U})$, where $X$ is a set and $\mathcal{U}$ is a collection of subsets of $X$, referred to as \textit{open sets}. The open sets satisfy the following conditions: (a) both the empty subset $\empty$ and $X$ are elements of $\mathcal{U}$, (b) any union of elements of $\mathcal{U}$ is an element of $\mathcal{U}$, and (c) the intersection of a finite collection of elements of $\mathcal{U}$ is an element of $\mathcal{U}$.

A \textit{cell complex} provides a combinatorial representation of a topological space by decomposing it into cells (points, edges, faces, etc.) together with boundary operators describing how each $k$-cell is attached to the $(k-1)$-skeleton. These boundary maps generate chain complexes whose homology groups capture the topological invariants used in persistent homology.

For a cell complex $X_\alpha$, the $k$-th \emph{chain group}
$C_k(X_\alpha)$ is the vector space generated by all $k$-cells of $X_\alpha$.
The \emph{boundary operator} $\partial_k : C_k(X_\alpha) \to C_{k-1}(X_\alpha)$ maps each $k$-cell to the formal sum of its boundary $(k-1)$-cells and
extends linearly. The \emph{cycle group} is
$Z_k(X_\alpha) = \ker(\partial_k)$, and the \emph{boundary group} is
$B_k(X_\alpha) = \mathrm{im}(\partial_{k+1})$.
The $k$-th \emph{homology group} of $X_\alpha$ is the quotient $H_k(X_\alpha) = Z_k(X_\alpha)/B_k(X_\alpha)$,
whose elements represent $k$-dimensional \textit{homological features} such as
connected components ($k=0$), loops ($k=1$), and voids ($k=2$).

\textit{Persistent homology}, a popular tool used in topological data analysis,  studies the evolution of homological features across a
filtration of cell complexes
$X_0 \subseteq X_1 \subseteq \cdots \subseteq X_n = X$.
In this work, we consider filtrations arising from sublevel sets of a function
$g : X \to \mathbf{R}$, defined by $
X_\alpha = g^{-1}((-\infty,\alpha])$.
Each $X_\alpha$ is required to be a cell complex, meaning that whenever a cell
is included in $X_\alpha$, all of its boundary cells are also included.  As the
parameter $\alpha$ increases, new homology classes may be created or destroyed,
and persistent homology records the scales at which such events occur.
Given a filtration $\{X_i\}$, a class $[c]\in H_k(X_i)$ is \emph{born} at
index $i$ if it does not lie in the image of
$H_k(X_{i-1}) \to H_k(X_i)$. It \emph{dies entering} $j>i$ if its
image in $H_k(X_{j-1})$ is nonzero but becomes zero in $H_k(X_j)$.
Its \emph{persistence interval} is $[i,j)$. Next, we formally define the \textit{persistence diagram}, a visual summary (shown using scatter plots) of persistent homology that shows which topological features (such as connected components, loops, or voids) appear and how long they persist across scales. See Fig.~\ref{fig:example} for an example.

\begin{definition}[Persistence diagram]
For each dimension $k$, the \emph{persistence diagram} $D_k$ is the multiset
\[
D_k = \{(b,d) \mid \text{a $k$-dimensional homology class is born at $b$
and dies at $d$} \},
\]
augmented with diagonal points $(t,t)$ for stability.
Each point $(b,d)$ corresponds to a $k$-dimensional feature with
\emph{birth time} $b$, \emph{death time} $d$, and \emph{lifetime} $d-b$.
\end{definition}

In Topo-ADV, the alpha complex is deployed as the filtration for persistent homology. We now present its definition. 

\begin{definition}[Alpha complex] Let $X=\{x_1,\dots,x_n\}\subset \mathbb{R}^d$ be a finite point set and let $\textsf{DT}$ denote its Delaunay triangulation~\cite{de2008computational}.
For each simplex $\sigma\in\textsf{DT}$, let $B_\sigma$ be the smallest closed ball whose boundary passes through all vertices of $\sigma$ (its circumscribed ball), and let $r_\sigma$ be its radius.
For $\alpha\ge 0$, the \emph{alpha complex} $A_\alpha$ is the subcomplex of $\textsf{DT}$ defined by,
\[
A_\alpha \;=\; \bigl\{\, \sigma\in\textsf{DT}\ \big|\ r_\sigma \le \alpha \ \text{ and }\ B_\sigma\cap X = \mathrm{vert}(\sigma)\,\bigr\},
\]
i.e., those Delaunay simplices whose circumscribed balls have radius at most $\alpha$ and are empty of sample points except for their own vertices.
As $\alpha$ increases, $\{A_\alpha\}_{\alpha\ge 0}$ forms a nested filtration of simplicial complexes, which can be used to compute persistent homology.
\end{definition}

\section{Proposed Methodology for Topo-ADV}

Adversarial perturbations on point clouds often exploit local geometric instabilities, such as shifts in point positions, changes in sampling density, or distortions of surface curvature that subtly alter the classifier's decision boundary. However, point clouds also encode \emph{global} structural information not captured by purely geometric metrics. This global structure is naturally described using \emph{persistent homology} (refer to Section~\ref{sec:prelims}).

Conventional adversarial attacks on 3D point clouds rely exclusively on geometric constraints (e.g., Chamfer distance, smoothness priors)~\cite{lou2024hide,wen2020geometry,xiang2019generating}, which regulate local perturbations but do not control how the global organization of the shape is modified. Our approach, {Topo-ADV}, explicitly integrates persistent homology into the attack objective, allowing us to steer perturbations toward or away from specific topological configurations. By embedding persistence diagrams in a differentiable representation, we turn global topological signals into gradients that interact directly with the classifier. Thus, unlike prior geometry-driven attacks, Topo-ADV treats topology as an \emph{optimization target}.
The attack optimizes three objectives simultaneously: (a) misclassification of the input, (b) manipulation of persistent homology through differentiable embeddings, and (c) preservation of geometric fidelity and perceptual realism.

In the following, we present the components of the Topo-ADV pipeline. We start with the components that leverage topology.

\medspace

\noindent
\textbf{Differentiable Persistent Homology Module.} To capture 3D global structure, we compute persistent homology up to dimension $2$ using the \emph{Alpha complex filtration}  (refer to Section~\ref{sec:prelims}). Consequently, we obtain:
\vspace*{-5pt}
\[
\mathrm{Dgm}(P) := \mathrm{Dgm}_0(P) \cup \mathrm{Dgm}_1(P) \cup \mathrm{Dgm}_2(P),
\]
where $\mathrm{Dgm}_i$ contains the birth-death pairs in dimension $i$. 
Now, persistence diagrams are multisets and therefore not directly compatible with gradient-based optimization. To integrate them into the attack pipeline, we apply a differentiable persistence embedding:
\vspace*{-5pt}
\[
\phi(P) = \Phi(\mathrm{Dgm}(P)) \in \mathbf{R}^d,
\]
implemented by a learnable module. 
  Each birth-death pair $(b_i, d_i)$ is transformed into a birth-persistence pair $(b_i, p_i)$ with $p_i = d_i - b_i$.  The quantity $p_i$ is also known as the \textit{lifetime} in the literature. These pairs are passed through an MLP $\psi$, producing per-feature vectors $\psi(b_i, p_i)$.  
To emphasize stable topological features, each vector is weighted by its persistence $p_i$, and all contributions are summed as: \vspace*{-5pt}

$$\phi(P) = \sum_i p_i\,\psi(b_i,p_i).$$
This approach is permutation-invariant, smooth, and fully differentiable, enabling gradient flow from the topological signature to the point cloud.


Let $\phi_{\text{clean}} = \phi(P)$ denote the topological embedding of the clean cloud and 
$\phi_{\text{adv}} = \phi(P_{\text{adv}})$ that of the adversarial cloud.
To enforce topological change, we penalize embedding similarity:
\begin{equation*}
\mathcal{L}_{\text{div}} = \|\phi_{\text{adv}} - \phi_{\text{clean}}\|_2^2.
\end{equation*}
It encourages the adversarial example to deviate from the clean topological signature.

Beyond measuring the overall deviation between clean and adversarial persistence
embeddings, as described above, Topo-ADV incorporates a mechanism to directly influence how the
topology evolves during optimization. Specifically, it modulates the lifetimes
of persistent homology classes across dimensions $k \in \{0,1,2\}$. Let
$\ell_k$ denote the vector of lifetimes extracted from the $k$-dimensional
persistence diagram, restricted to the $K$ most persistent features
to emphasize structurally significant elements. Dimension-specific weights
$w_k$ allow the attack to prioritize modifications in particular homological
degrees. We define the directional topology manipulation term as:
\vspace{-2pt}
\[
\mathcal{L}_{\mathrm{dir}}
    = \sum_{k=0}^{2} w_k \, s_k,
\qquad
s_k =
\begin{cases}
\sum \ell_k, & \text{destruction mode},\\[4pt]
-\sum \ell_k, & \text{creation mode}.
\end{cases}
\]
In destruction mode, the loss encourages the shortening of dominant lifetimes,
thereby destabilizing prominent topological features. In creation mode, the sign
inversion promotes the enlargement of these lifetimes, encouraging the emergence
of new or more persistent structures. This provides a principled means of
steering the adversarial perturbation toward systematically reducing or
enhancing the topological complexity of the shape.

To balance these opposing directions and avoid overly aggressive distortions,
Topo-ADV adopts an automatic two-phase schedule: the optimization begins in a
destruction phase for a short patience interval, weakening existing topological
features, and subsequently switches to a creation phase to promote the formation
of alternative, stable structures. This directional modulation ensures controlled
yet effective evolution of the shape's topology while maintaining geometric
plausibility.

The complete topological loss is:
\vspace*{-5pt}
\[
\mathcal{L}_{\text{PH}} = \alpha\,\mathcal{L}_{\text{div}} + \beta\,\mathcal{L}_{\text{dir}}.
\]
The parameters $\alpha$ and $\beta$ balance the topological objective, with $\alpha$ scaling the embedding‑based divergence between clean and adversarial persistence signatures and $\beta$ weighting the direct manipulation of feature lifetimes in the persistence diagrams.

\medspace

\noindent
\textbf{Misclassification Objective.} To promote adversarial misclassification, we adopt the confidence-based
Carlini--Wagner margin loss~\cite{carlini2017towards}.  
Let $f(\cdot)$ denote the classifier's output logits, and let
$f_k(P_{\mathrm{adv}})$ denote the logit assigned to class $k$ when evaluating the
adversarial point cloud $P_{\mathrm{adv}}$.  
If $y$ is the true class label, the loss is defined as:
\vspace{-4pt}
\begin{equation*}
\mathcal{L}_{\mathrm{cls}}
    = 
    \max\!\left(
        f_y(P_{\mathrm{adv}}) 
        - 
        \max_{j \neq y} f_j(P_{\mathrm{adv}}),
        \, -\kappa
    \right),
\end{equation*}
where the parameter $\kappa \ge 0$ specifies the desired confidence margin.
A larger $\kappa$ enforces a stronger separation between the true class logit
and the highest competing logit, making the attack more confident.

\medspace

\noindent
 \textbf{Geometric Imperceptibility Constraints.} To ensure that adversarial perturbations remain visually plausible and do not
introduce artifacts inconsistent with the underlying surface geometry, we impose
a set of geometric regularizers that quantify how strongly the perturbed point
cloud deviates from clean local structure.  These terms are computed using
clean pre-processing statistics, such as $k$NN neighborhoods, PCA-estimated
normals, and a curvature proxy. 

\textit{Chamfer distance.} To limit global point-wise displacement, we penalize the bidirectional
nearest-neighbor discrepancy between the clean cloud $P$ and the adversarial
cloud $P_{\mathrm{adv}}$:
\vspace{-2pt}
\[
\mathcal{L}_{\mathrm{CD}}
\,=\,
\frac{1}{|P|} \left( \sum_{x \in P} 
    \min_{y \in P_{\mathrm{adv}}} \|x-y\|_2
\;+\;
 \sum_{y \in P_{\mathrm{adv}}} 
    \min_{x \in P} \|y-x\|_2 \right).
\]
\textit{Normal consistency.} Let $n_{\mathrm{clean}}(x)$ and $n_{\mathrm{adv}}(x)$ denote the PCA normals at
corresponding points in $P$ and $P_{\mathrm{adv}}$.  To preserve the local
tangent-plane orientation, we penalize their deviation:
\vspace{-2pt}
\[
\mathcal{L}_{\mathrm{norm}}
\,=\,
\frac{1}{|P|}
\sum_{x \in P}
\big\| n_{\mathrm{adv}}(x) - n_{\mathrm{clean}}(x) \big\|_2^2 .
\]

\textit{Curvature consistency.} Let $\kappa_{\mathrm{clean}}(x)$ and $\kappa_{\mathrm{adv}}(x)$ denote a scalar
curvature proxy computed from the $k$-nearest neighbors of $x$ in $P$ and
$P_{\mathrm{adv}}$, respectively.  To maintain the second-order differential
structure of the surface, we enforce:
\vspace{-2pt}
\[
\mathcal{L}_{\mathrm{curv}}
\,=\,
\frac{1}{|P|}
\sum_{x \in P}
\big\|
\kappa_{\mathrm{adv}}(x) - \kappa_{\mathrm{clean}}(x)
\big\|_2^2 .
\]

\textit{Laplacian smoothness.}
To suppress high-frequency noise in the perturbation field and encourage
coherent local motion, we apply a graph-Laplacian penalty over the clean
$k$NN graph $\mathcal{N}(x)$:
\vspace{-2pt}
\[
\mathcal{L}_{\mathrm{lap}}
\,=\,
\frac{1}{|P|}
\sum_{x \in P}
\left\|
\delta(x)
-
\frac{1}{|\mathcal{N}(x)|}
\sum_{x' \in \mathcal{N}(x)} \delta(x')
\right\|_2^2,
\]
where $\delta(x)$ denotes the perturbation applied at point $x$.

\paragraph{Combined geometric regularizer.}
The full geometric fidelity term is then given by:
\vspace{-2pt}
\[
\mathcal{L}_{\mathrm{geom}}
=
\mathcal{L}_{\mathrm{CD}}
+
\mathcal{L}_{\mathrm{norm}}
+
\mathcal{L}_{\mathrm{curv}}
+
\mathcal{L}_{\mathrm{lap}},
\]
which ensures that the adversarial point cloud remains consistent with both the
global shape and the local differential structure of the original surface, while
still permitting meaningful topological modification.

\medspace

\noindent
\textbf{Full Objective.} The final attack objective combines classification, topology, and geometry:
\vspace{-2pt}
\begin{equation*}
\mathcal{L} =
\lambda_1 \mathcal{L}_{\text{cls}} +
\lambda_2 \mathcal{L}_{\text{PH}} +
\lambda_3 \mathcal{L}_{\text{geom}},
\end{equation*}
where $\lambda_1,\lambda_2,\lambda_3$ control task trade-offs.

\medspace


\noindent \textbf{Optimization Procedure.}
Topo-ADV generates adversarial examples using projected gradient descent (PGD),
where at iteration $t$ the perturbation is updated as
\[
\delta^{(t+1)}
\,=\, \delta^{(t)} - \eta_t \nabla_{\!\delta}\mathcal{L},
\]
with $\nabla_{\!\delta}\mathcal{L}$ denoting the gradient of the full objective
$\mathcal{L}$ with respect to the perturbation $\delta$, and $\eta_t$ a step
size following a prescribed decay schedule. Following each gradient update, two projection operators are applied to maintain
the physical plausibility of the perturbed surface. The first is a
\emph{tangent-plane projection}, which ensures that perturbations remain
consistent with the underlying local surface geometry. Given the clean
per-point normals $n_{\mathrm{clean}}$, the update $\delta$ is decomposed into
tangential and normal components. Since displacements along the normal direction
tend to lift points off the surface and introduce visually unrealistic
artifacts, the normal component is removed:
\[
\delta \;\leftarrow\; \delta - \langle \delta, n_{\mathrm{clean}} \rangle \, 
n_{\mathrm{clean}}.
\]
This operation confines all perturbations to the estimated tangent spaces of the
clean shape, preserving surface smoothness and embedding structure. The second projection enforces the perturbation budget by constraining each
per-point displacement to lie inside an $\ell_2$ ball of radius $\epsilon$:
\[
\|\delta_i\|_2 \leq \epsilon,\qquad \forall\, i.
\]
This guarantees that no point undergoes excessive movement, thereby bounding the
visual footprint of the attack. The PGD loop terminates early if a stable misclassification is detected, and the
final adversarial example is selected as the first successful result across the
$R$ restarts (or the last iterate if none succeed). This optimization procedure
jointly incorporates classification, geometric, and topological objectives,
enabling Topo-ADV to produce adversarial point clouds that are both effective
and geometrically credible. A high-level pseudocode is provided in Algo. \ref{alg:topoadv-restarts-clean}.

\begin{algorithm}[t]
\caption{A high-level pseudo-code of Topo-ADV}
\label{alg:topoadv-restarts-clean}
\DontPrintSemicolon



\For{$r\gets 1$ \KwTo $R$}{

  \lIf{$r=1$}{
    $\delta_0 \leftarrow 0$;
  }
  \Else{
    Initialize $\delta_0$ by sampling a small random perturbation at each point, projecting it onto the local tangent plane using the clean normals, and normalizing each vector so its $\ell_2$ norm does not exceed $\epsilon$;
  }

  $t\leftarrow 0$;

  \While{$t < T$}{

    $P^{\mathrm{adv}} \leftarrow P + \delta_t$;


    $\delta' \leftarrow \delta_t - \eta_t\, \nabla_{\delta}\bigl(\lambda_1\mathcal{L}_{\mathrm{cls}}
    + \lambda_2\mathcal{L}_{\mathrm{PH}} + \lambda_3\mathcal{L}_{\mathrm{geom}}\bigr)$;

    Remove from $\delta'$ the component pointing along each clean normal:
    $\delta' \leftarrow \delta' - \langle \delta', n_{\mathrm{clean}}\rangle n_{\mathrm{clean}}$;

    For each point $i$, scale $\delta'^{(i)}$ if needed so that $\|\delta'^{(i)}\|_2 \le \epsilon$;

    $\delta_{t+1} \leftarrow \delta'$;

    \If{$\arg\max f(P+\delta_{t+1}) \neq y$ and the label flip is stable}{
      \Return{$P^{\mathrm{adv}} = P + \delta_{t+1}$};
    }

    $t \leftarrow t+1$;
  }
}
\textbf{return} {$P^{\mathrm{adv}} = P + \delta_T \text{ from the final restart}$;}
\end{algorithm}
\section{Experiments and Results}

\noindent
\textbf{Datasets. }We conduct experiments on three widely used point cloud benchmark datasets: {ModelNet40}~\cite{wu20153d}, {ShapeNet Part}~\cite{chang2015shapenet}, and {ScanObjectNN}~\cite{uy2019revisiting}. The ModelNet40 dataset consists of 12{,}311 synthetic CAD models belonging to 40 object categories, split into 9{,}843 samples for training and 2{,}468 for testing. The ShapeNet Part dataset contains 16{,}881 pre-aligned shapes covering 16 object categories, with 12{,}137 shapes for training and 2{,}874 for testing; although originally introduced for part segmentation, it is widely employed to assess robustness against geometric perturbations. To evaluate robustness under realistic conditions, we further include the ScanObjectNN dataset (PB-T50-RS, the hardest variant), which comprises approximately 15{,}000 objects captured from real-world scans across 15 categories. Due to background clutter, sensor noise, and occlusions, ScanObjectNN poses substantial challenges and serves as a rigorous benchmark for adversarial robustness evaluation.

\medskip
\noindent
\textbf{Victim Models. }For our classification backbone, we adopt two widely used and representative point cloud models: the MLP-based {PointNet}~\cite{qi2017pointnet} and the graph-based {DGCNN}~\cite{wang2019dynamic}. Prior to conducting adversarial evaluations, we train both models on clean training data following the original implementations to ensure fair and consistent comparisons.

\medskip
\noindent
\textbf{Evaluation Metrics. }To evaluate the effectiveness of adversarial attacks, we measure the \emph{attack success rate} (ASR), which denotes the percentage of adversarial examples that successfully fool the classifier. In addition, we assess the imperceptibility of generated perturbations using three distance-based metrics. Following prior work, we employ the curvature standard deviation distance (CSD) as a shape-related metric, along with a geometric distance measure: the uniform~\cite{li2019pu} distance. The CSD metric compares the consistency of curvature statistics between clean and adversarial point clouds, providing both a global similarity measure and an indication of whether the perturbations remain reasonable with respect to the underlying surface complexity. Formally, CSD is defined as:
\begin{equation}
    \mathrm{CSD} = \left\| \mathcal{C}_{\mathrm{std}}(P) - \mathcal{C}_{\mathrm{std}}(P_{\text{adv}}) \right\|_{2},
\end{equation}
where $\mathcal{C}_{\mathrm{std}}(P)$ and $\mathcal{C}_{\mathrm{std}}(P_{\text{adv}})$ denote the curvature standard deviations of the clean point cloud $P$ and the adversarial point cloud $P_{\text{adv}}$, respectively.

\medskip
\noindent
\textbf{Implementation Details.} The experiments were run on a Dell laptop equipped with an Intel i7 13650 HX 14-core CPU, 96 GB DDR5 RAM, and an 8 GB
NVIDIA RTX 4060 GPU. We set $R=3$, $T=300$, and $\epsilon=0.55$ (same as \cite{lou2024hide}). The values for $\lambda_1,\lambda_2,\lambda_3$ were set to $10, 0.001, 5$, respectively. The $\eta_t$ in the algorithm was set to $0.001$. The constants $w_1,w_2,w_k,K$ were set to $0.3,1,1,50$, respectively. The parameter $\kappa$ used in $\mathcal{L}_\text{cls}$ was set to $0.05$. For $\mathcal{L}_\text{PH}$, $\alpha=1, \beta=1$ was used. The $k$-value used for $k$NN neighborhoods was 16. The average time taken to generate an adversarial point cloud was around 8 seconds. The code will be released upon acceptance.

\begin{table*}[ht!]
\centering
\small
\setlength{\tabcolsep}{4pt}
\begin{tabular}{c|l|ccc}
\toprule
\textbf{Model} & \textbf{Method} 
& ASR(\%)$\uparrow$ & CSD$\downarrow$ & Uniform$\downarrow$ \\
\midrule
\multirow{11}{*}{PointNet}
& IFGSM($l_\infty$) & 99.01 & 2.6341 & 0.3134 \\
& IFGM($l_2$)~\cite{liu2019extending} & 99.68 & 1.6502 & 0.3173 \\
& 3D-ADV~\cite{xiang2019generating} & \textbf{100.00} & 0.9500 & 0.2965 \\
& KNN~\cite{tsai2020robust} & 99.68 & 2.3133 & 0.3695 \\
& GeoA$^3$~\cite{wen2020geometry} & \textbf{100.00} & 1.7112 & 0.2919 \\
& SI-ADV~\cite{huang2022shape} & 99.32 & 1.4601 & 0.3059 \\
& AOF~\cite{liu2022boosting} & \textbf{100.00} & 2.8213 & 0.3809 \\
& MeshAttack~\cite{zhang20233d} & 96.64 & 2.2161 & 0.3008 \\
& Eps-iso~\cite{dong2022isometric} & 97.17 & 2.2514 & 0.3053 \\
& HiT-ADV~\cite{lou2024hide} & \textbf{100.00} & \textbf{0.4709} & {0.2883} \\
& Topo-ADV (Ours)  & 99.86 & {1.2328} & \textbf{0.0121} \\
\midrule
\multirow{11}{*}{DGCNN}
& IFGSM($l_\infty$) & 98.96 & 2.7243 & 0.3112 \\
& IFGM($l_2$)~\cite{liu2019extending} & 98.96 & 2.0193 & 0.2983 \\
& 3D-ADV~\cite{xiang2019generating} & \textbf{100.00} & 1.0206 & 0.2919 \\
& KNN~\cite{tsai2020robust} & 96.15 & 3.2322 & 0.3987 \\
& GeoA$^3$~\cite{wen2020geometry} & 99.71 & 1.0286 & 0.2887 \\
& SI-ADV~\cite{huang2022shape} & 96.08 & 2.3211 & 0.5557 \\
& AOF~\cite{liu2022boosting} & 97.09 & 2.3148 & 0.3226 \\
& MeshAttack~\cite{zhang20233d} & \textbf{100.00} & 1.0411 & 0.3192 \\
& Eps-iso~\cite{dong2022isometric} & \textbf{100.00} & 1.0430 & 0.3001 \\
& HiT-ADV~\cite{lou2024hide} & \textbf{100.00} & \textbf{0.1946} & {0.2867} \\
& Topo-ADV (Ours) & \textbf{100.00} & 2.8549 & \textbf{0.0137} \\
\bottomrule
\end{tabular}
\caption{Comparison of adversarial attacks on \textbf{ModelNet40}. Best results per column are highlighted in bold.}
\label{tab:modelnet40_attack}
\end{table*}



\medskip

\noindent
\textbf{Performance Analysis.} First, we compare Topo-ADV against ten existing attacks (Tables~\ref{tab:modelnet40_attack} and~\ref{tab:shapenet_attack}) using ASR, CSD, and Uniform metrics~\cite{lou2024hide}. Topo-ADV consistently achieves $100\%$ ASR across multiple victim architectures on both ModelNet40 and ShapeNet Part, revealing a strong and consistent vulnerability of current point cloud classifiers under topology-driven perturbations.

\begin{table*}[ht!]
\centering
\small
\setlength{\tabcolsep}{4pt}
\begin{tabular}{c|l|ccc}
\toprule
\textbf{Model} & \textbf{Method} 
& ASR(\%)$\uparrow$ & CSD$\downarrow$ & Uniform$\downarrow$ \\
\midrule
\multirow{11}{*}{PointNet}
& IFGSM($l_\infty$) & 95.20 & 2.8831 & 0.3035 \\
& IFGM($l_2$)~\cite{liu2019extending} & 97.49 & 2.3401 & 0.2404 \\
& 3D-ADV~\cite{xiang2019generating} & \textbf{100.00} & 2.1603 & 0.2103 \\
& KNN~\cite{tsai2020robust} & 96.40 & 2.5067 & 0.2723 \\
& GeoA$^3$~\cite{wen2020geometry} & \textbf{100.00} & 3.7291 & 0.2658 \\
& SI-ADV~\cite{huang2022shape} & 96.38 & 2.8476 & 0.2945 \\
& AOF~\cite{liu2022boosting} & 99.85 & 3.3172 & 0.3601 \\
& MeshAttack~\cite{zhang20233d} & 97.24 & 3.0671 & 0.3444 \\
& Eps-iso~\cite{dong2022isometric} & 98.01 & 2.6093 & 0.2908 \\
& HiT-ADV~\cite{lou2024hide} & \textbf{100.00} & \textbf{0.9810} & 0.1874 \\
& Topo-ADV (Ours) & \textbf{100.00} & 2.3936 & \textbf{0.0156} \\
\midrule
\multirow{11}{*}{DGCNN}
& IFGSM($l_\infty$) & 98.46 & 3.3550 & 0.2094 \\
& IFGM($l_2$)~\cite{liu2019extending} & 99.51 & 2.6947 & 0.2080 \\
& 3D-ADV~\cite{xiang2019generating} & \textbf{100.00} & 2.3788 & 0.1886 \\
& KNN~\cite{tsai2020robust} & 98.09 & 3.4718 & 0.2372 \\
& GeoA$^3$~\cite{wen2020geometry} & \textbf{100.00} & 2.9907 & 0.1895 \\
& SI-ADV~\cite{huang2022shape} & 96.43 & 3.0193 & 0.2456 \\
& AOF~\cite{liu2022boosting} & 98.54 & 3.2227 & 0.2444 \\
& MeshAttack~\cite{zhang20233d} & 99.43 & 3.1836 & 0.2232 \\
& Eps-iso~\cite{dong2022isometric} & \textbf{100.00} & 3.0629 & 0.2171 \\
& HiT-ADV~\cite{lou2024hide} & \textbf{100.00} & \textbf{0.6629} & 0.1870 \\
& Topo-ADV (Ours) & \textbf{100.00} & 3.8679 & \textbf{0.0146} \\
\bottomrule
\end{tabular}
\caption{Comparison of adversarial attacks on \textbf{ShapeNet Part}. Best results per column are highlighted in bold.}
\label{tab:shapenet_attack}
\end{table*}

\begin{figure}
\centering
\begin{tabular}{c|c|c|c|c|c|c}

$P$ & \includegraphics[width=0.12\linewidth]{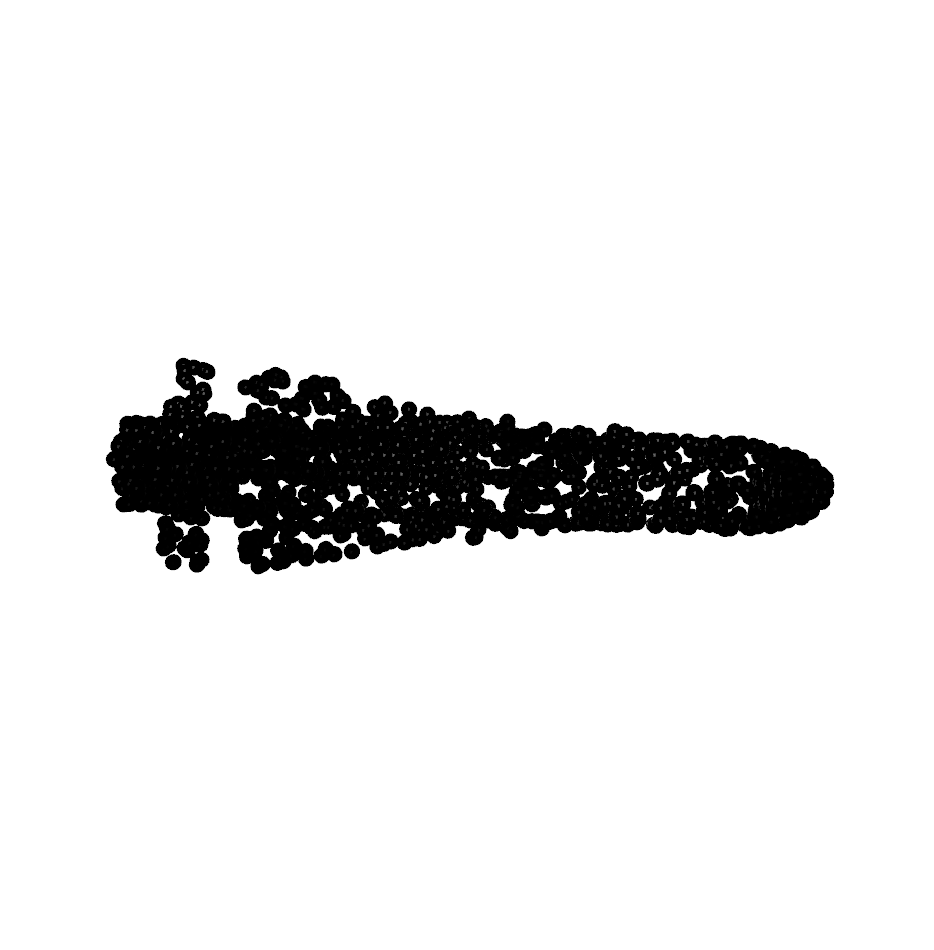} &
\includegraphics[width=0.12\linewidth]{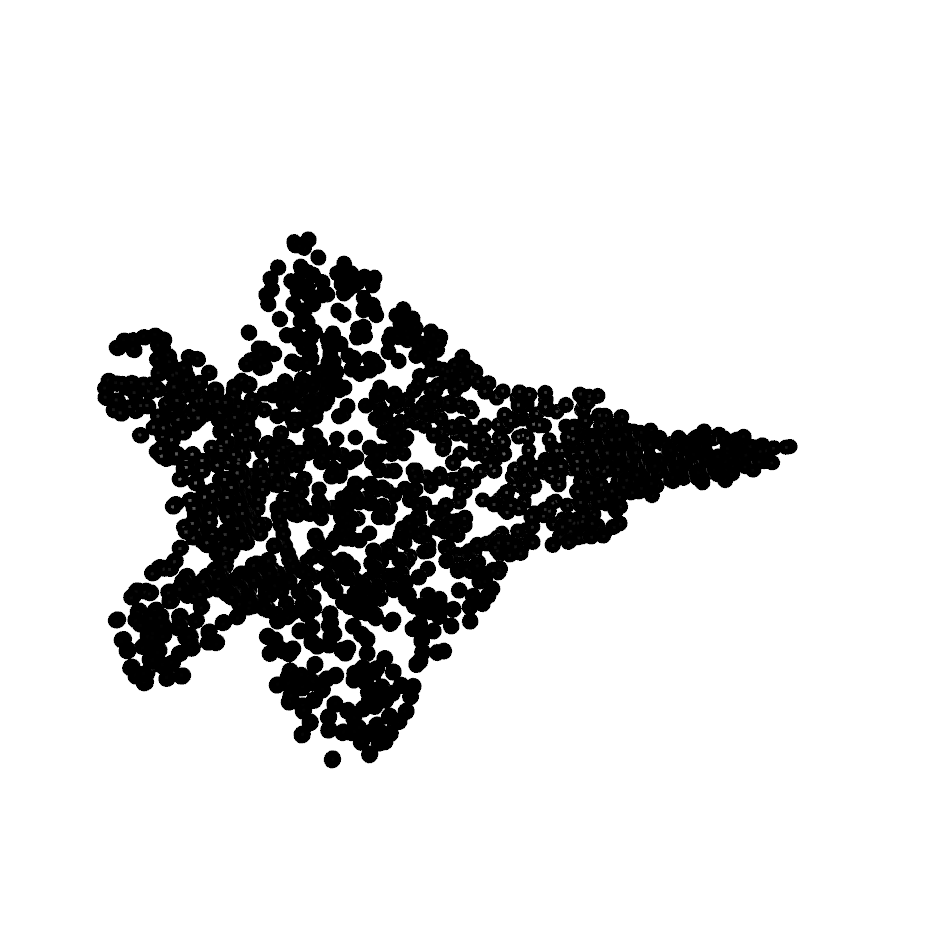} &
\includegraphics[width=0.12\linewidth]{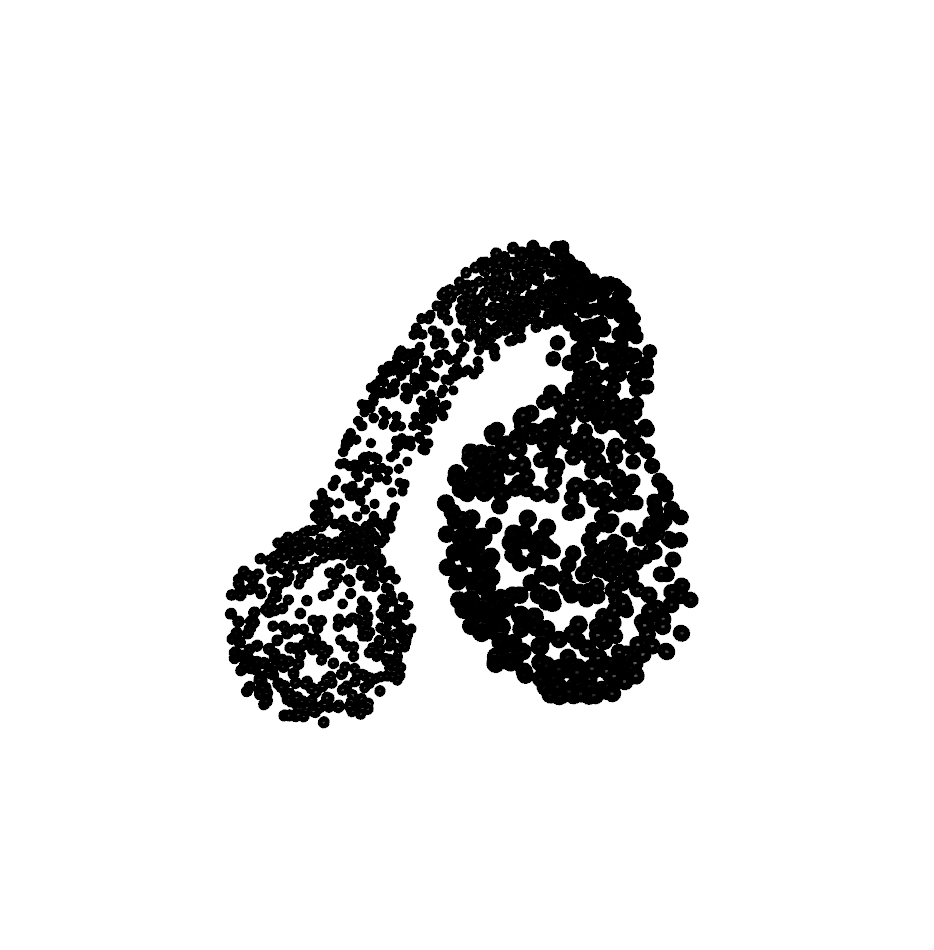} &
\includegraphics[width=0.12\linewidth]{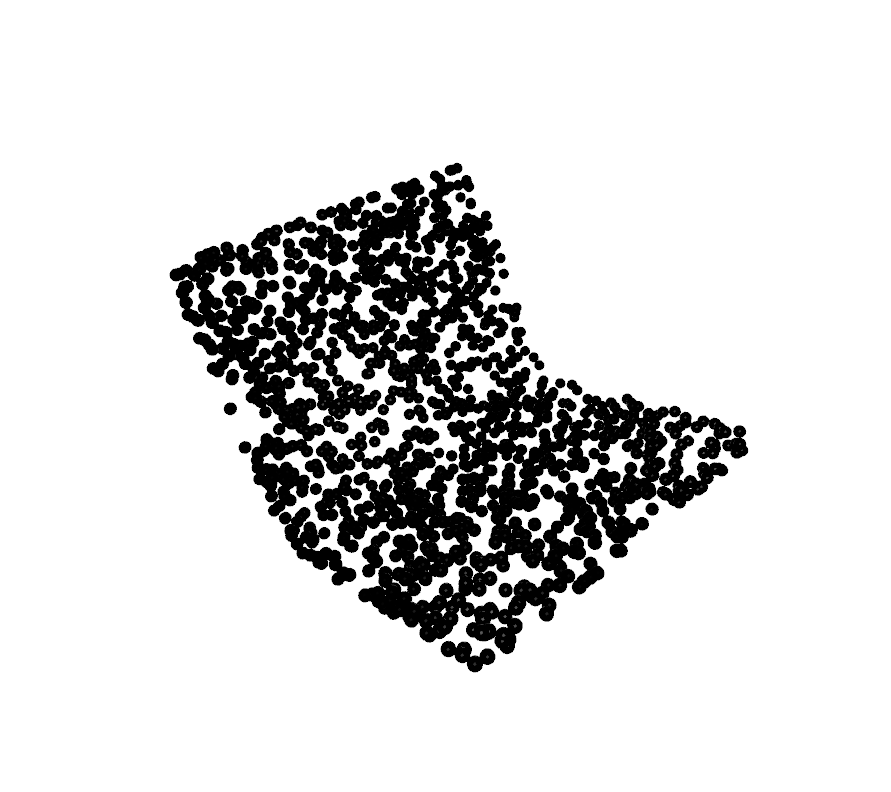} &
\includegraphics[width=0.12\linewidth]{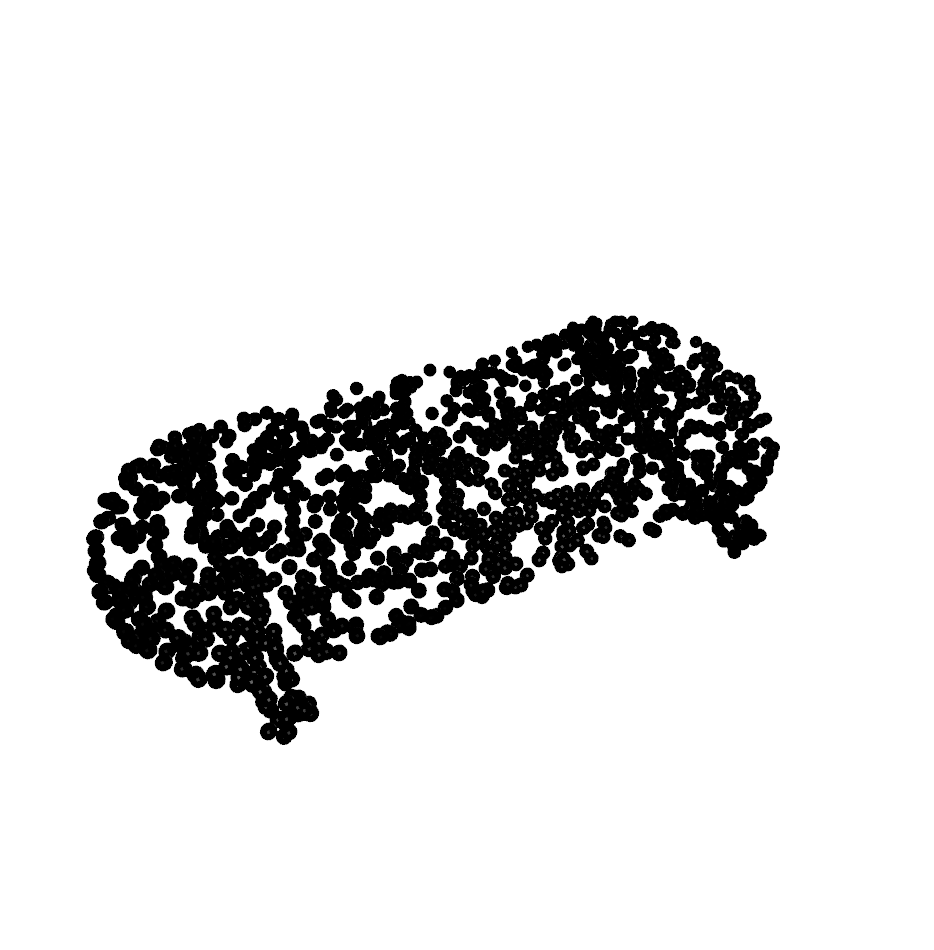} & 
\includegraphics[width=0.12\linewidth]{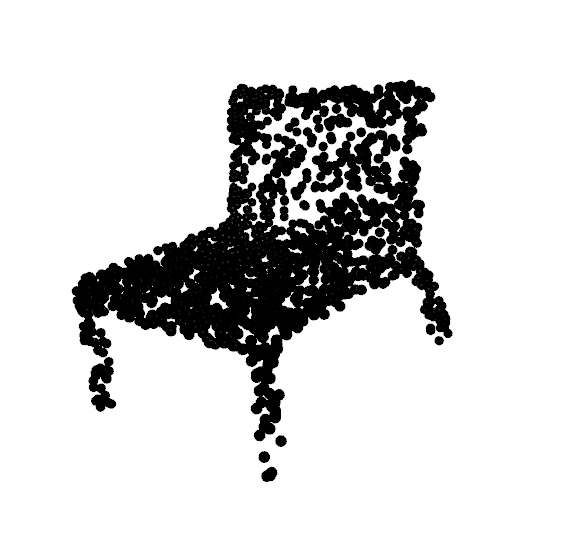}
\\

&  \textbf{\scriptsize {Rocket}} &
\textbf{\scriptsize Airplane} &
\textbf{\scriptsize Earphone} &
\textbf{\scriptsize Laptop} &
\textbf{\scriptsize Skateboard} &  \textbf{\scriptsize Chair} \\


\parbox{2cm}{\centering {$P_\text{adv}$} \\ \scriptsize{(PointNet)}} & \includegraphics[width=0.12\linewidth]{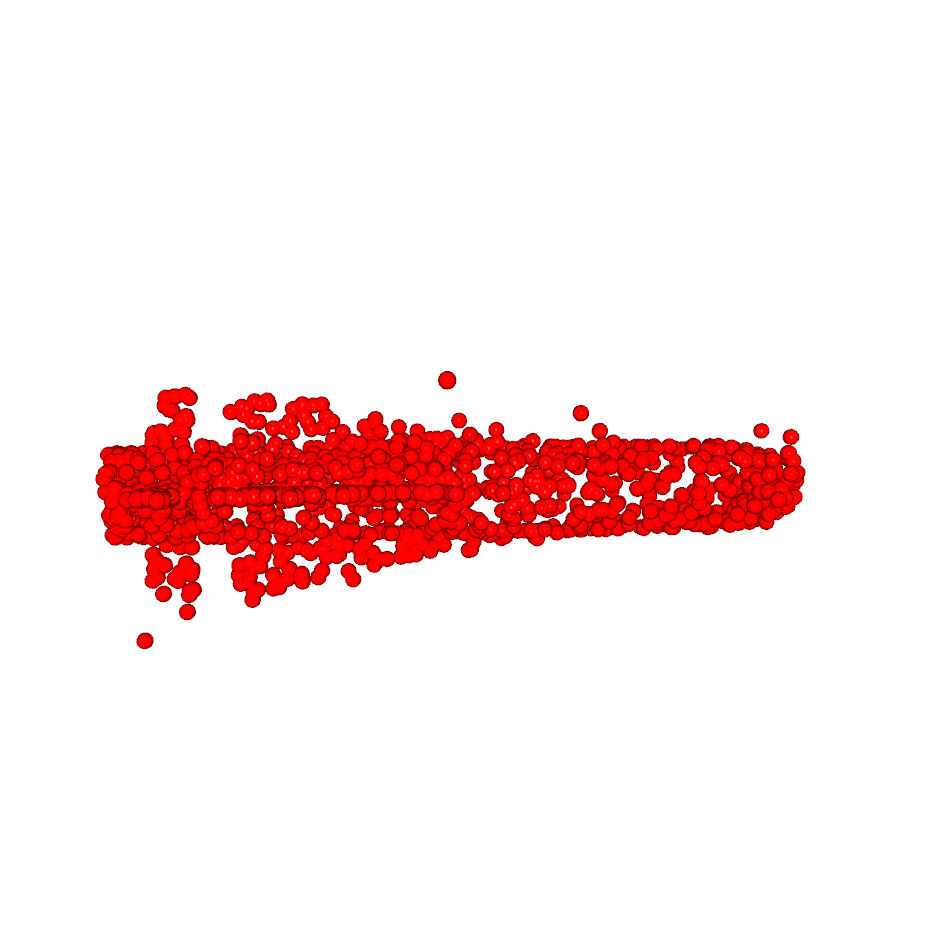} &
\includegraphics[width=0.12\linewidth]{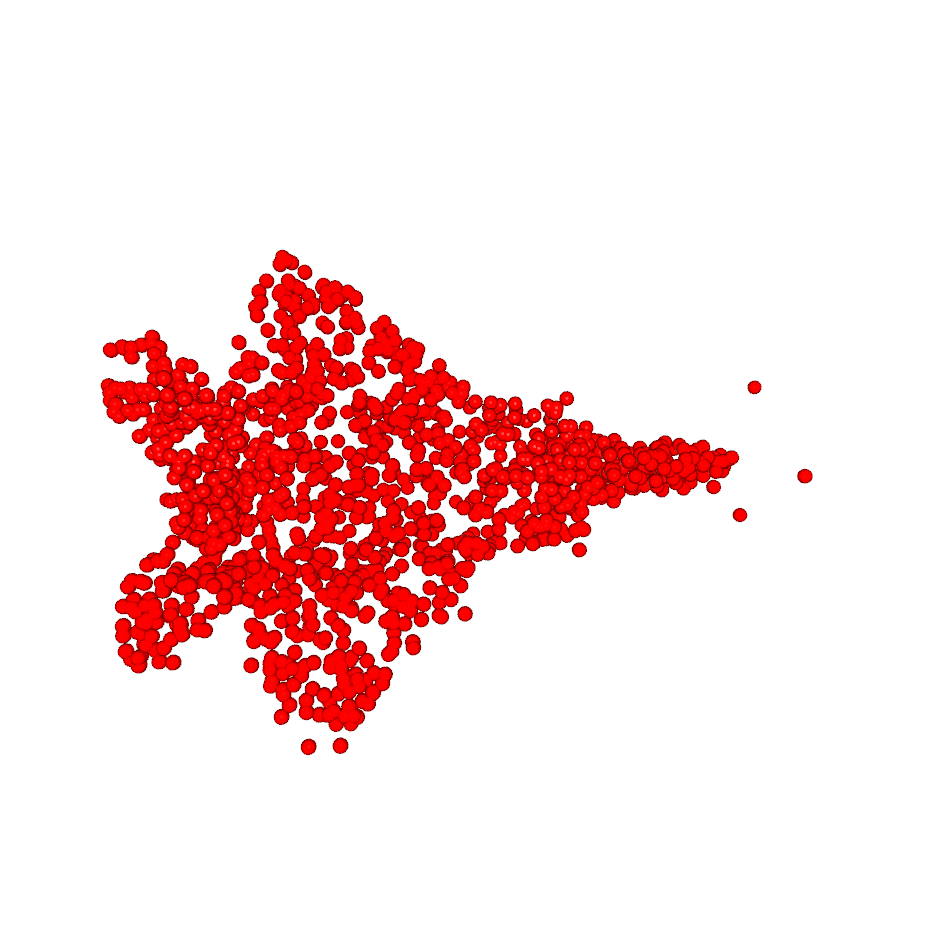} &
\includegraphics[width=0.12\linewidth]{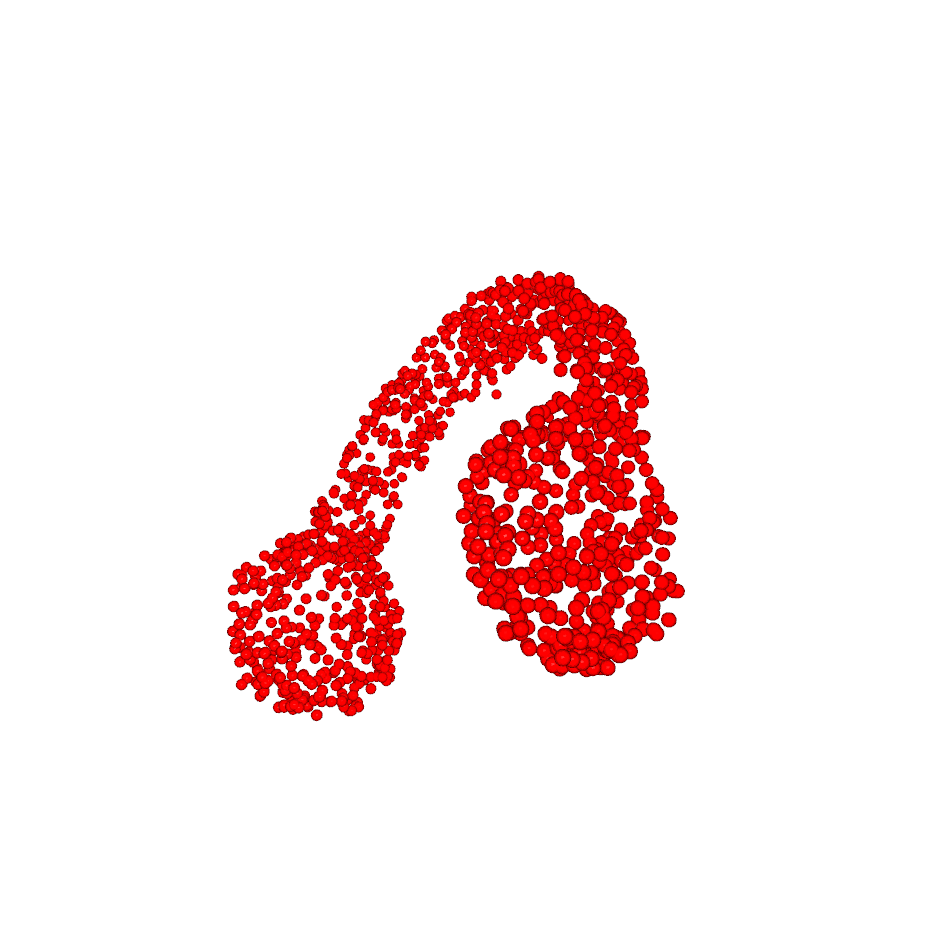} &
\includegraphics[width=0.12\linewidth]{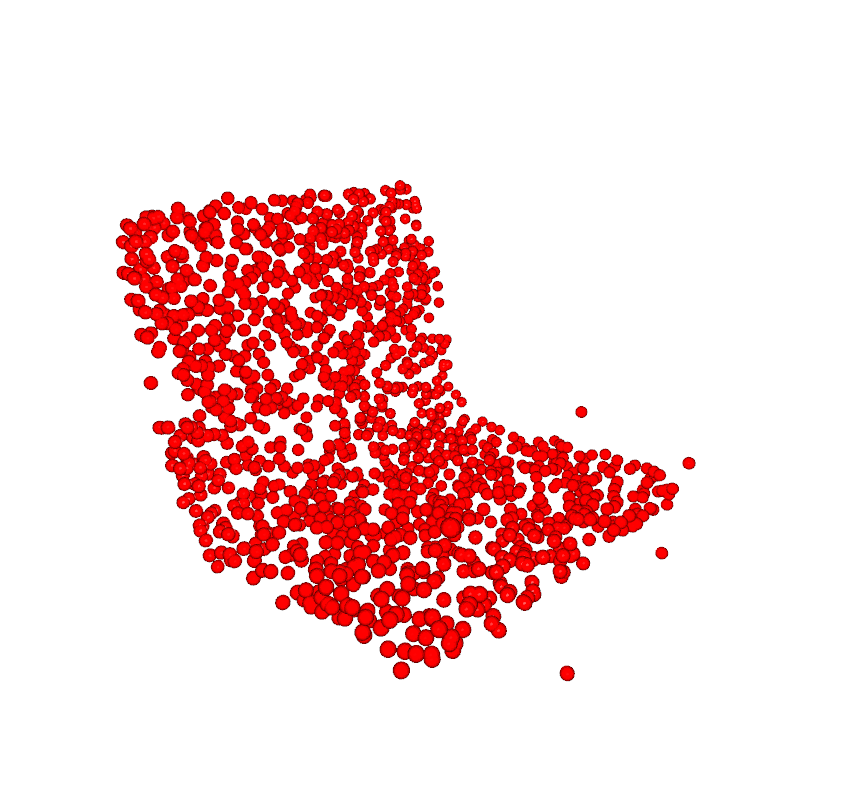} &
\includegraphics[width=0.12\linewidth]{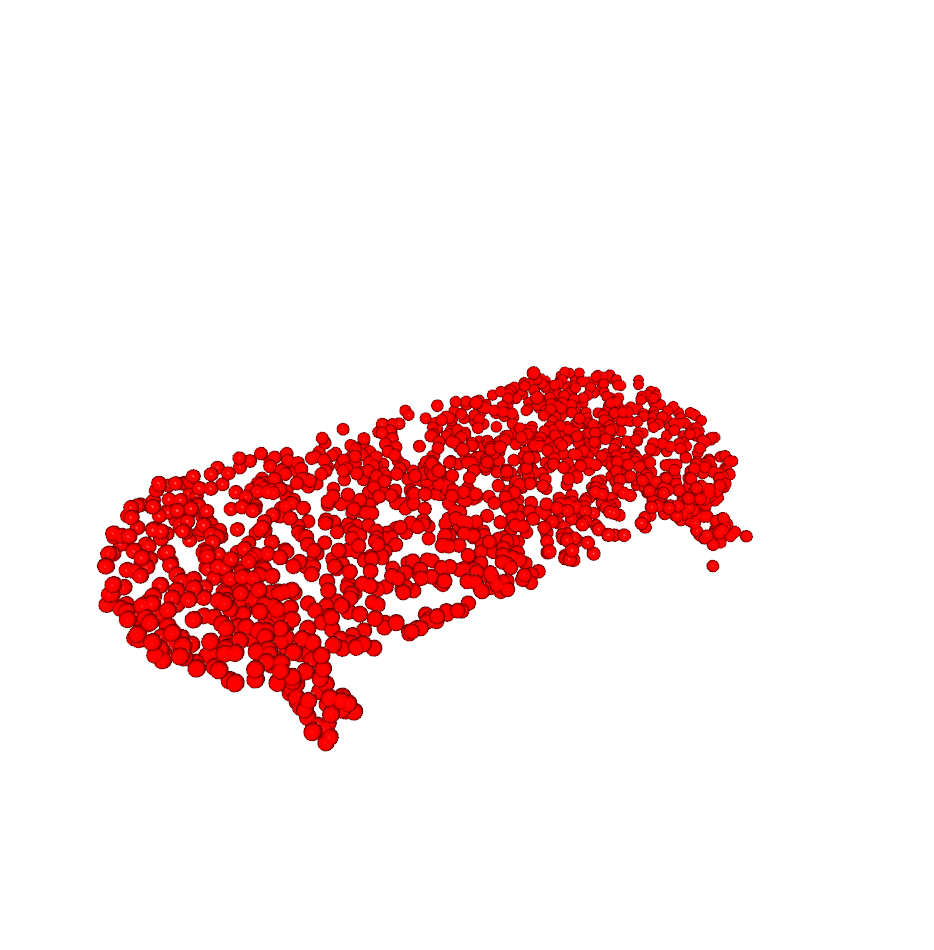} & 
\includegraphics[width=0.12\linewidth]{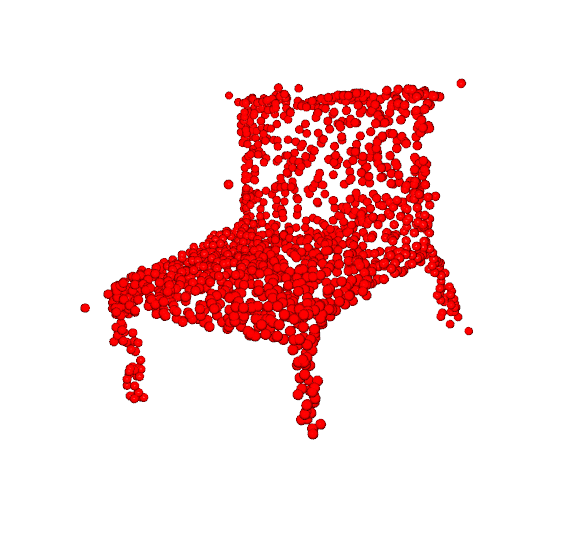}\\

& \textbf{\textcolor{red}{\scriptsize Motorbike}} &
\textbf{\textcolor{red}{\scriptsize Car}} &
\textbf{\textcolor{red}{\scriptsize Table}} &
\textbf{\textcolor{red}{\scriptsize Chair}} &
\textbf{\textcolor{red}{\scriptsize Car}} & \textbf{\textcolor{red}{\scriptsize Table}}\\


\parbox{2cm}{\centering {$P_\text{adv}$} \\ \scriptsize{(DGCNN)}} &  \includegraphics[width=0.12\linewidth]{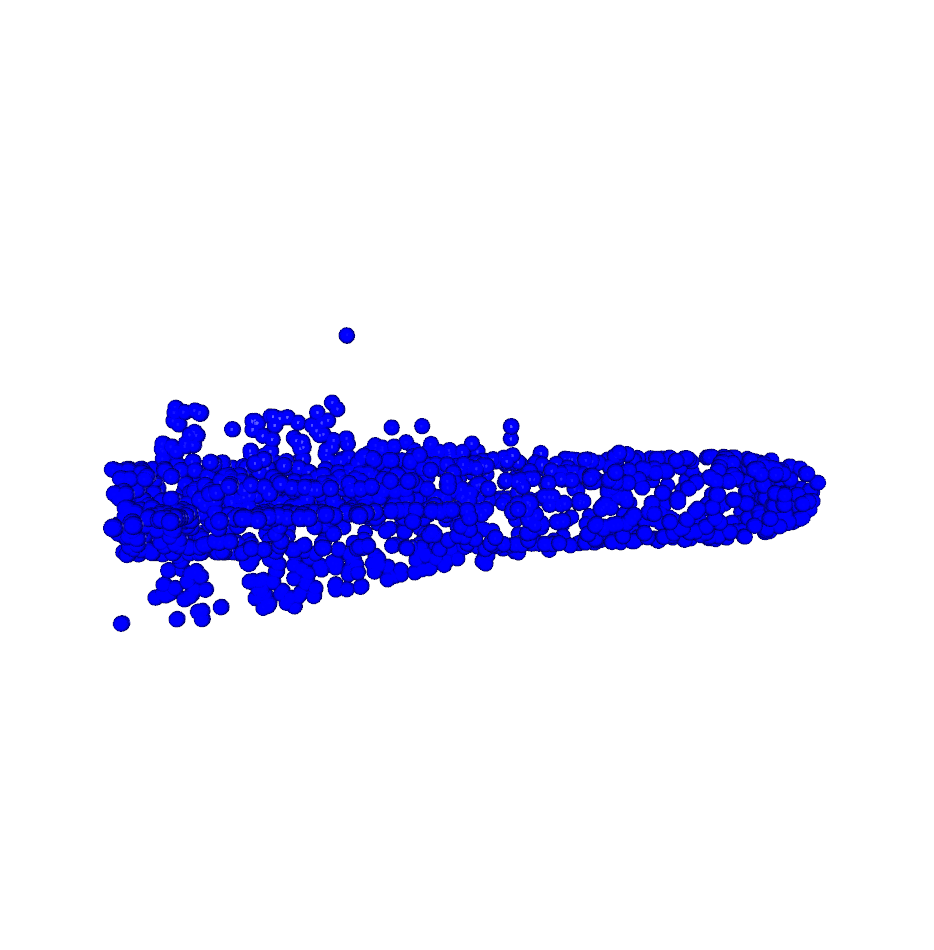} &
\includegraphics[width=0.12\linewidth]{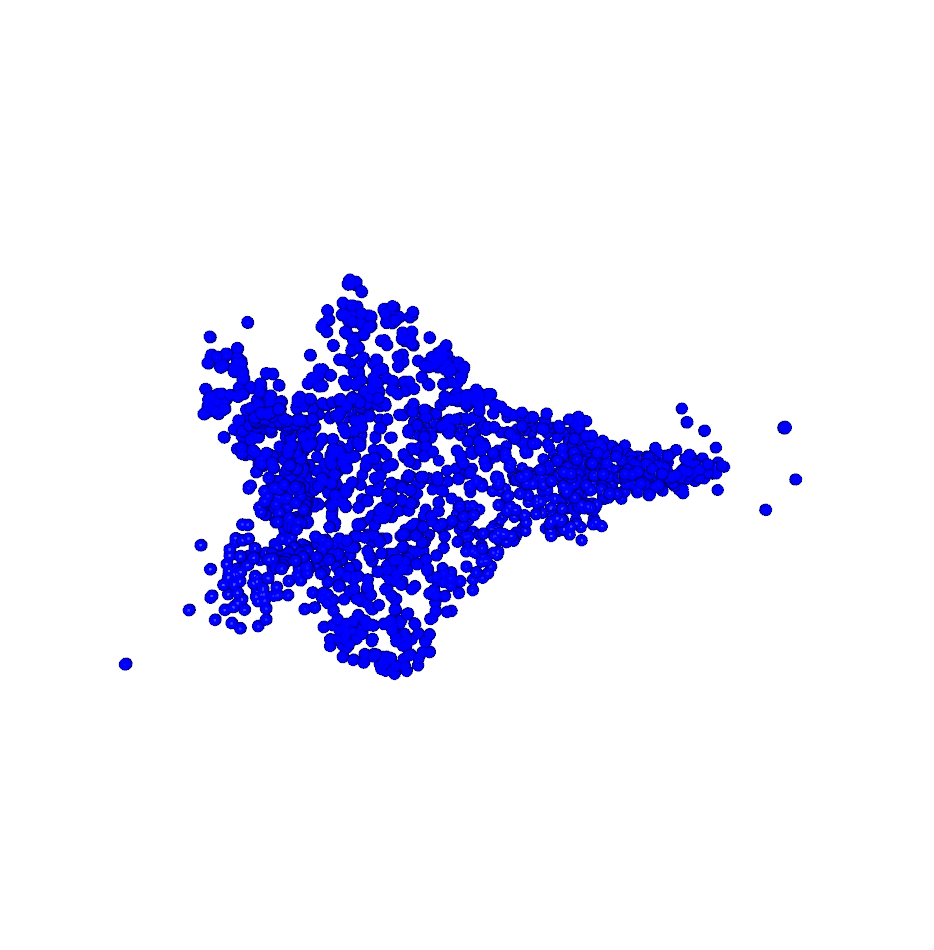} &
\includegraphics[width=0.12\linewidth]{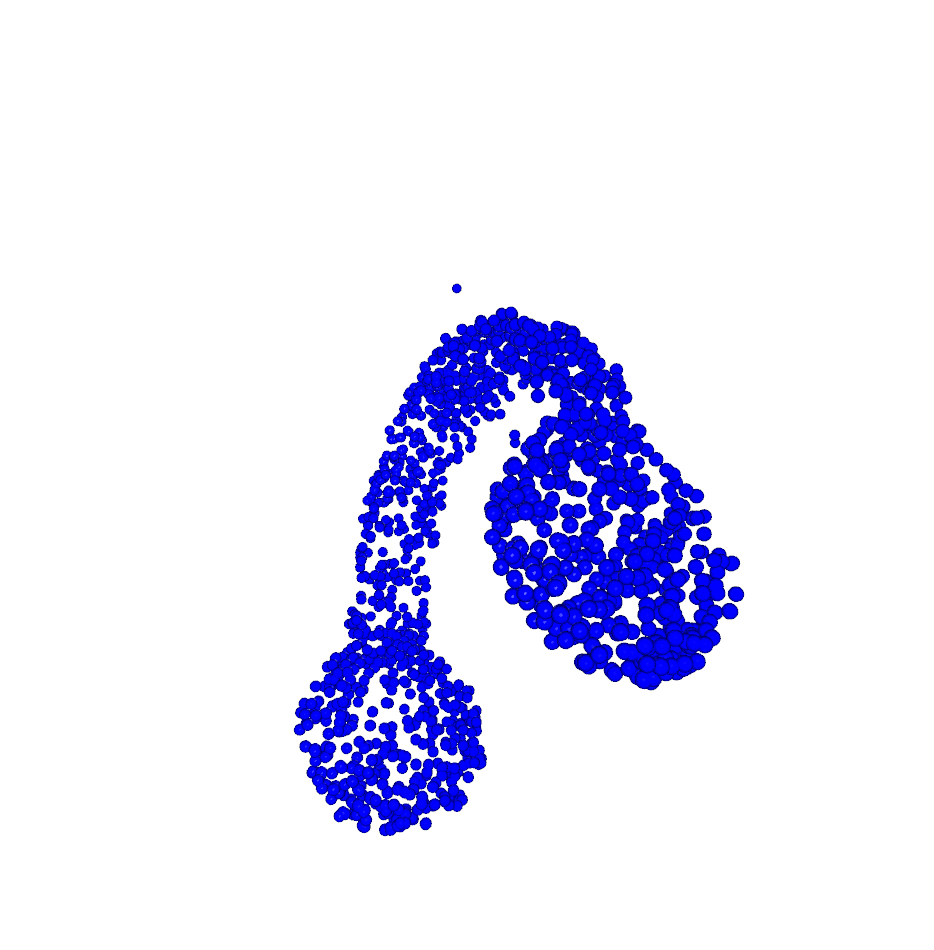} &
\includegraphics[width=0.1\linewidth]{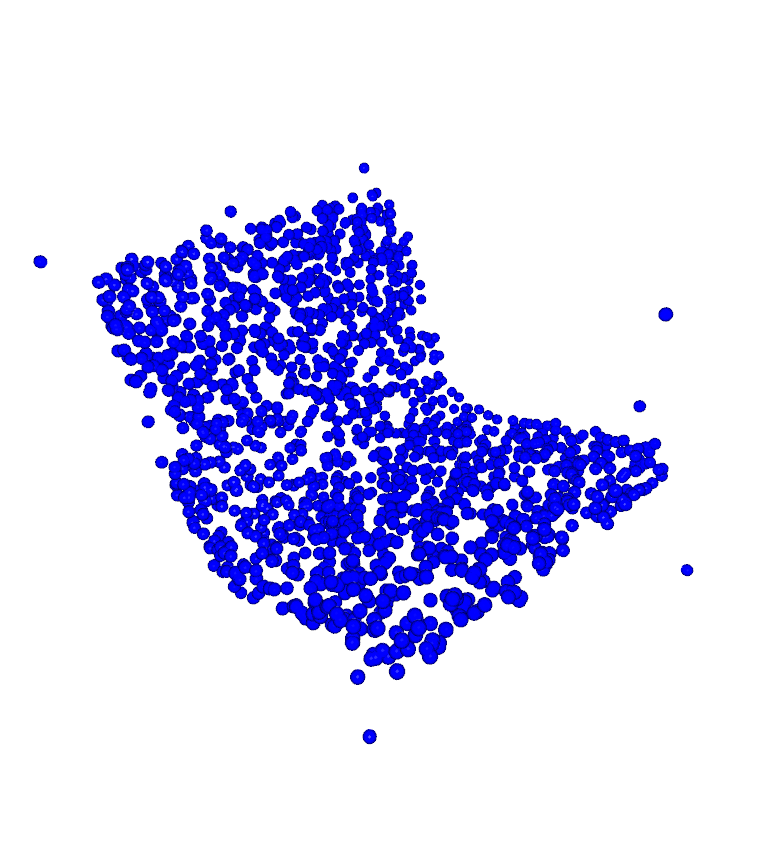} &
\includegraphics[width=0.12\linewidth]{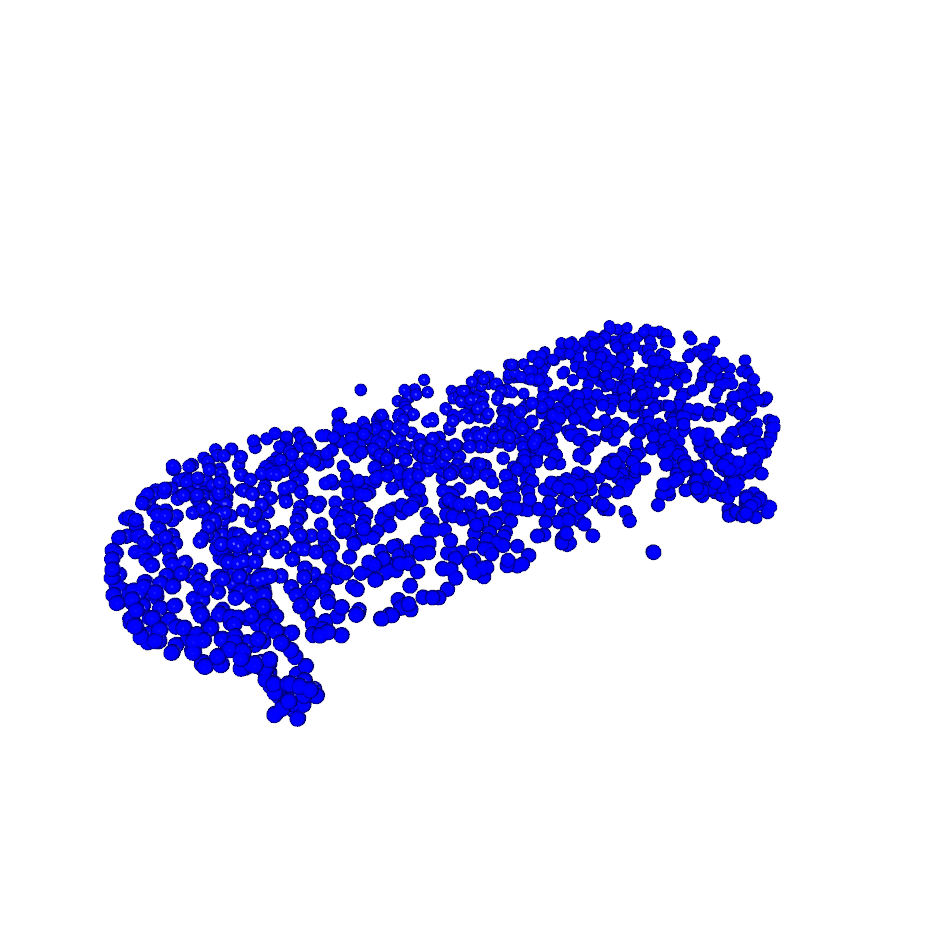} & 
\includegraphics[width=0.12\linewidth]{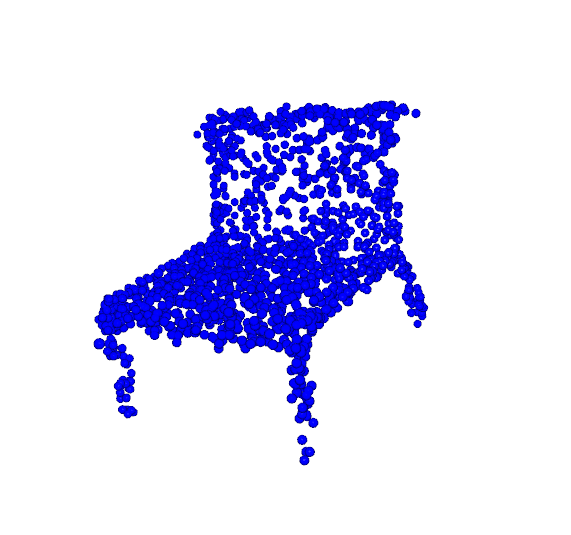}\\

&\textbf{\textcolor{blue}{\scriptsize Pistol}} &
\textbf{\textcolor{blue}{\scriptsize Table}} &
\textbf{\textcolor{blue}{\scriptsize Lamp}} &
\textbf{\textcolor{blue}{\scriptsize Chair}} &
\textbf{\textcolor{blue}{\scriptsize Lamp}} &
\textbf{\textcolor{blue}{\scriptsize Table}}\\

\end{tabular}

\caption{Visualization of representative original and adversarial ShapeNet Part point clouds generated by Topo-ADV, together with their misclassified labels. The top row represents the clean point clouds, the middle and bottom rows present the adversarial point clouds generated with PointNet and DGCNN as victim models, respectively.}
\label{fig:adv_examples}
\end{figure}

Despite topology manipulation, the proposed method maintains competitive geometric fidelity. While the CSD metric is comparable to several strong baselines (third lowest among the eleven in Table \ref{tab:modelnet40_attack}), Topo-ADV consistently achieves substantially improved uniformity preservation. For example, under PointNet on ModelNet40, Topo-ADV achieves a uniform distance of $0.0121$ compared to $0.2883$ for HiT-ADV (Table~\ref{tab:modelnet40_attack}), representing an order-of-magnitude ($23.83\times$) improvement. On ShapeNet, Topo-ADV achieves perfect ASR while producing imperceptible attacks, which is evident by the lowest uniform distance score (Table ~\ref{tab:shapenet_attack}). Representative visualizations are presented in Fig. \ref{fig:adv_examples}.

A low Uniform score indicates that the adversarial point cloud preserves the spatial sampling density of the original surface extremely well, i.e., the local distribution of points within neighborhoods remains statistically consistent. This reflects high perceptual plausibility, since humans are sensitive to irregular density patterns on surfaces. In contrast, a slightly higher CSD suggests that the perturbation induces modest variations in local curvature statistics. Because curvature captures second‑order geometric structure rather than sampling regularity, this indicates that the attack modifies the underlying surface shape just enough to affect topology or classification, while still maintaining highly uniform point spacing.
Together, this combination implies that the perturbation is globally subtle and surface‑consistent (low Uniform), yet locally effective in altering geometric complexity (moderate CSD), achieving a strong trade‑off between imperceptibility and attack strength. 

Topo-ADV demonstrates significantly improved geometric imperceptibility compared to CFG~\cite{pang2025towards} on ModelNet40 (Table \ref{tab:cd_pang}). In particular, it reduces the Chamfer distance from $0.32$ to $0.0024$ and and achieves a lower Hausdorff distance ($0.23$ vs.\ $0.36$), indicating substantially better surface fidelity. Although the $\mathcal{L}_2$ displacement is moderately higher, the results suggest that topology-driven perturbations remain visually subtle while preserving overall geometric consistency.

\noindent
\textbf{ScanObjectNN Results. }The results on the real-world ScanObjectNN dataset\cite{uy2019revisiting} further confirm the robustness of the proposed attack. Topo-ADV achieves a $100\%$ ASR across both PointNet and DGCNN while maintaining low geometric distortion (e.g., Chamfer distance of $0.0037$ and $0.0396$, respectively), indicating that topology-driven perturbations remain highly effective even on noisy, real-world point cloud data.



\begin{table*}[ht!]
    \centering
    \begin{tabular}{c|ccc}
    \toprule
        Method & Chamfer distance$\downarrow$ & ~Hausdorff distance$\downarrow$ & ~$\mathcal{L}_2$ distance$\downarrow$\\
        \midrule CFG~\cite{pang2025towards}&0.32 & 0.36 & \textbf{0.22}\\
         Topo-ADV (ours) & \textbf{0.0024} & \textbf{0.23} & 0.54\\
         \bottomrule
    \end{tabular}
    \caption{Comparison of imperceptibility on ModelNet40 dataset with \cite{pang2025towards}.}
    \label{tab:cd_pang}
\end{table*}




\begin{table*}[ht!]
    \centering
    \begin{tabular}{cccccc}
    \toprule
        Metric \textbackslash~Config~~ & Default & ~$R=0$~ & ~$R=0$, $T=100$~ & ~w/ $\mathcal{L}_{\text{cls}}$,   $\mathcal{L}_{\text{PH}}$~ & ~w/ $\mathcal{L}_{\text{cls}}$~\\
        \midrule
        ASR (\%) $\uparrow$ & 99.86 & 99.47 & 99.04 & 99.09 & 99.19\\
        \bottomrule
    \end{tabular}
    \caption{Ablation results with PointNet and ModelNet40.}
    \label{tab:ablation}
\end{table*}

\begin{table}[ht!]
    \centering
    \begin{tabular}{cccc}
    \toprule
        $\epsilon$ & ASR (\%) $\uparrow$ & CSD $\downarrow$ & Uniform $\downarrow$\\
        \midrule
         0.55 (default) & 99.86 & 1.2328 & 0.0121\\ 
         0.75 & 99.90 & 1.2300 & 0.0121\\
         1.0 & 99.95 & 1.2322 & 0.0121\\
         \bottomrule
         
    \end{tabular}
    \caption{Analyzing the effect of $\epsilon$ with PointNet and ModelNet40.}
    \label{tab:eps_effect}
\end{table}

\medskip
\noindent
\textbf{Ablation Study.}
To better approximate the worst-case vulnerability under a fixed perturbation budget, we report attack success with $R\in\{0,2\}$ random restarts. Each restart uses identical hyperparameters and constraints and differs only in the random initialization. A sample is considered successfully attacked if any of the $R$ restarts yield misclassification. Even without restarts ($R=0$), the proposed method achieves an ASR of $99.47\%$, demonstrating that a single optimization trajectory is already sufficient to reliably fool the classifier (Table \ref{tab:ablation}). Decreasing the iteration budget to $T=100$ with $R=0$, a slightly lower ASR of $99.04\%$ is achieved, as expected. The default configuration ($R=2$, $T=300$) provides only marginal gains, reaching $99.86\%$ (Table \ref{tab:ablation}). These results indicate that Topo-ADV does not rely on aggressive multi-start optimization and remains highly effective under reduced computational budgets.

We further analyze the sensitivity of the proposed attack to the perturbation budget $\epsilon$ on PointNet using ModelNet40 (Table~\ref{tab:eps_effect}). The results show that Topo-ADV remains highly effective across different perturbation budgets, achieving consistently high ASR values exceeding $99.8\%$. Increasing $\epsilon$ provides only marginal improvements in attack success while maintaining nearly identical uniformity preservation and comparable CSD. This indicates that the proposed topology-driven perturbations are already sufficient to reliably fool the classifier under relatively small displacement constraints.

\medskip

\noindent
\textbf{Measuring Change in Persistence After Attack.} 
For a given point cloud $P$, the \textit{persistence entropy}~\cite{rucco2016characterisation} of its persistence diagram $D(P,k)=\{(b_i,d_i)\}_{i\in I}$ in dimension $k$ is defined as: $$E(P,D(P,k))=-\sum_{i\in I} p_i\log(p_i), \text{ where } p_i = \frac{d_i-b_i}{L} \text{ and } L = \sum_{i\in I} (d_i-b_i).$$

We slightly abuse the $E(P,D(P,k))$ notation by using $E_k$ only. 
In our case, we say that $P$ and $P_\text{adv}$ have different persistence (hence, different topology), if $E(P,D(P,k))\neq E(P_\text{adv}, D(P_\text{adv},k))$, for some $k \in \{0,1,2\}$. Across all evaluated settings, the proposed attack produces minimal entropy shifts in $H_0$ and $H_1$ (avg.\ $\Delta E_0 \approx 0.06$, $\Delta E_1 \approx -0.03$), while inducing a substantially larger shift in $H_2$ (avg.\ $\Delta E_2 \approx -1.49$). This consistent pattern across ModelNet40/ShapeNet and PointNet/DGCNN indicates that Topo-ADV primarily disrupts higher-order topological structures.

\section{Conclusion}

In this work, we introduced a novel topology-driven adversarial attack framework for 3D point cloud learning that treats persistent homology as an explicit optimization objective rather than a post-hoc analytical tool. By integrating differentiable persistent homology embeddings within an end-to-end optimization pipeline, the proposed method manipulates topological structure while preserving geometric plausibility through carefully designed imperceptibility constraints.

Extensive evaluations across multiple synthetic and real-world 3D point cloud datasets and notable victim architectures demonstrate that the proposed approach consistently achieves near-perfect attack success rates under identical perturbation budgets. Despite explicitly altering topological signatures, the generated adversarial examples maintain strong geometric fidelity, achieving many significantly improved imperceptibility metrics compared to existing geometry-driven attacks. Sensitivity analysis further shows that the method remains highly effective even under small perturbation budgets, highlighting topology manipulation as a powerful and previously overlooked vulnerability surface in point cloud deep learning.

\bibliographystyle{abbrv}
\bibliography{refs}

\newpage
\section*{Appendix}
\subsection*{Transferability}
Table~\ref{tab:transfer} reports the transferability results on ModelNet40 with $\epsilon = 0.18$, where adversarial point clouds are generated using DGCNN and evaluated on PointNet as the victim model. Our proposed Topo-ADV achieves a transfer-ASR of 21.2\%, substantially outperforming prior geometry-based approaches such as 3D-Adv (9.2\%), KNN (7.2\%), GeoA$^3$ (4.4\%), and AdvPC (19.6\%), while practically matching the transferability performance of PF-Attack (21.3\%), indicating improved cross-model generalization of the generated adversarial perturbations. 

\begin{table}[ht!]
    \centering
    \begin{tabular}{cc}
    \toprule
        Method & ASR (\%)\\
        \midrule
         3D-Adv~\cite{xiang2019generating} & 9.2\\
         KNN~\cite{tsai2020robust} & 7.2\\
         GeoA$^3$~\cite{wen2020geometry} & 4.4\\
         AdvPC~\cite{hamdi2020advpc} & 19.6\\
         Topo-ADV (Ours) & 21.2\\
         PF-Attack~\cite{he2023generating} & 21.3\\
         \bottomrule
    \end{tabular}
    \caption{Transferability results with $\epsilon = 0.18$ and ModelNet40, where the adversarial point clouds were generated using DGCNN and the victim model was PointNet. The numbers for the baselines in this table are from \cite{he2023generating}, and we used the same $\epsilon$ for this comparison.}
    \label{tab:transfer}
\end{table}

For the converse setting (Table \ref{tab:transfer_attackPN2victimDGCNN}), where adversarial point clouds are generated using PointNet and evaluated on the victim model DGCNN, our Topo-ADV achieves a transfer-ASR of 22.9\%, outperforming several prior methods, including 3D-Adv (6.8\%), KNN (6.0\%), GeoA$^3$ (7.2\%), and AdvPC (22.4\%). The result is also competitive with recent strong baselines such as PF-Attack (24.8\%), indicating that the proposed topological perturbations generalize well across architectures.

\begin{table}[ht!]
    \centering
    \begin{tabular}{cc}
    \toprule
        Method & ASR (\%)\\
        \midrule
         3D-Adv~\cite{xiang2019generating} & 6.8\\
         KNN~\cite{tsai2020robust} & 6.0\\
         GeoA$^3$~\cite{wen2020geometry} & 7.2\\
         AdvPC~\cite{hamdi2020advpc} & 22.4\\
         Topo-ADV (Ours) & 22.9\\
         PF-Attack~\cite{he2023generating} & 24.8\\
         \bottomrule
    \end{tabular}
    \caption{Transferability results with $\epsilon = 0.18$ and ModelNet40, where the adversarial point clouds were generated using PointNet and the victim model was DGCNN. The numbers for the baselines in this table are from \cite{he2023generating}, and we used the same $\epsilon$ for this comparison.}
    \label{tab:transfer_attackPN2victimDGCNN}
\end{table}

\subsection*{Visualizations}
\begin{figure}[ht]
\centering
\begin{tabular}{c|c|c|c|c|c|c}

$P$ & \includegraphics[width=0.12\linewidth]{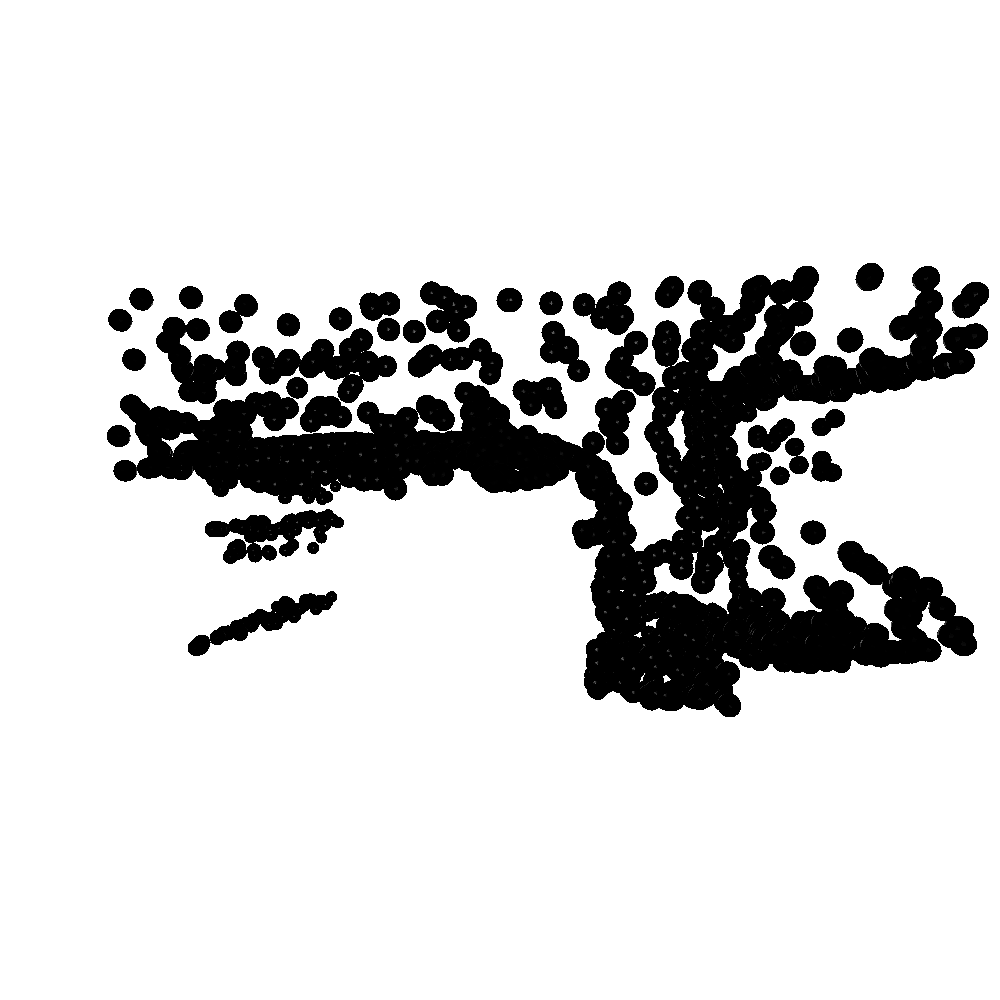} &
\includegraphics[width=0.12\linewidth]{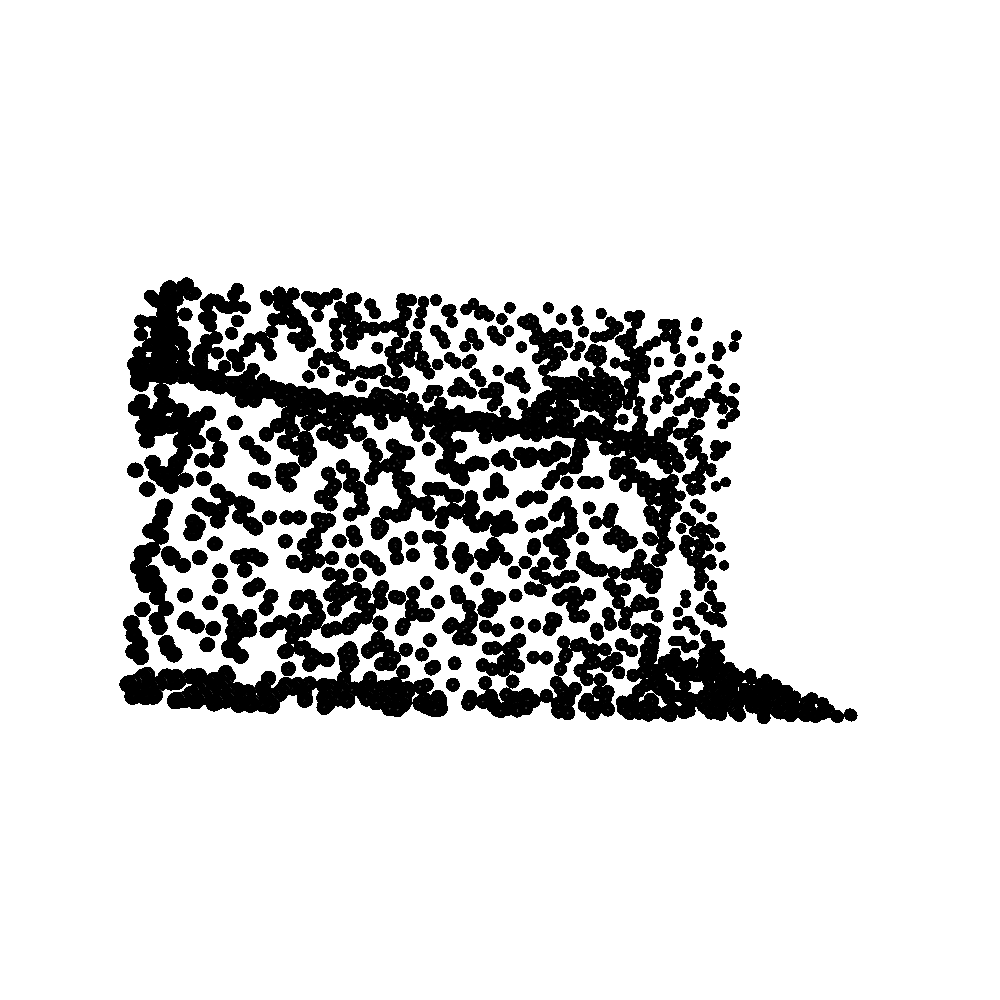} &
\includegraphics[width=0.12\linewidth]{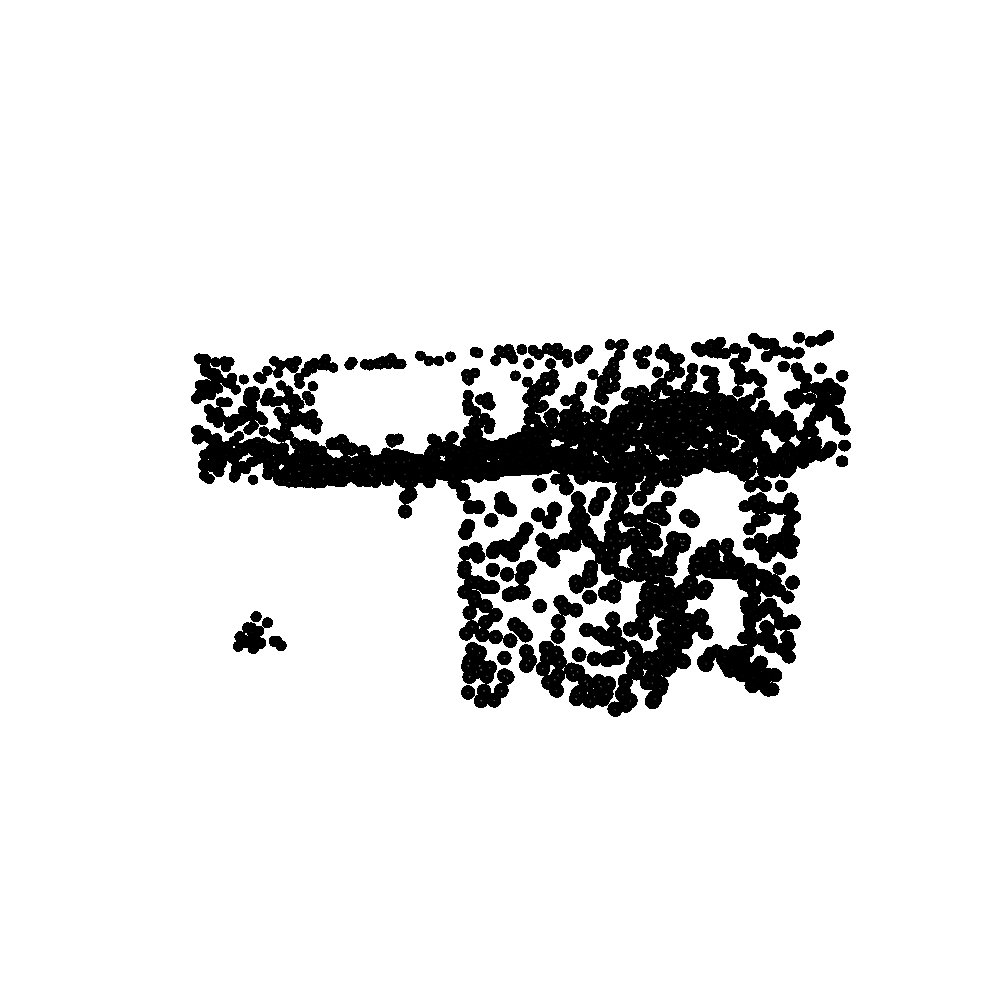} &
\includegraphics[width=0.12\linewidth]{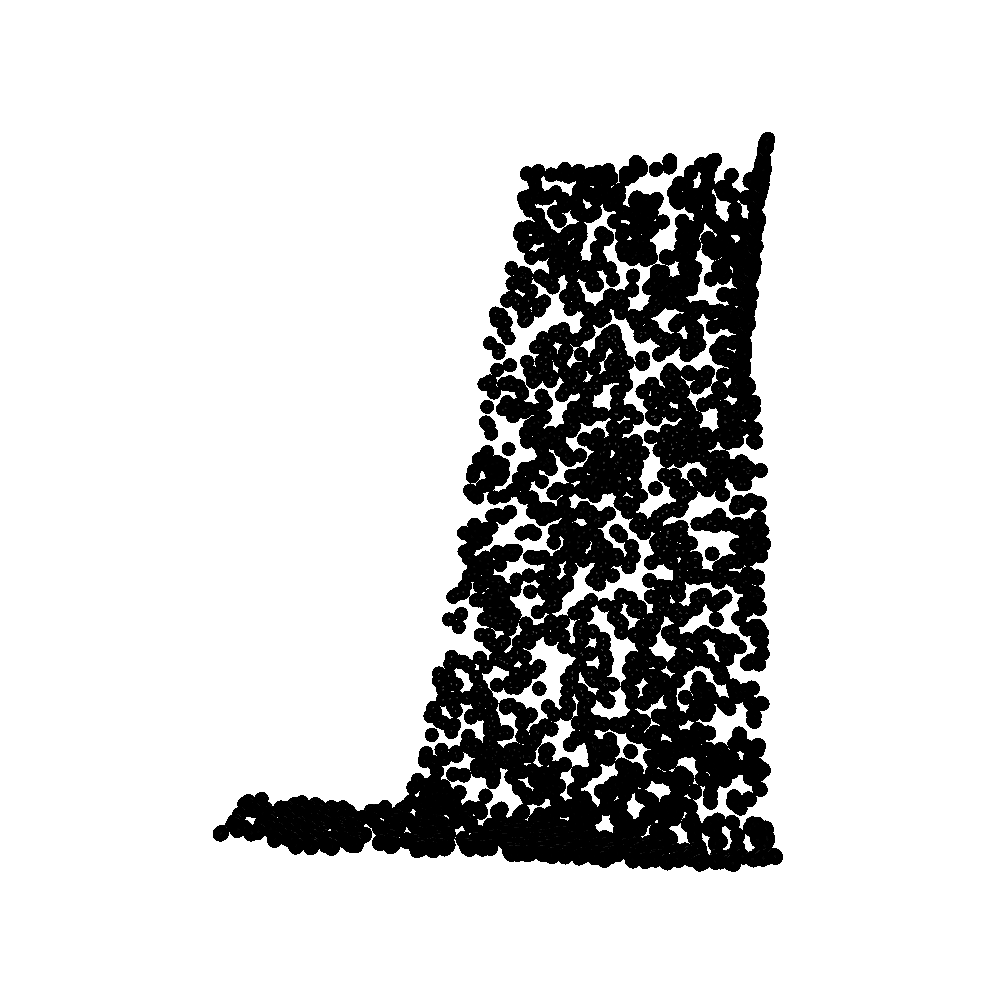} &
\includegraphics[width=0.12\linewidth]{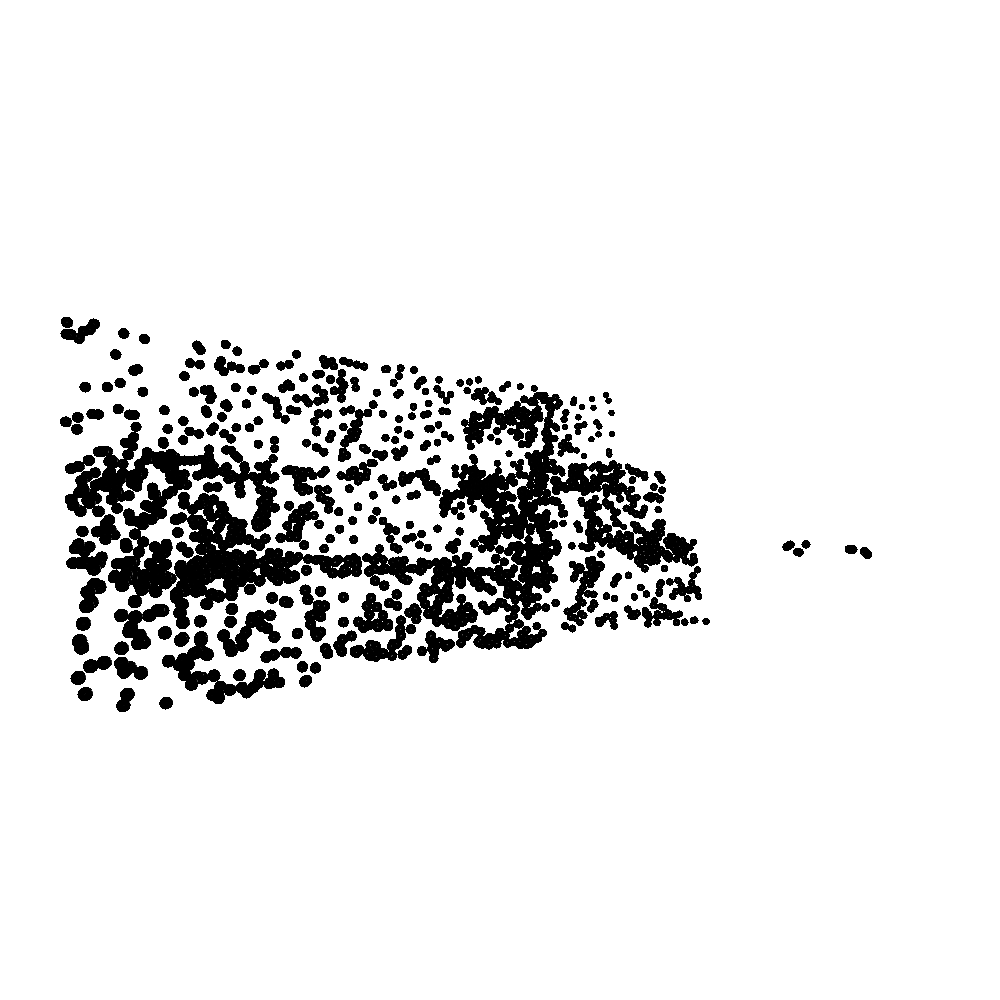} & 
\includegraphics[width=0.12\linewidth]{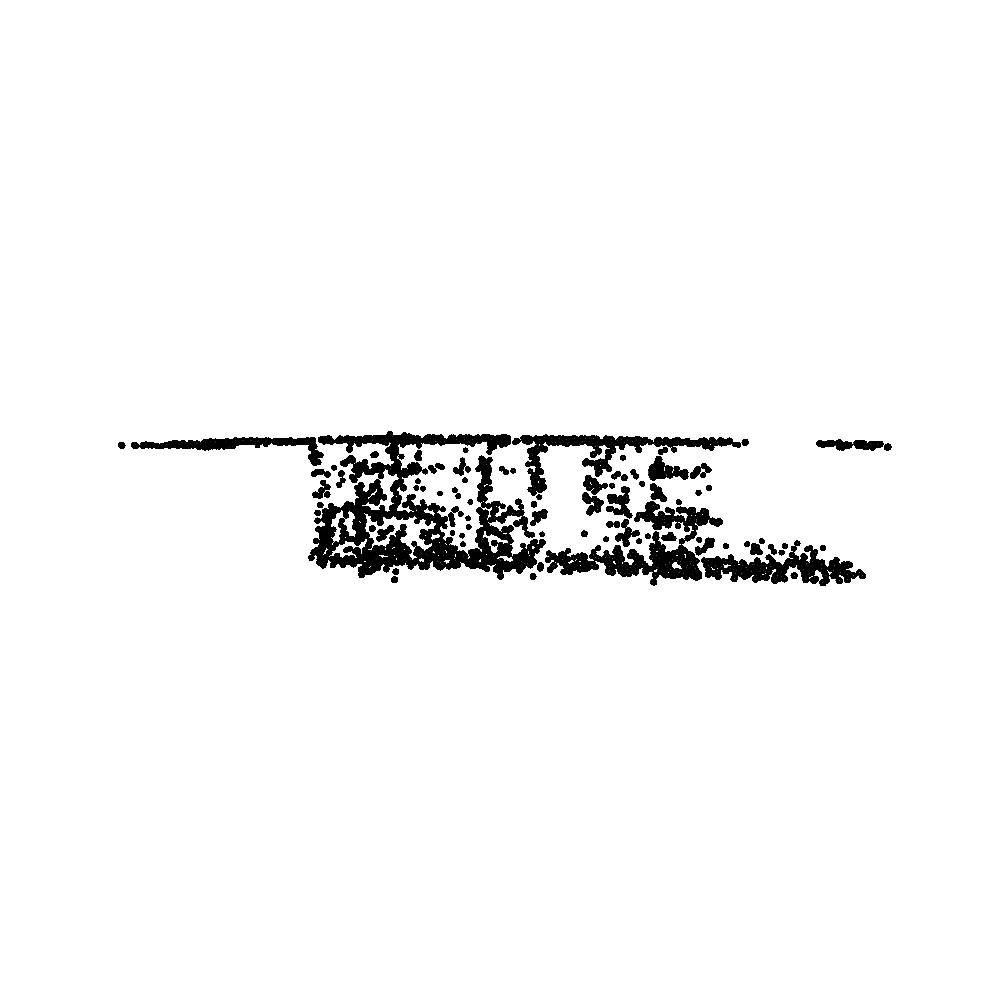}
\\

&  \textbf{\scriptsize {Bed}} &
\textbf{\scriptsize Cabinet} &
\textbf{\scriptsize Desk} &
\textbf{\scriptsize Door} &
\textbf{\scriptsize Sofa} &  \textbf{\scriptsize Table} \\


\parbox{2cm}{\centering {$P_\text{adv}$} \\ \scriptsize{(PointNet)}} & \includegraphics[width=0.12\linewidth]{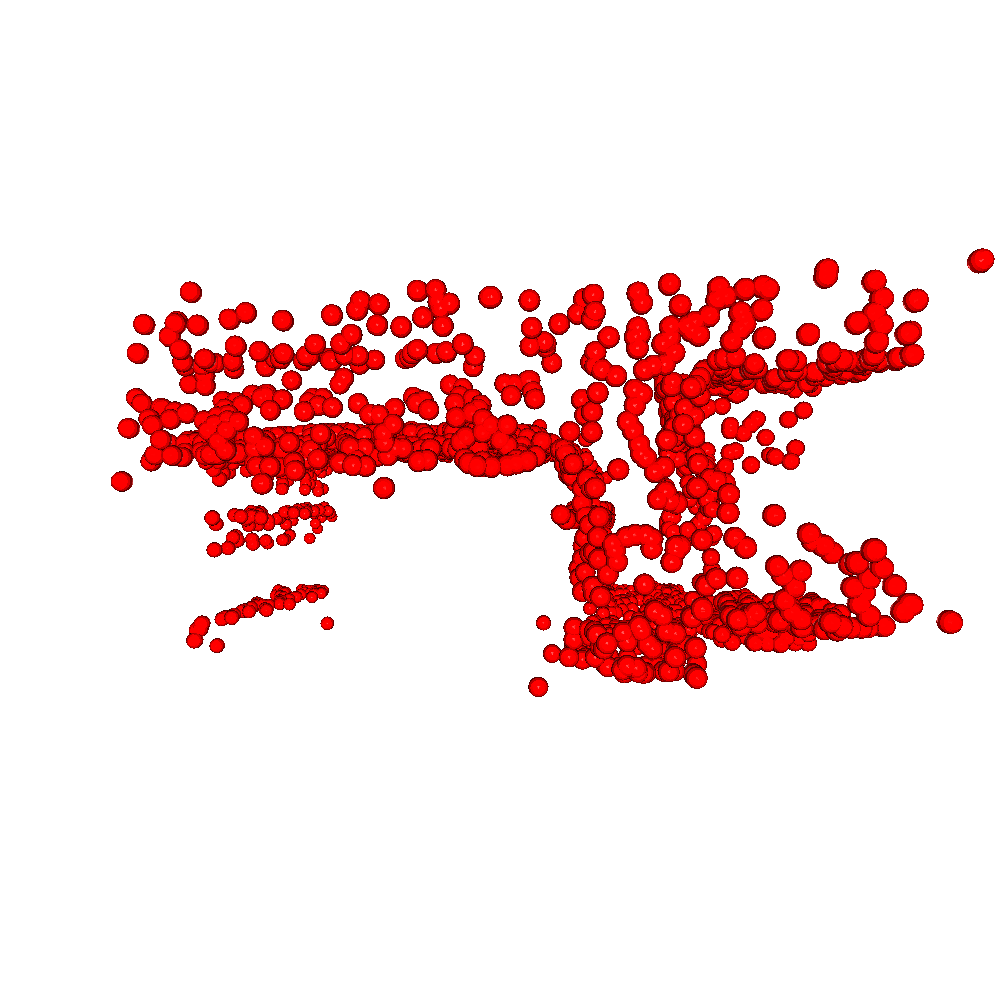} &
\includegraphics[width=0.12\linewidth]{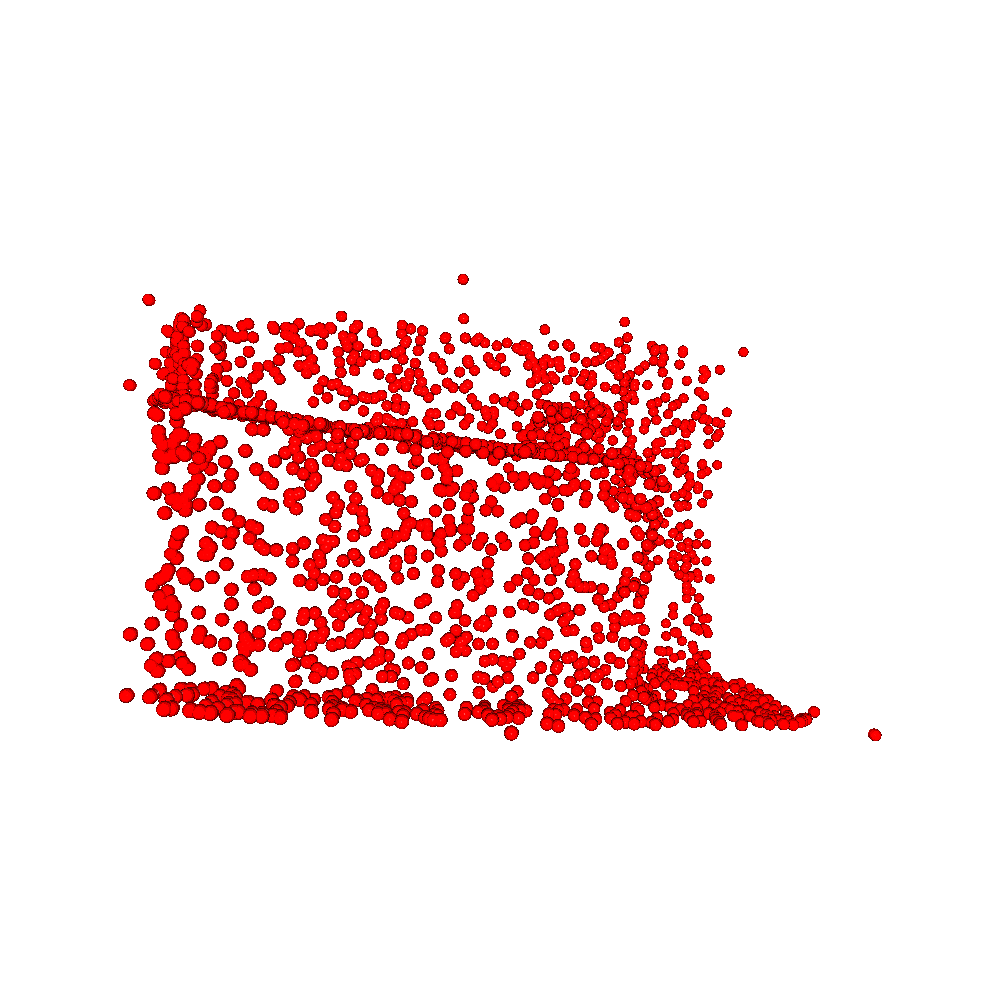} &
\includegraphics[width=0.12\linewidth]{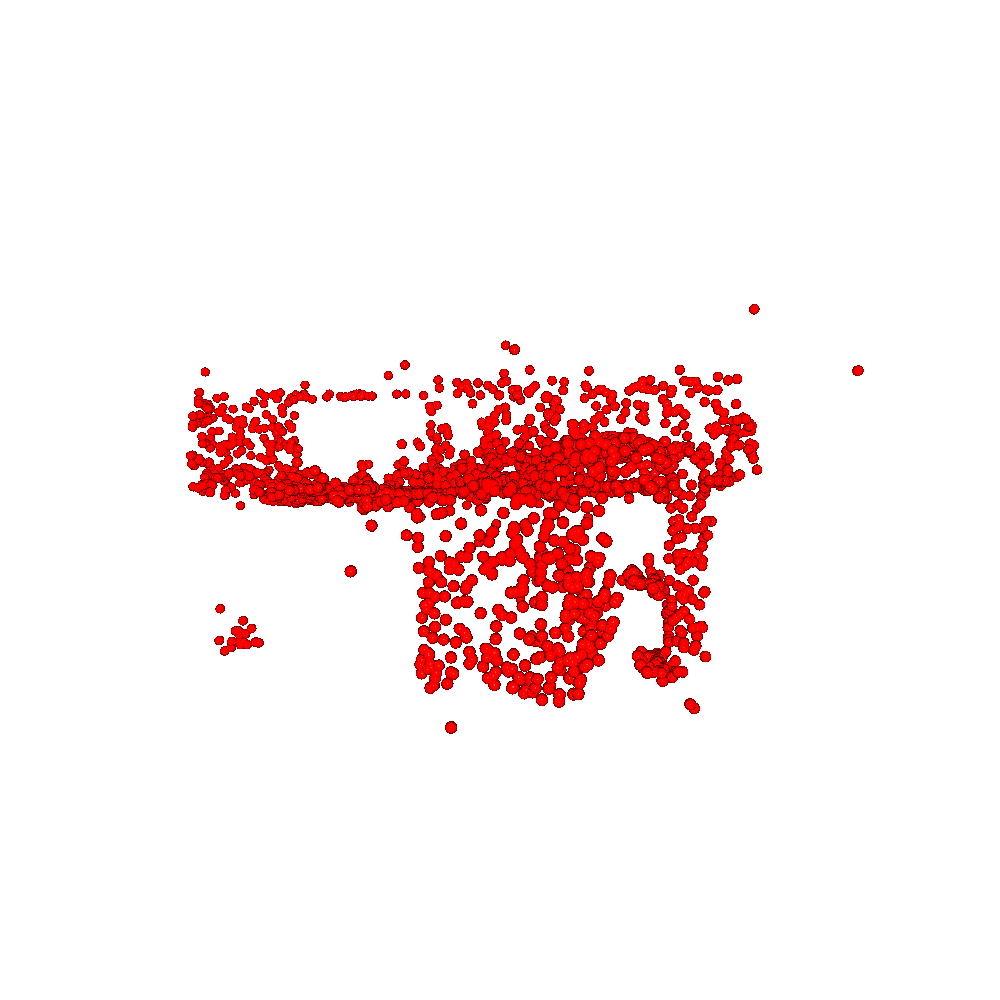} &
\includegraphics[width=0.12\linewidth]{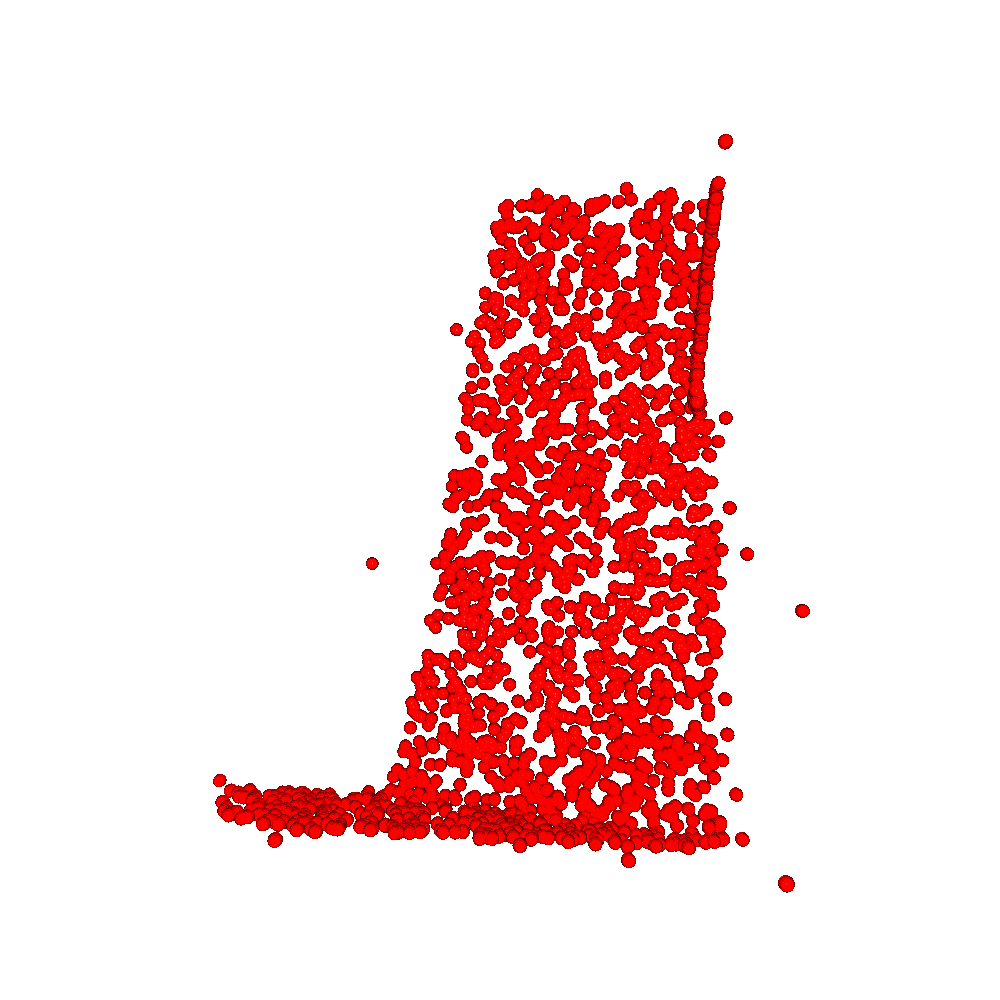} &
\includegraphics[width=0.12\linewidth]{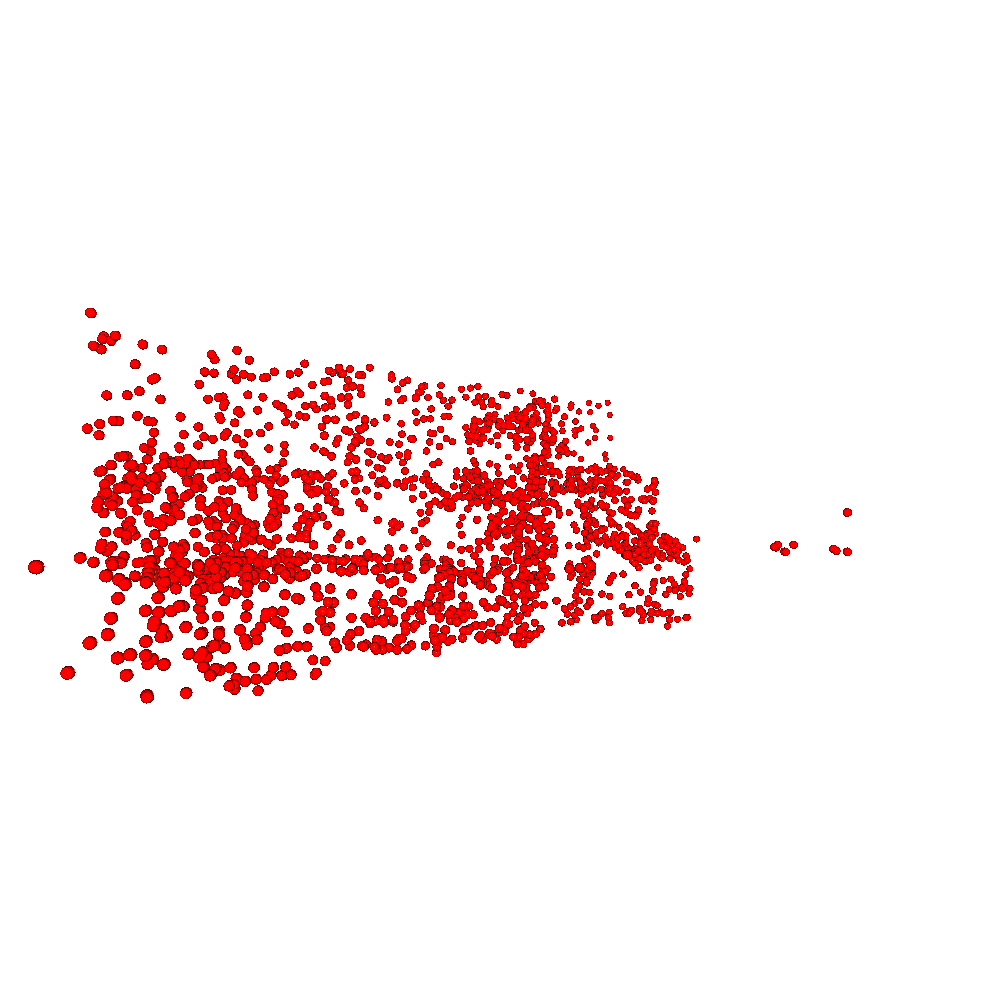} & 
\includegraphics[width=0.12\linewidth]{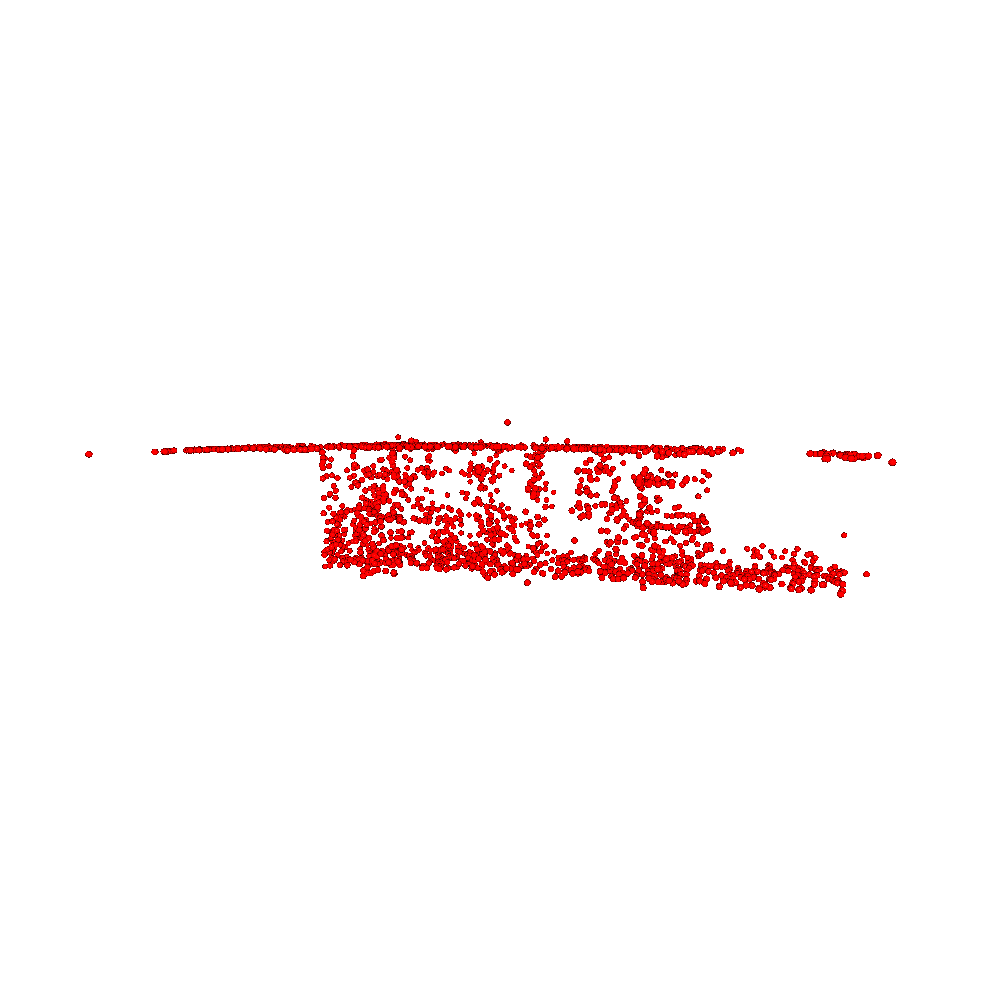}\\

& \textbf{\textcolor{red}{\scriptsize Table}} &
\textbf{\textcolor{red}{\scriptsize Box}} &
\textbf{\textcolor{red}{\scriptsize Box}} &
\textbf{\textcolor{red}{\scriptsize Cabinet}} &
\textbf{\textcolor{red}{\scriptsize Shelf}} & \textbf{\textcolor{red}{\scriptsize Shelf}}\\


\parbox{2cm}{\centering {$P_\text{adv}$} \\ \scriptsize{(DGCNN)}} &  \includegraphics[width=0.12\linewidth]{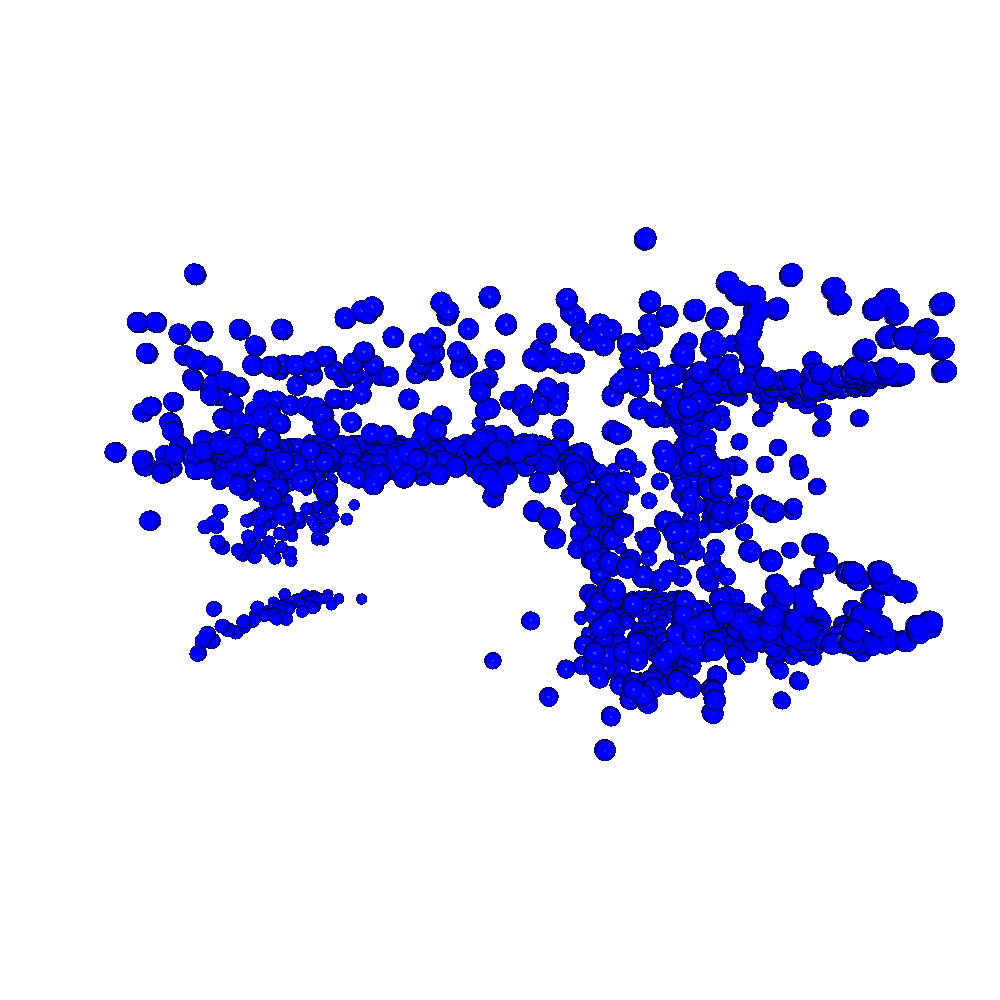} &
\includegraphics[width=0.12\linewidth]{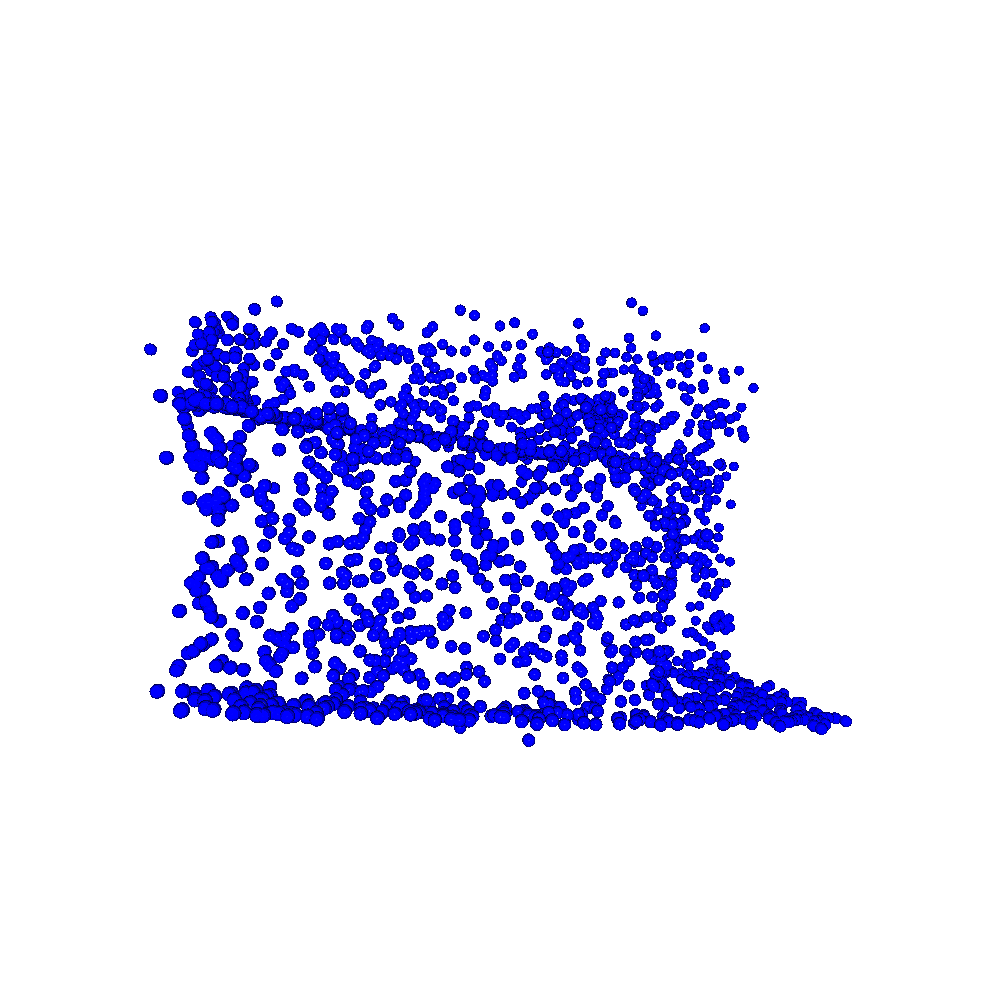} &
\includegraphics[width=0.12\linewidth]{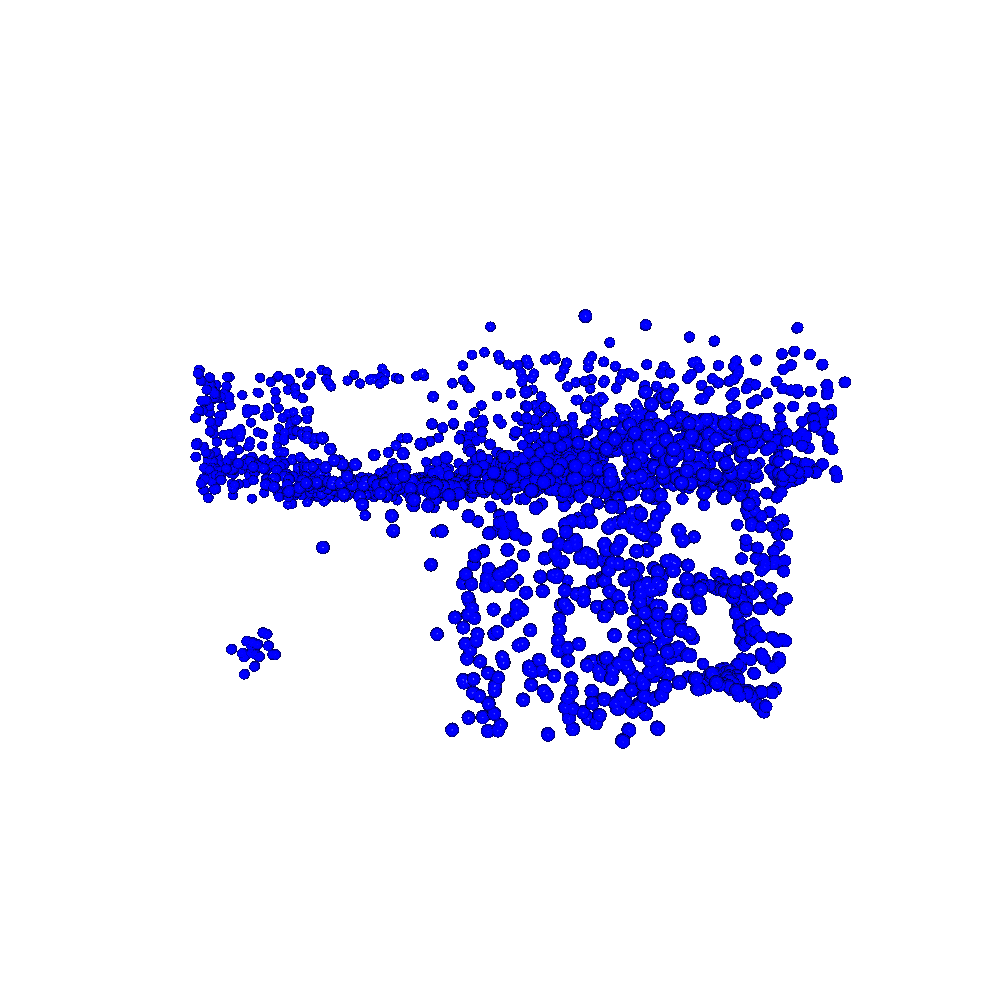} &
\includegraphics[width=0.1\linewidth]{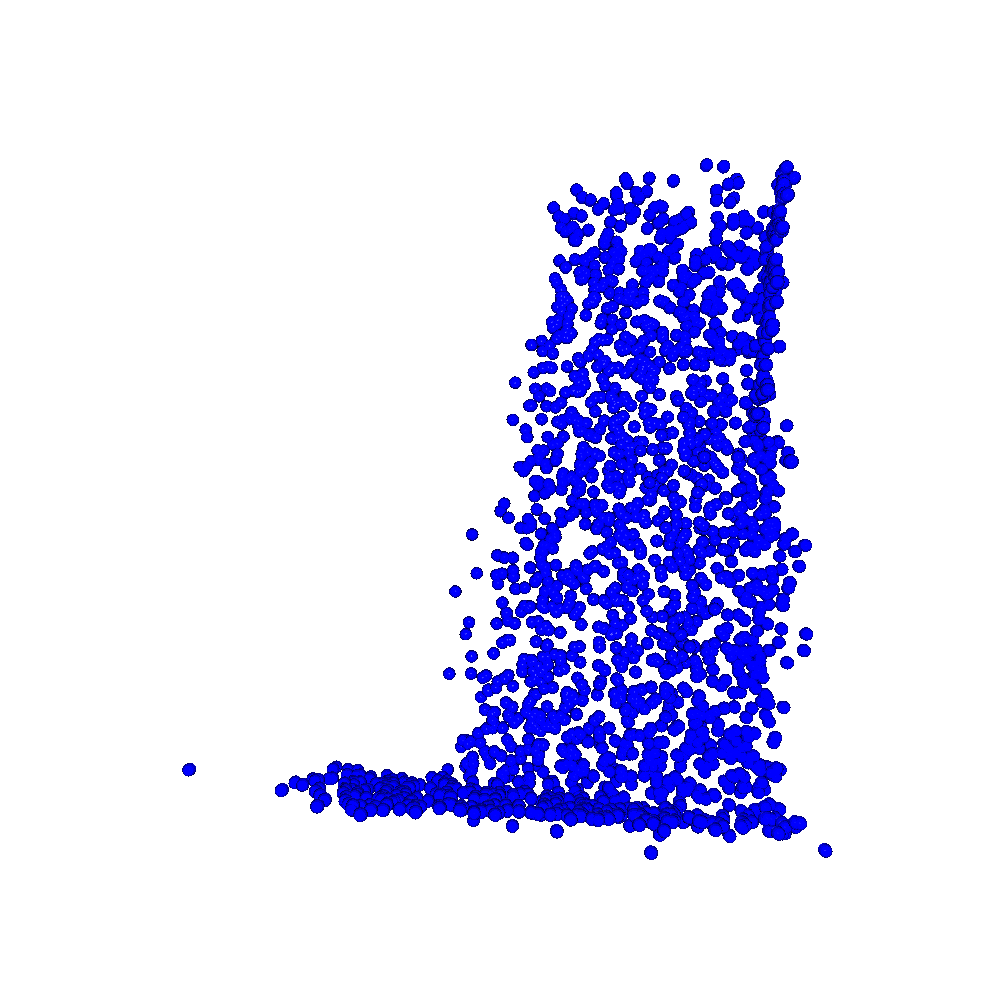} &
\includegraphics[width=0.12\linewidth]{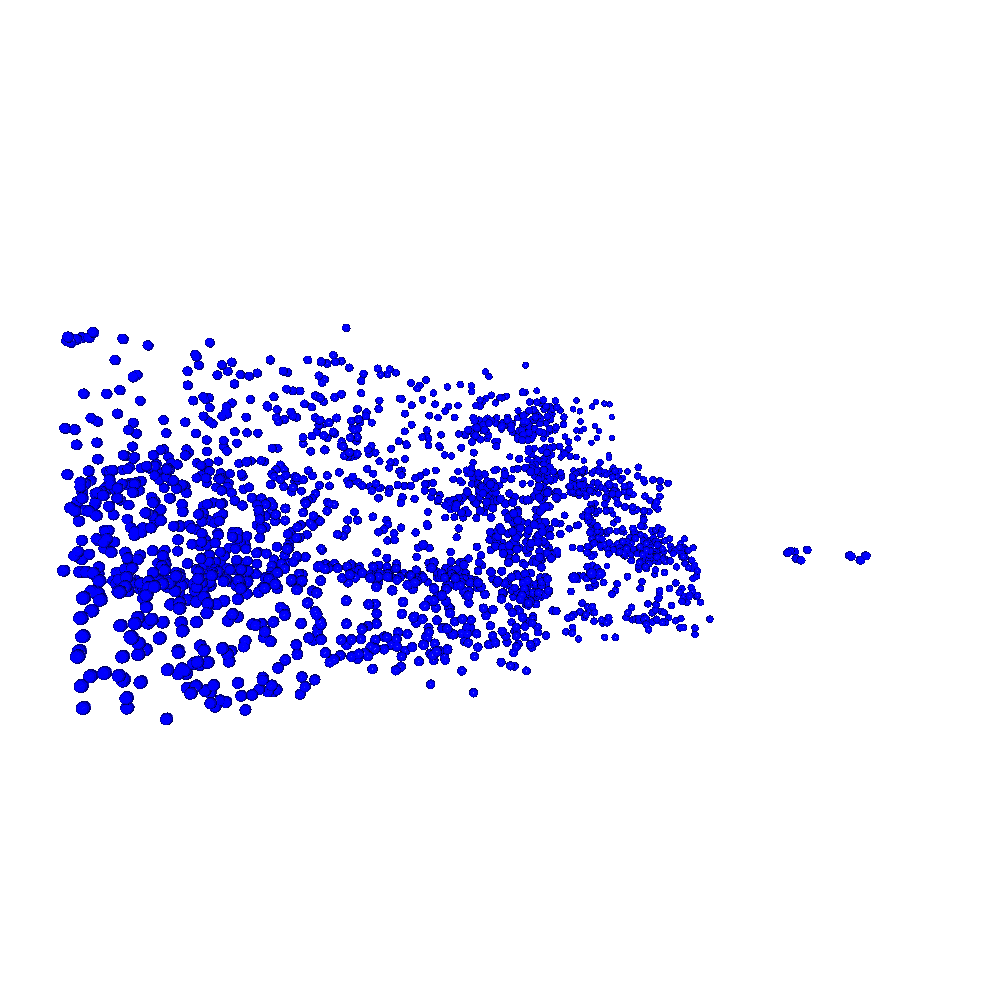} & 
\includegraphics[width=0.12\linewidth]{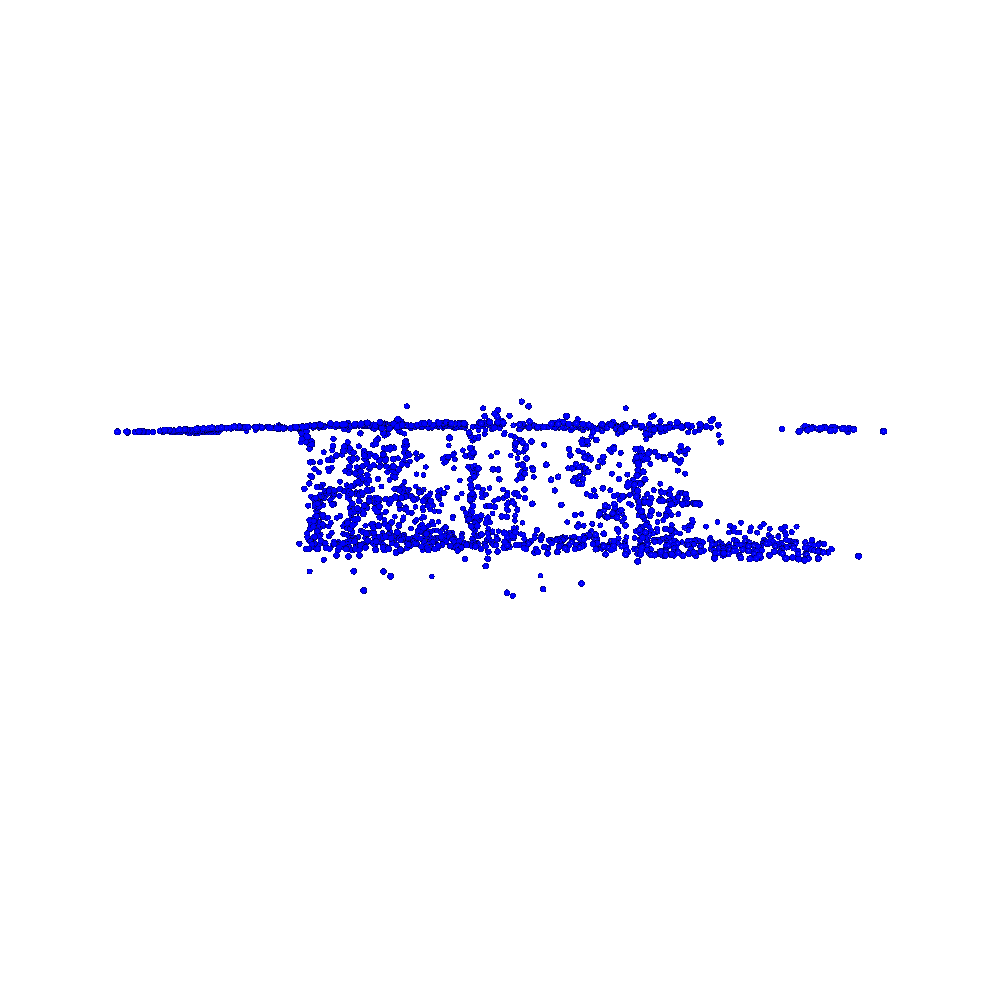}\\

&\textbf{\textcolor{blue}{\scriptsize Table}} &
\textbf{\textcolor{blue}{\scriptsize Shelf}} &
\textbf{\textcolor{blue}{\scriptsize Chair}} &
\textbf{\textcolor{blue}{\scriptsize Cabinet}} &
\textbf{\textcolor{blue}{\scriptsize Bed}} &
\textbf{\textcolor{blue}{\scriptsize Desk}}\\

\end{tabular}

\caption{Visualization of representative original and adversarial ScanObjectNN point clouds generated by Topo-ADV, together with their misclassified labels. The top row represents the clean point clouds, the middle and bottom rows present the adversarial point clouds generated with PointNet and DGCNN as victim models, respectively.}
\label{fig:adv_examplesSONN}

\end{figure}

Refer to Fig.~\ref{fig:adv_examplesSONN} to see examples of adversarial point clouds generated by Topo-ADV for the ScanObjectNN dataset (using PB-T50-RS, the most challenging variant~\cite{uy2019revisiting}) to attack PointNet and DGCNN. 

\begin{figure}
    \centering
\begin{tabular}{c|c|c|c}
    \textbf{Clean} & $\epsilon:0.2$ &  $\epsilon:0.55$ &  $\epsilon:1.0$\\
    \hline

\includegraphics[width=0.2\linewidth]{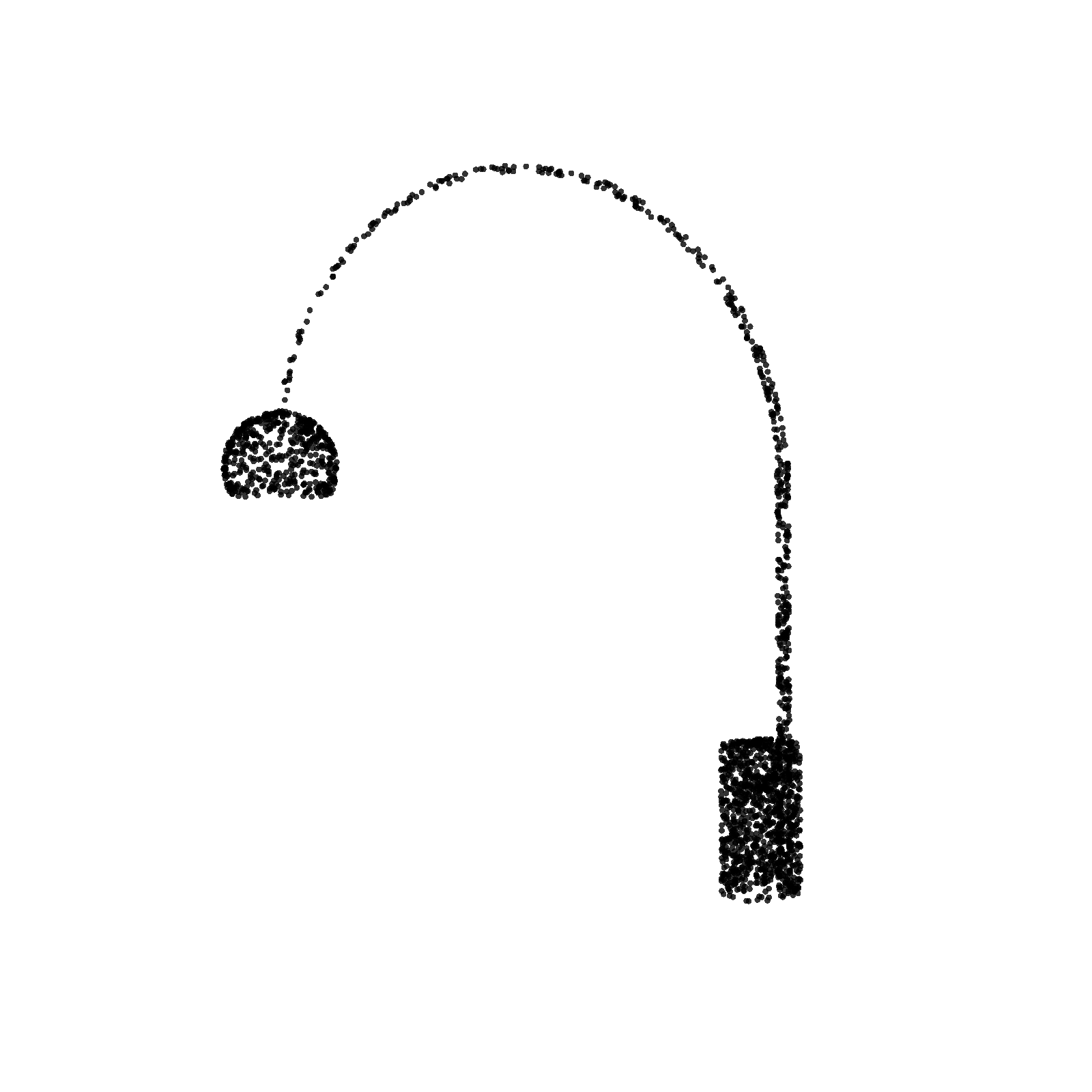} &
\includegraphics[width=0.2\linewidth]{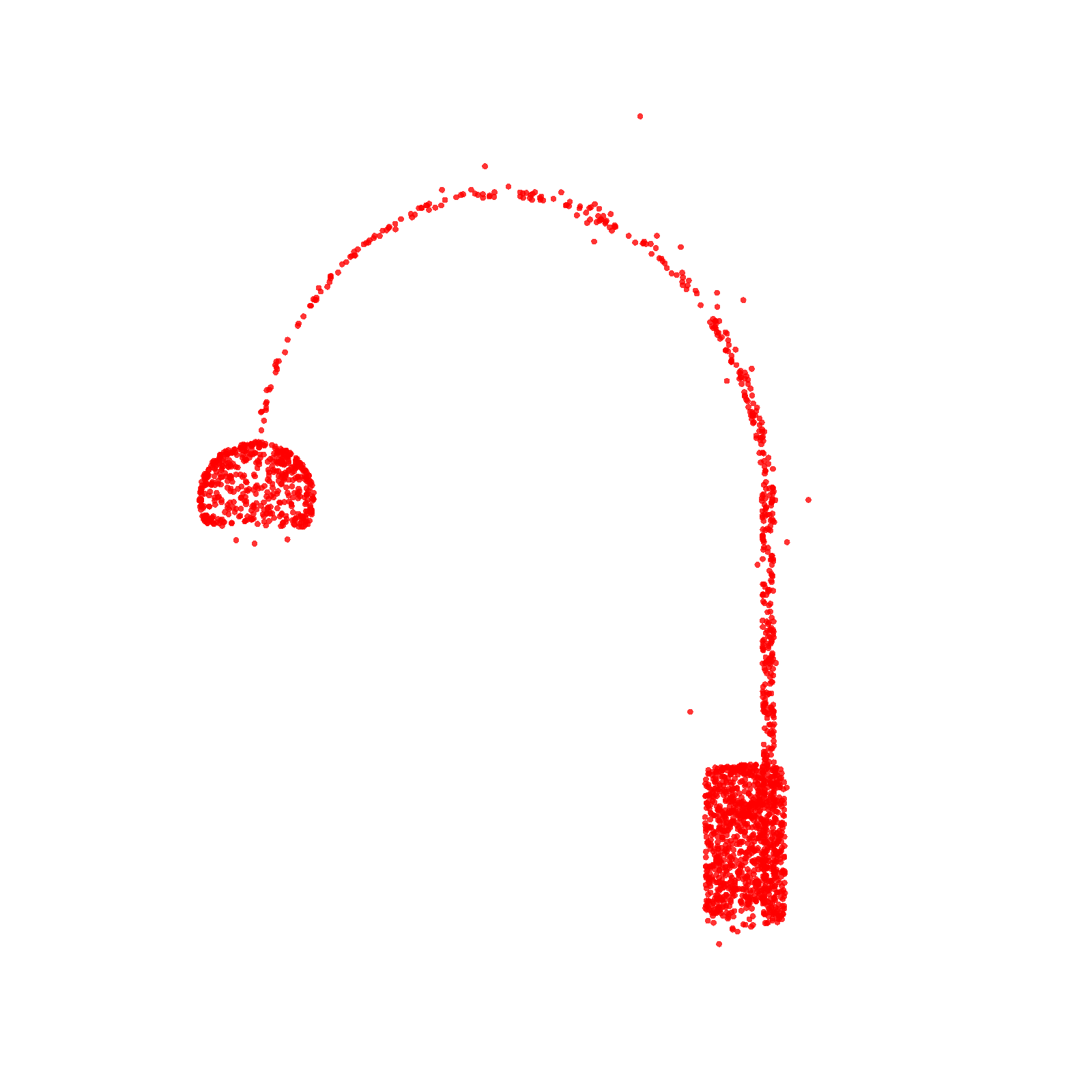} &
\includegraphics[width=0.2\linewidth]{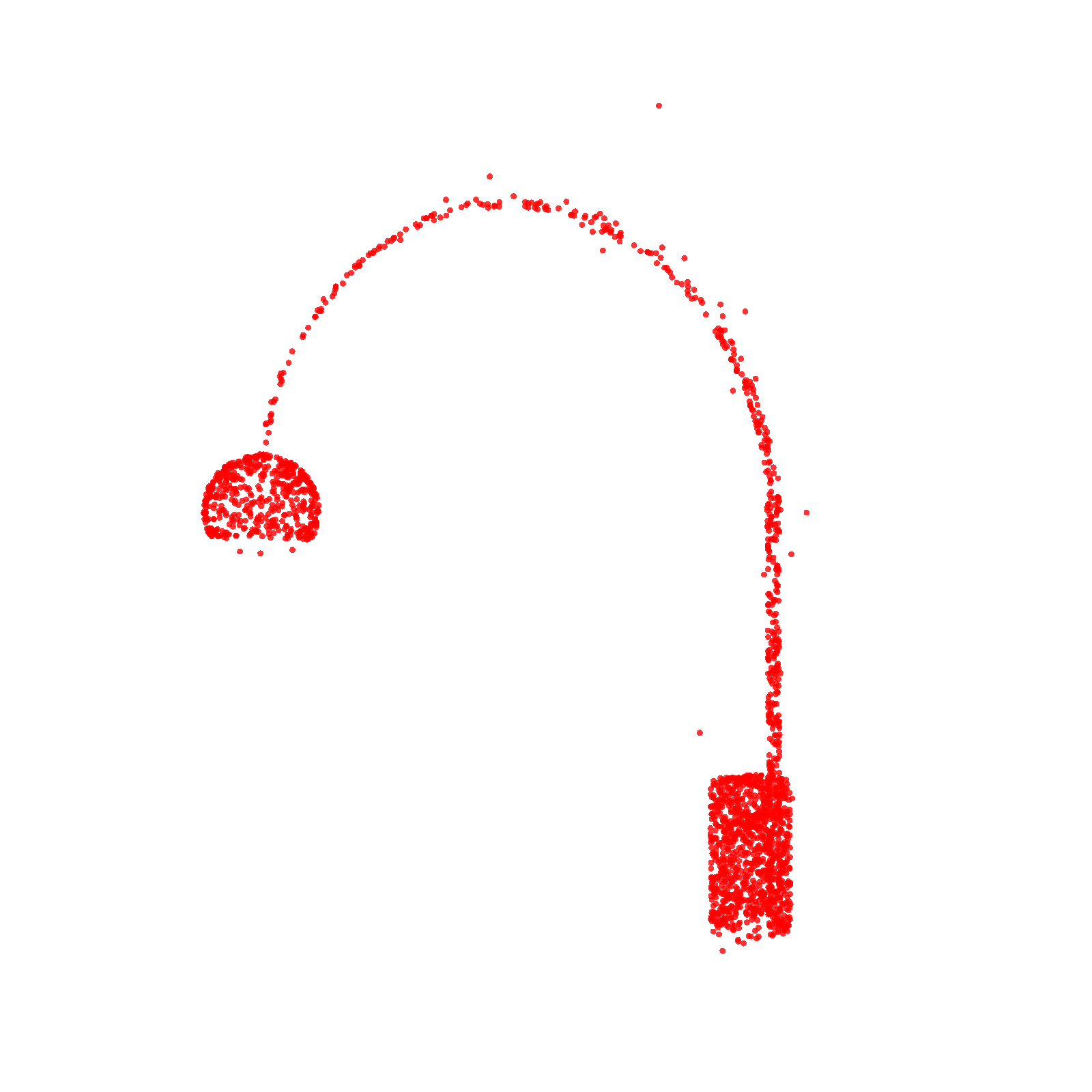} &
\includegraphics[width=0.2\linewidth]{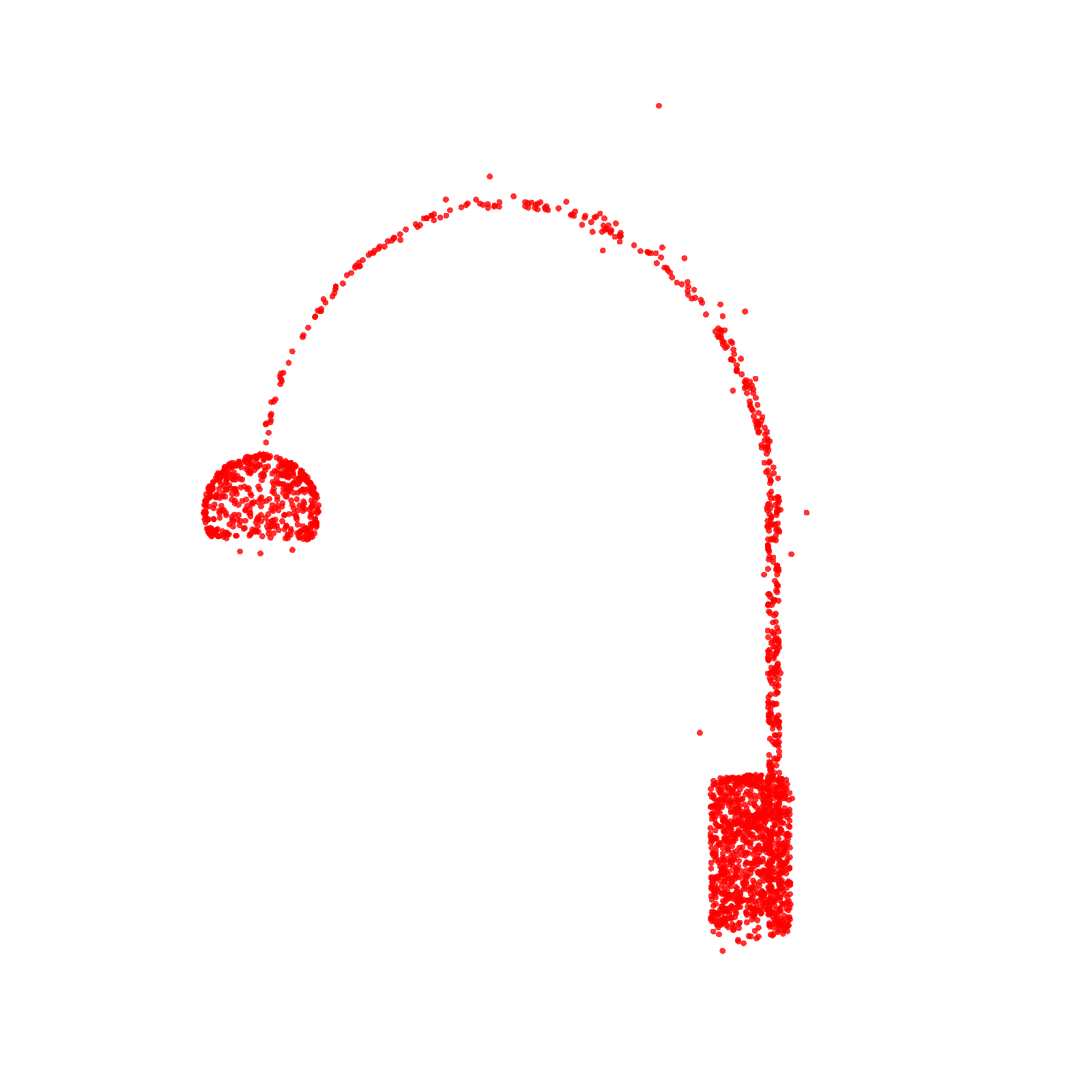} \\
    
\textbf{\scriptsize Lamp}&  \textbf{\textcolor{red}{\scriptsize Vase}} & \textbf{\textcolor{red}{\scriptsize Vase}} & \textbf{\textcolor{red}{\scriptsize Vase}}\\

\includegraphics[width=0.2\linewidth]{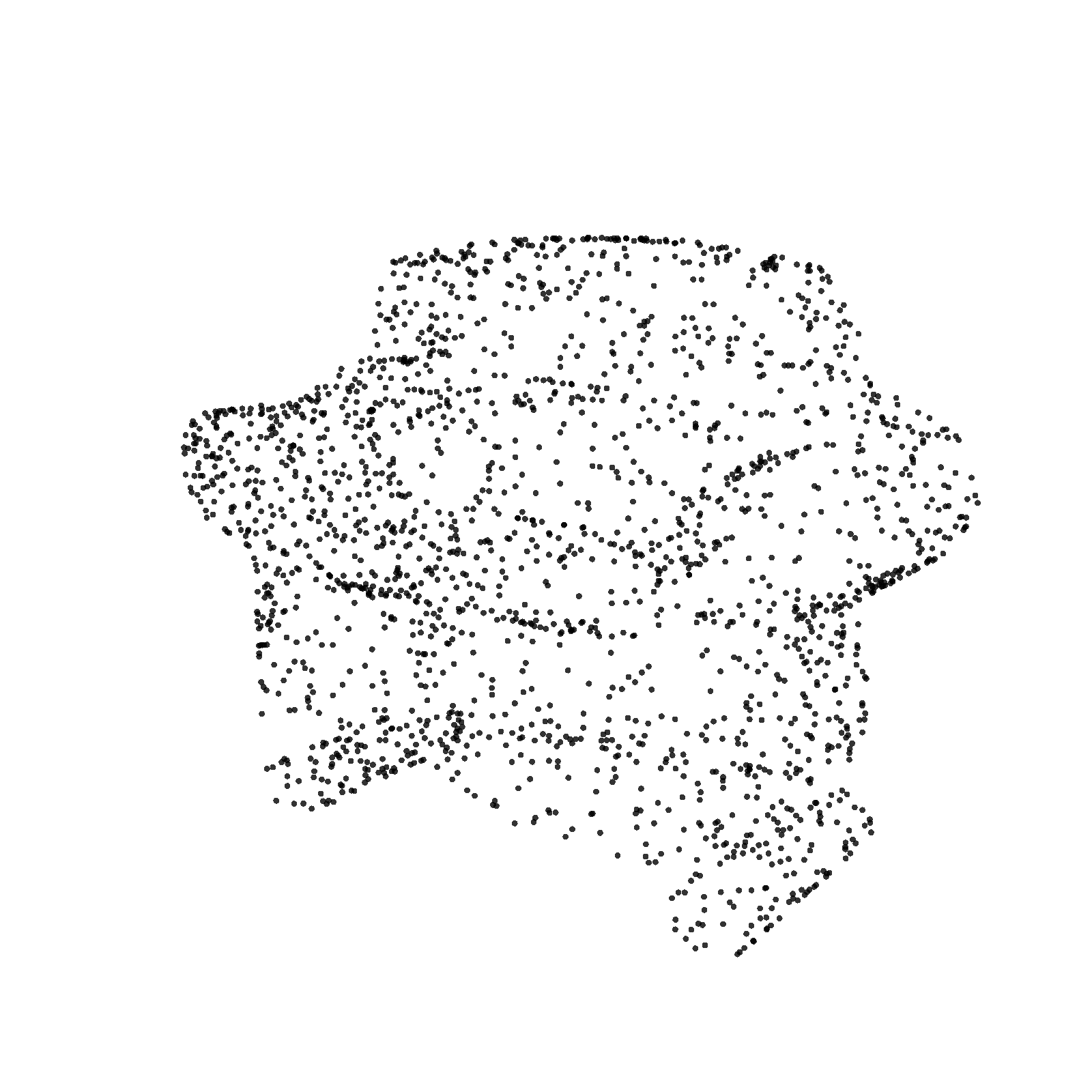} &
\includegraphics[width=0.2\linewidth]{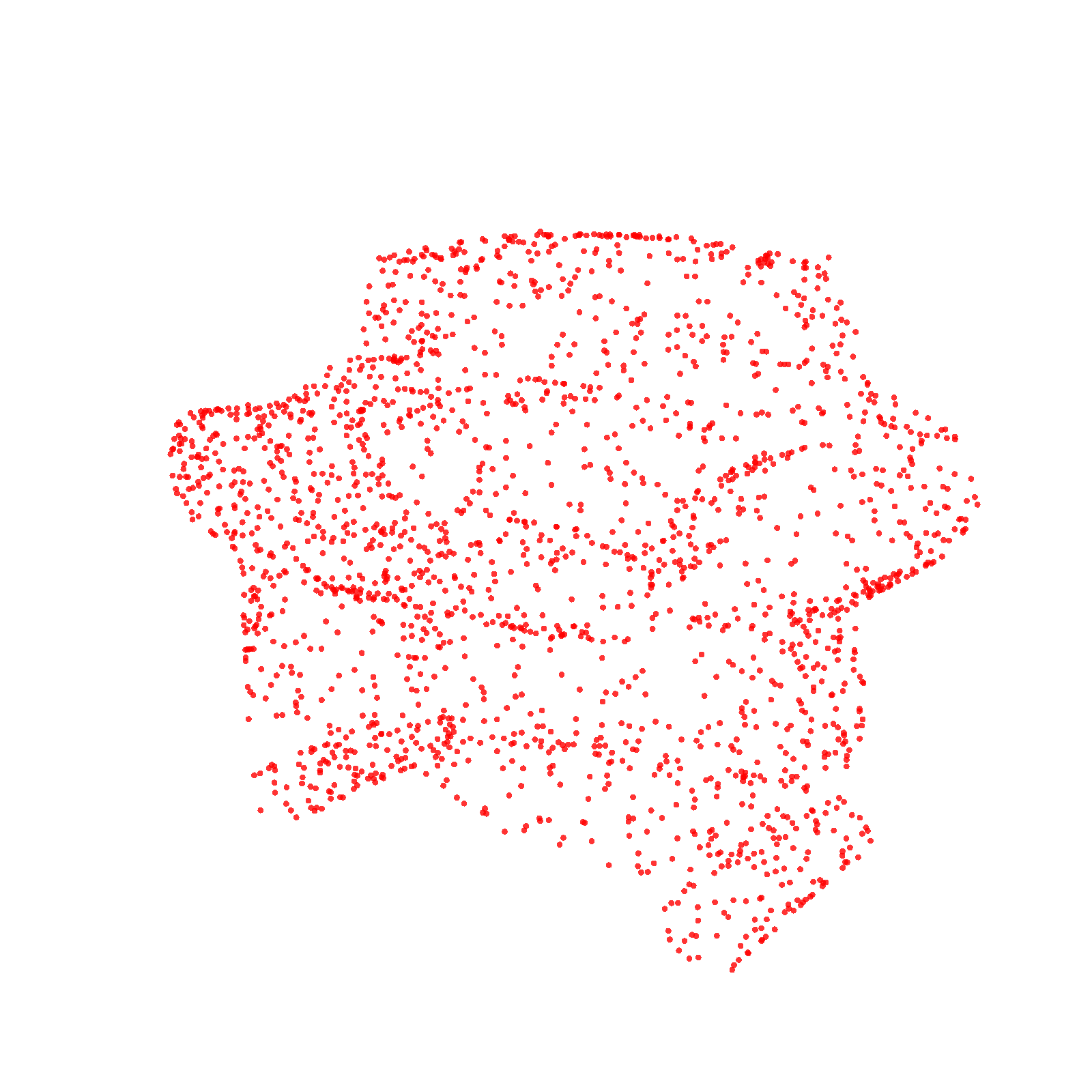} &
\includegraphics[width=0.2\linewidth]{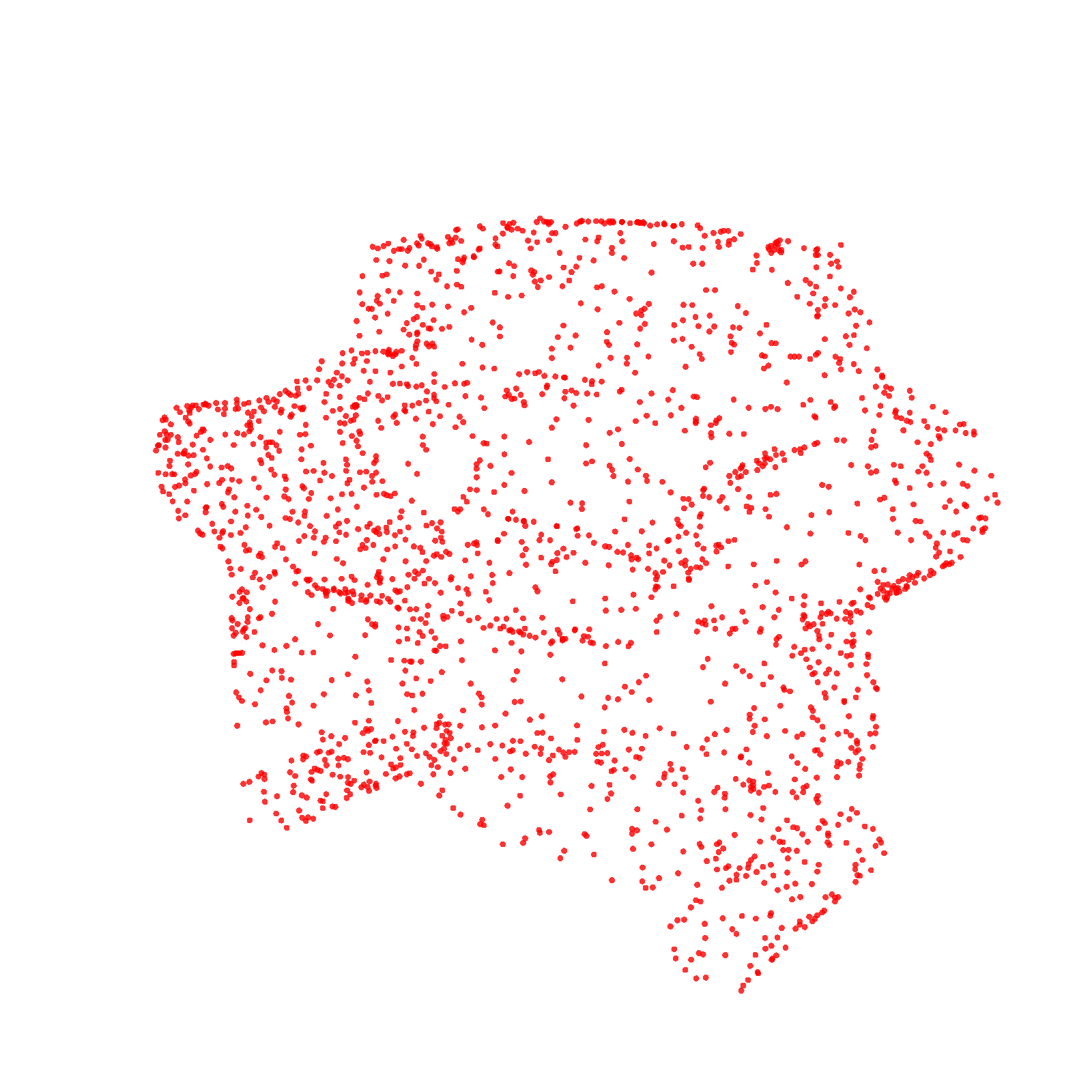} &
\includegraphics[width=0.2\linewidth]{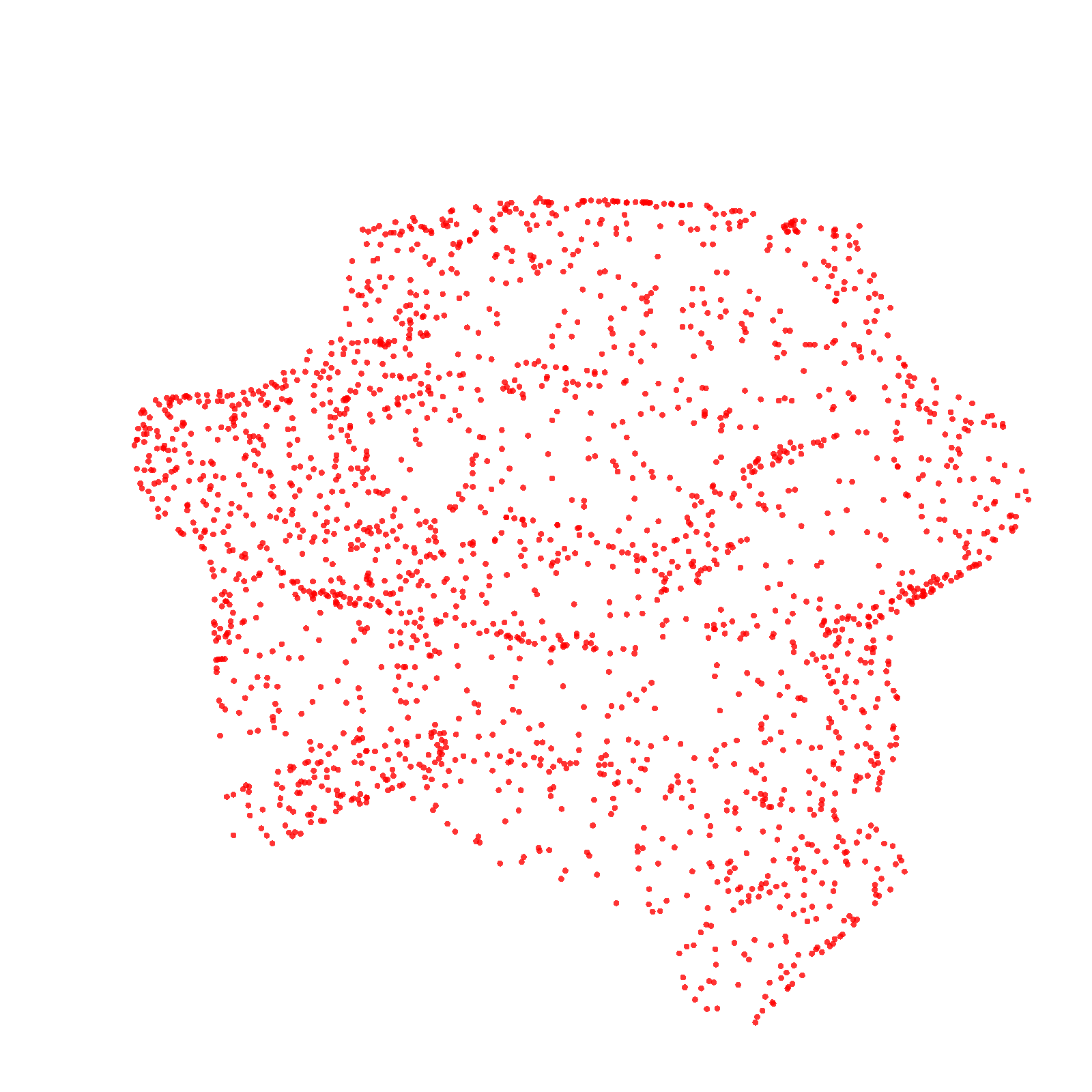} \\
    
\textbf{\scriptsize Sofa}&  \textbf{\textcolor{red}{\scriptsize Dresser}} & \textbf{\textcolor{red}{\scriptsize Dresser}} & \textbf{\textcolor{red}{\scriptsize Dresser}}\\

\includegraphics[width=0.2\linewidth]{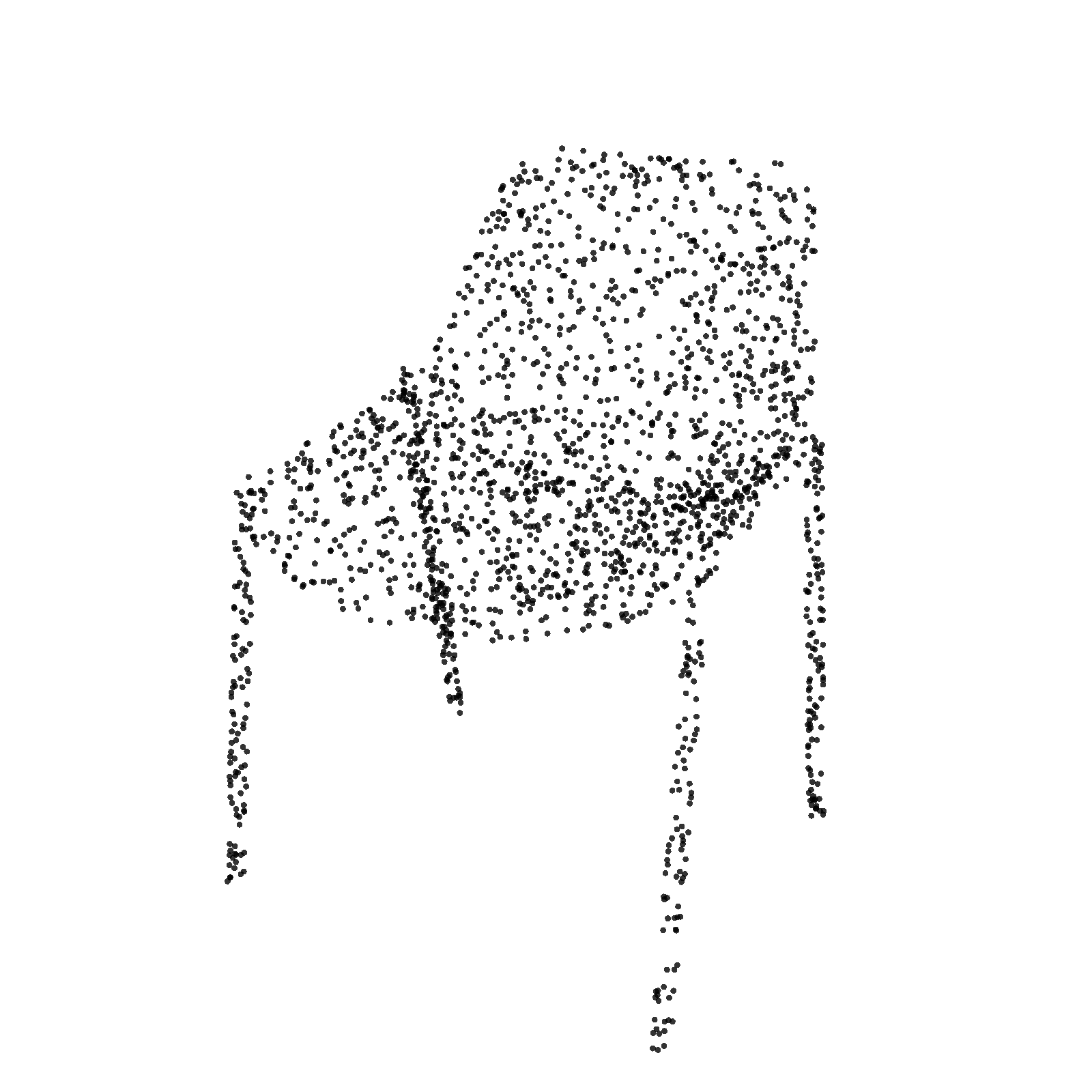} &
\includegraphics[width=0.2\linewidth]{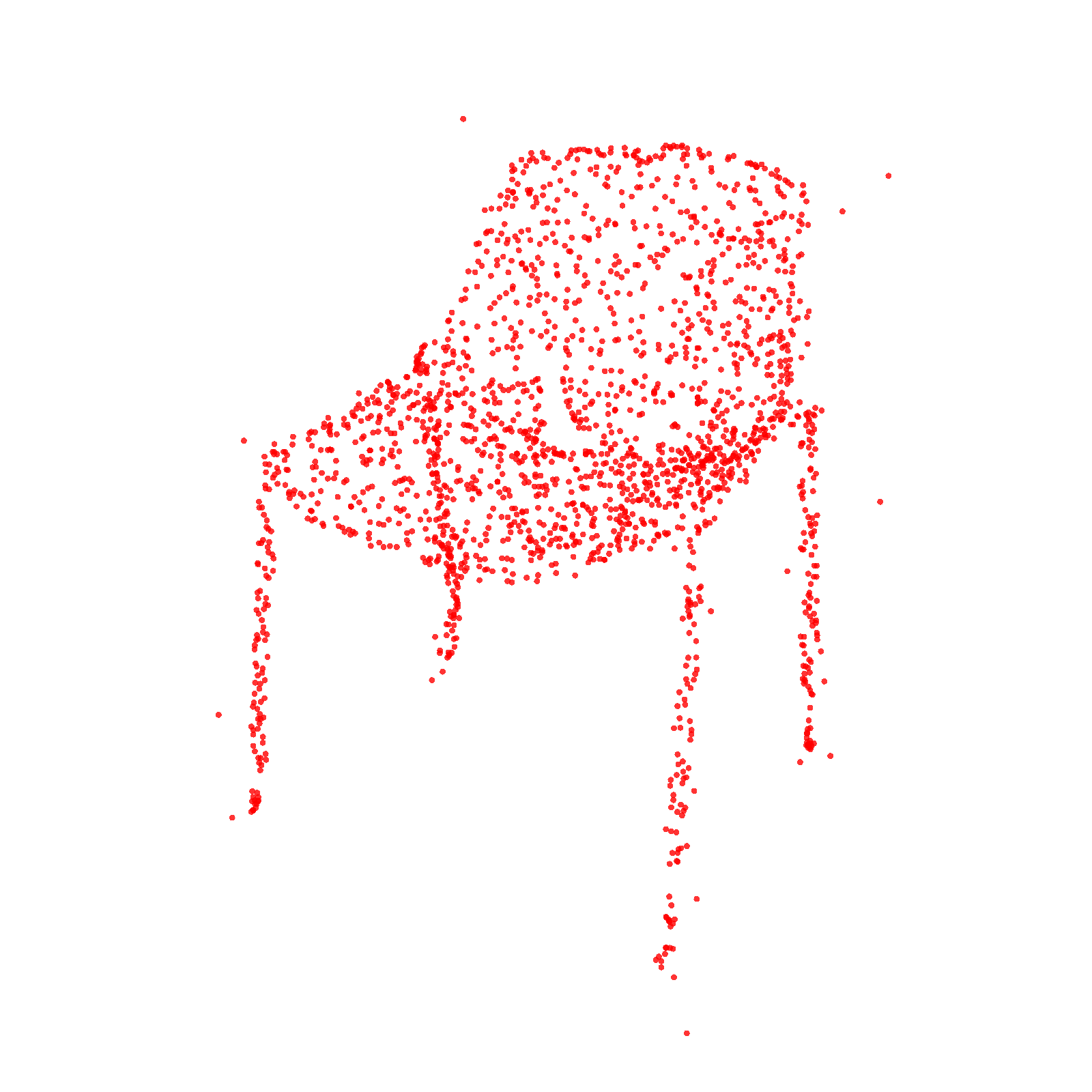} &
\includegraphics[width=0.2\linewidth]{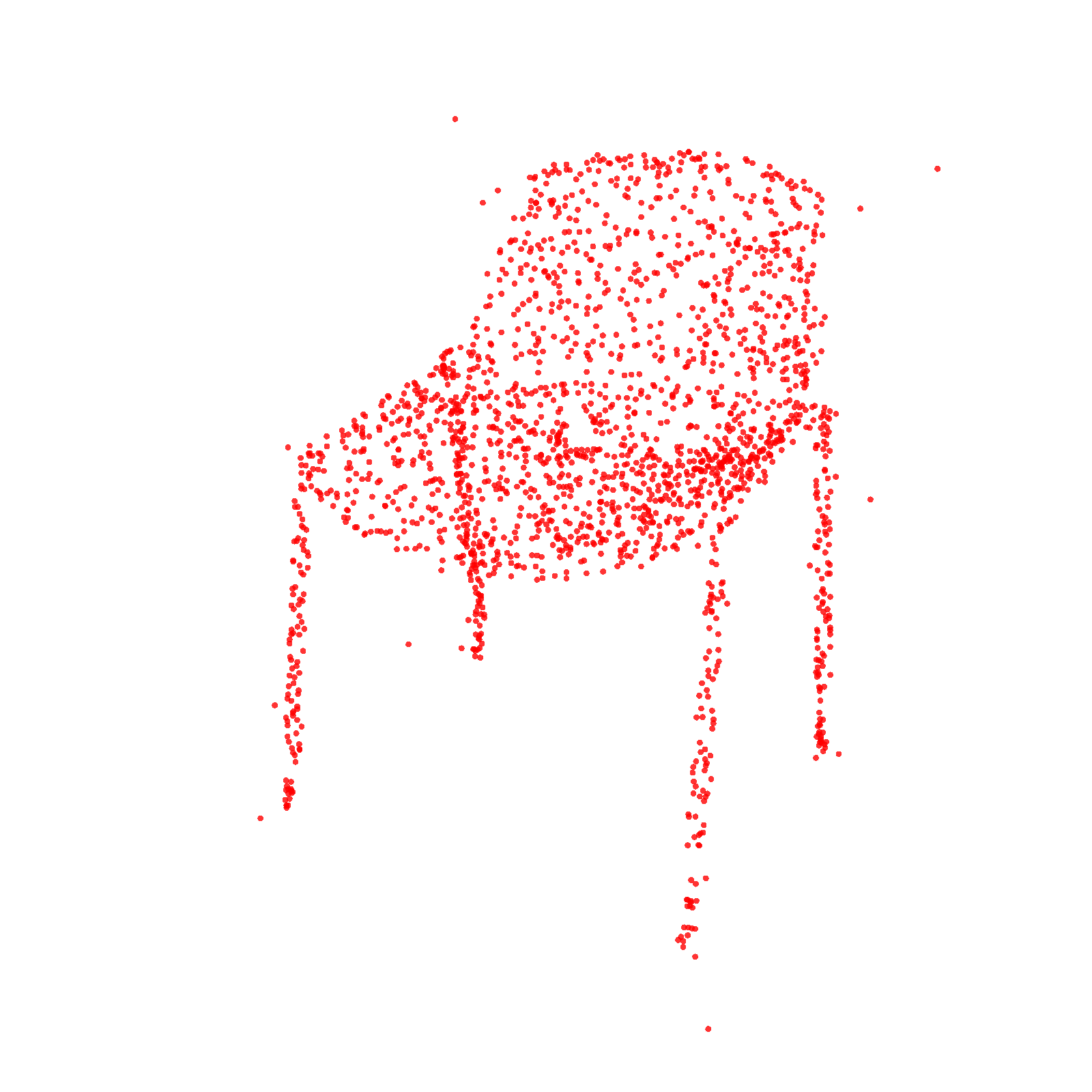} &
\includegraphics[width=0.2\linewidth]{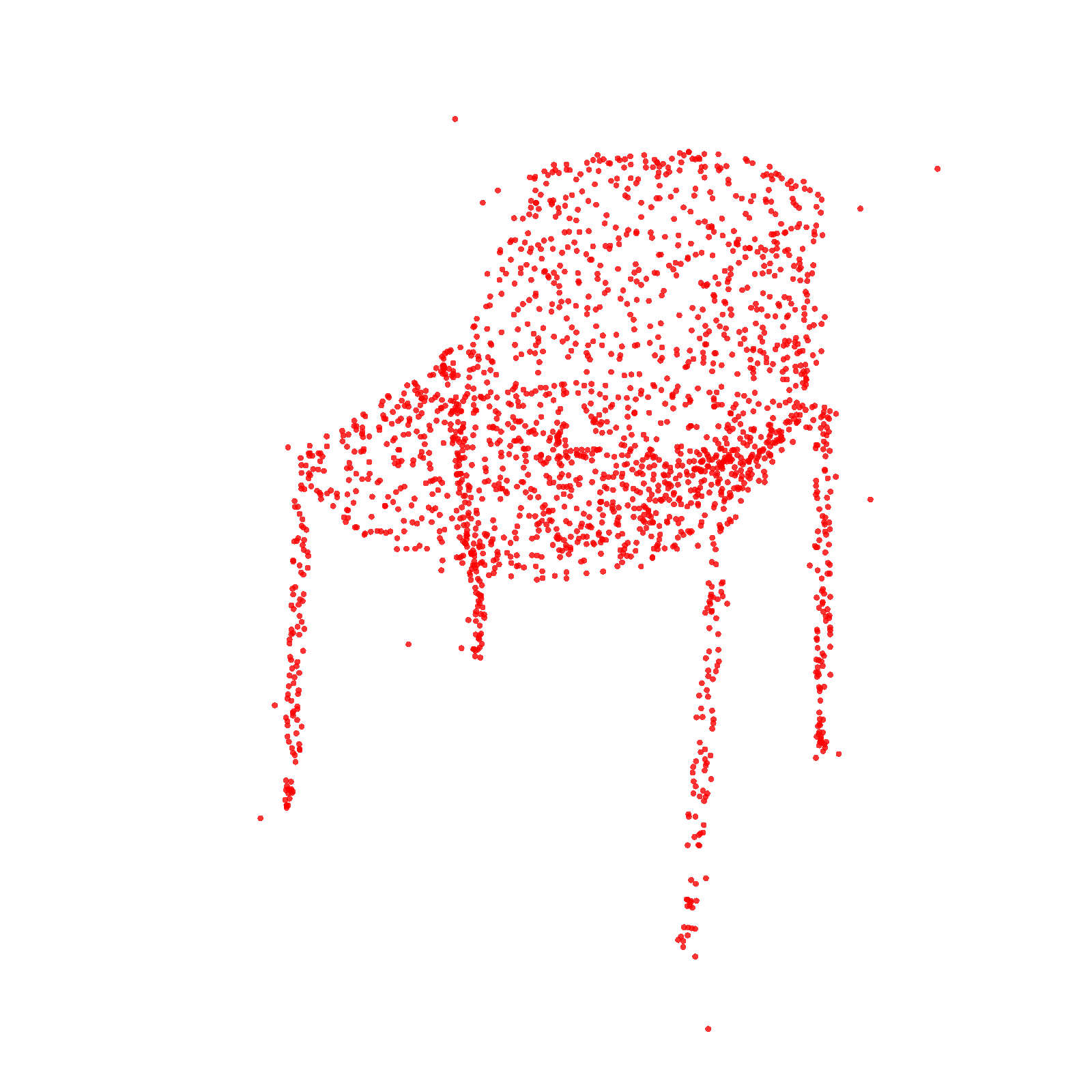} \\
    
\textbf{\scriptsize Chair}&  \textbf{\textcolor{red}{\scriptsize Bench}} & \textbf{\textcolor{red}{\scriptsize Bench}} & \textbf{\textcolor{red}{\scriptsize Bench}}\\

\includegraphics[width=0.2\linewidth]{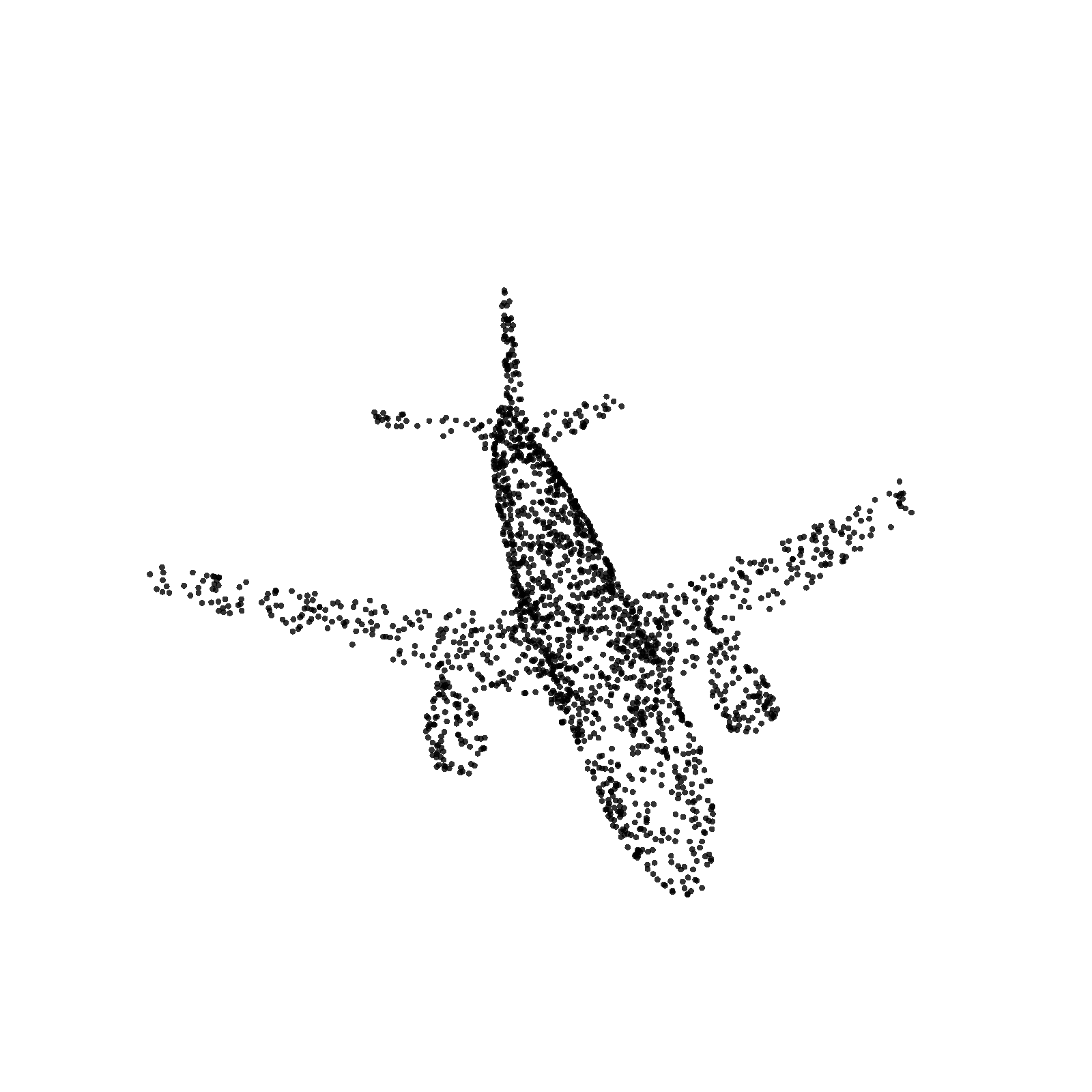} &
\includegraphics[width=0.2\linewidth]{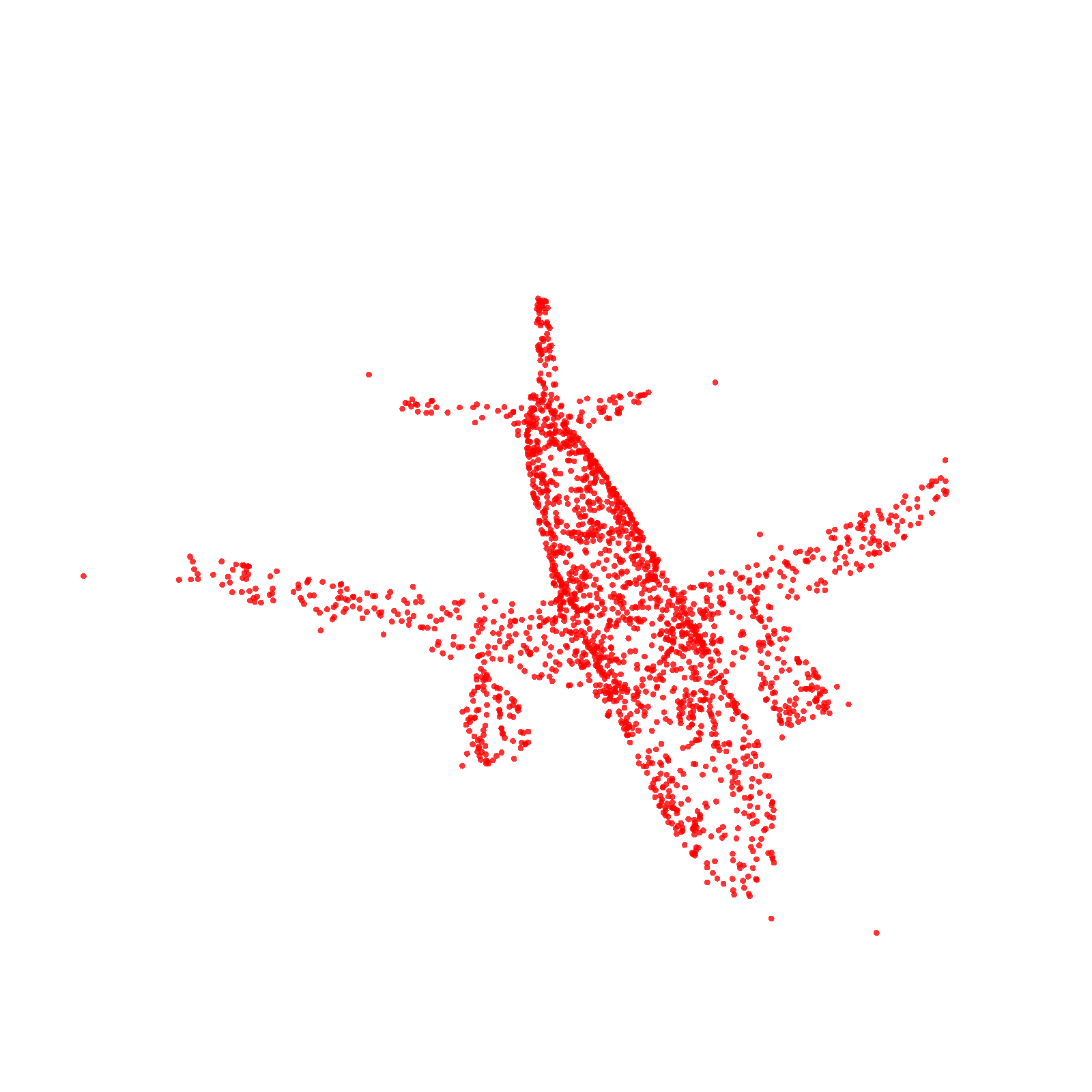} &
\includegraphics[width=0.2\linewidth]{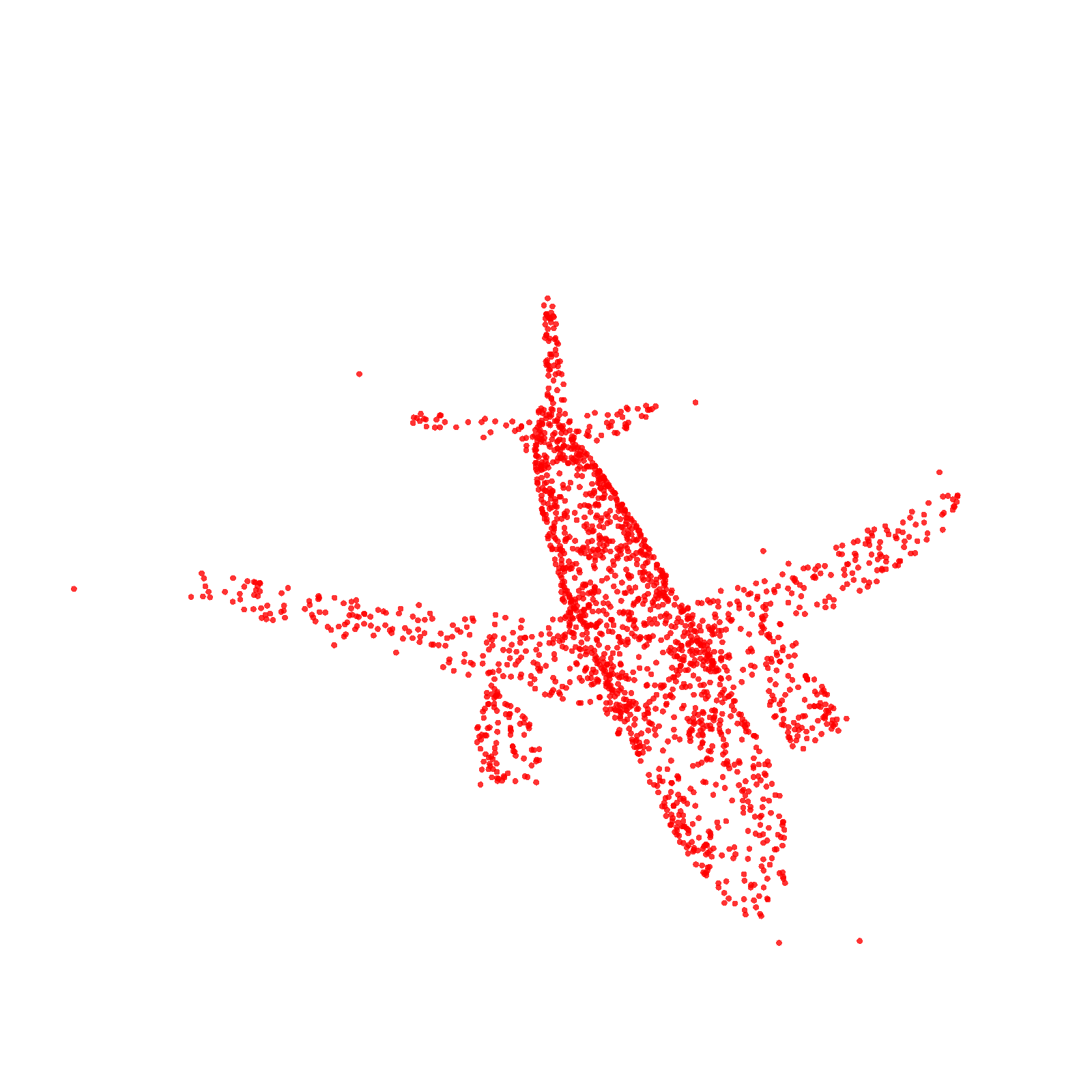} &
\includegraphics[width=0.2\linewidth]{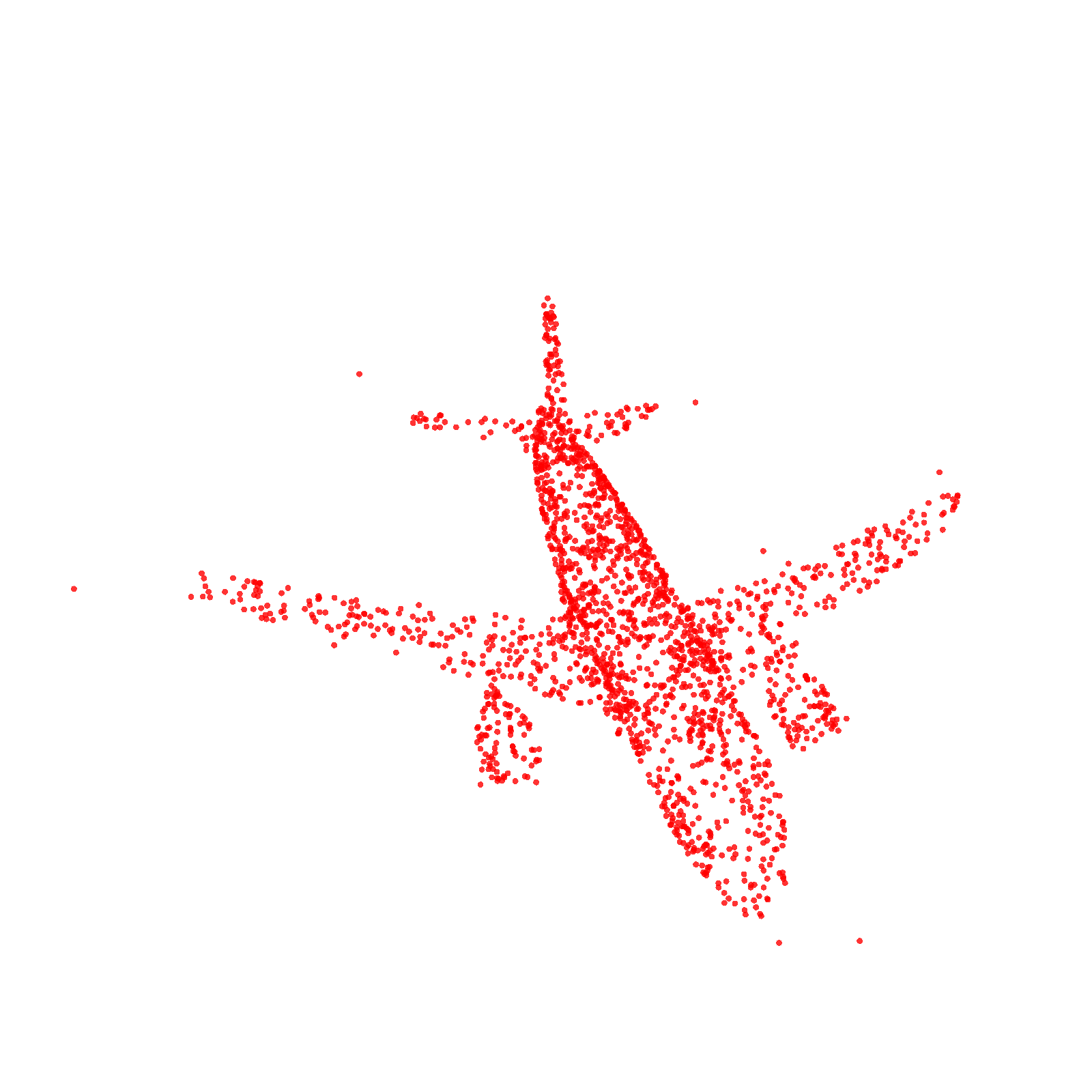} \\
    
\textbf{\scriptsize Airplane}&  \textbf{\textcolor{red}{\scriptsize Keyboard}} & \textbf{\textcolor{red}{\scriptsize Keyboard}} & \textbf{\textcolor{red}{\scriptsize Keyboard}}\\

\includegraphics[width=0.2\linewidth]{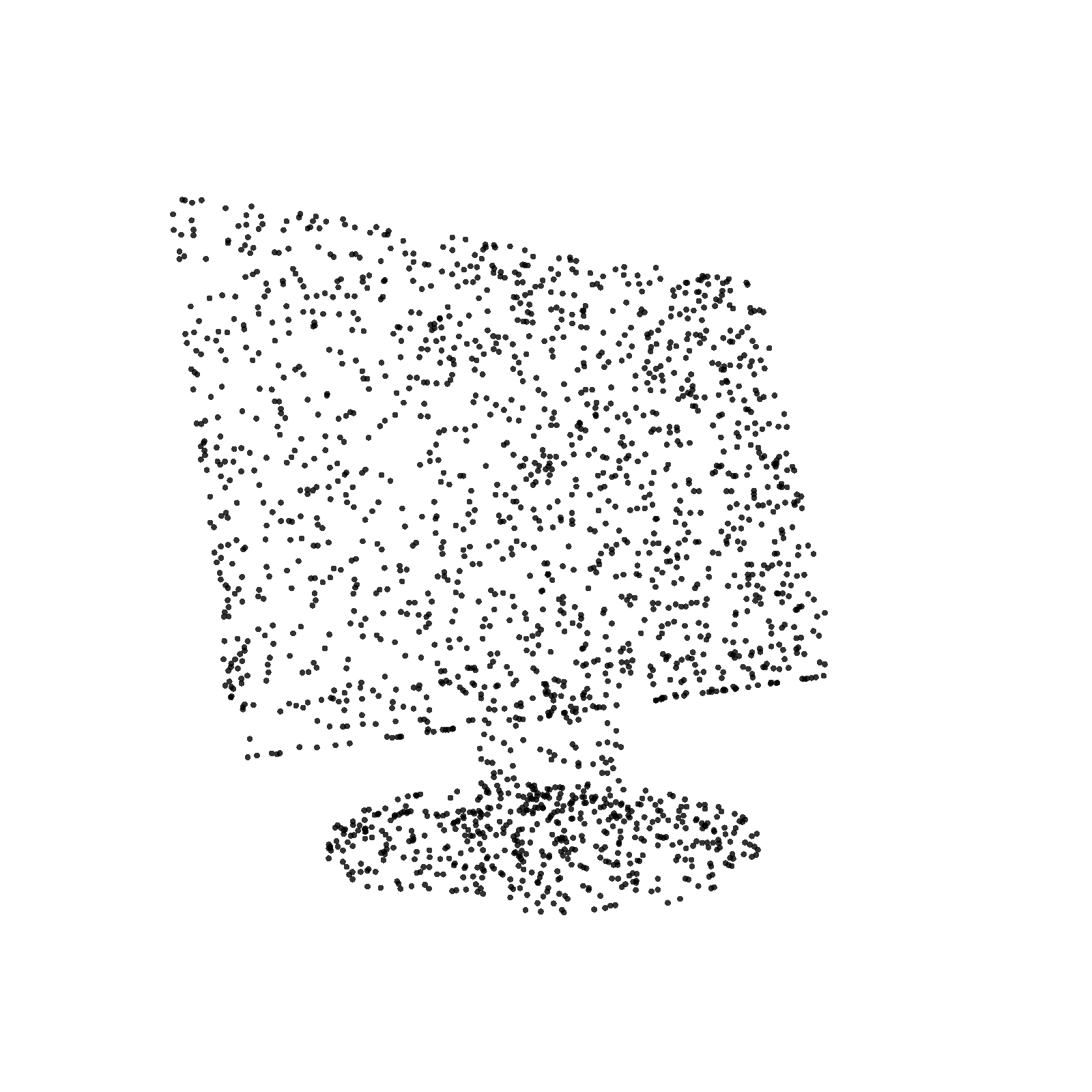} &
\includegraphics[width=0.2\linewidth]{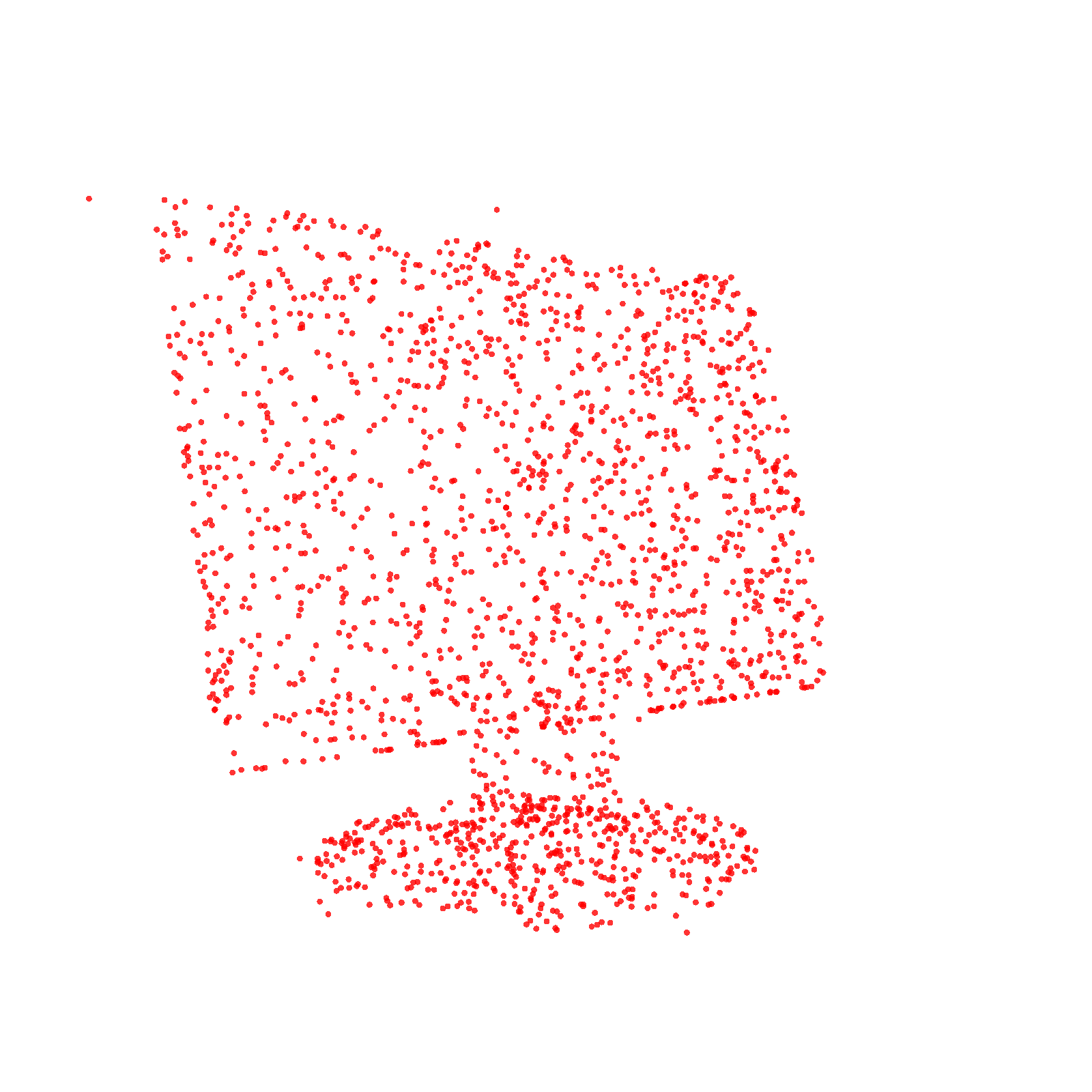} &
\includegraphics[width=0.2\linewidth]{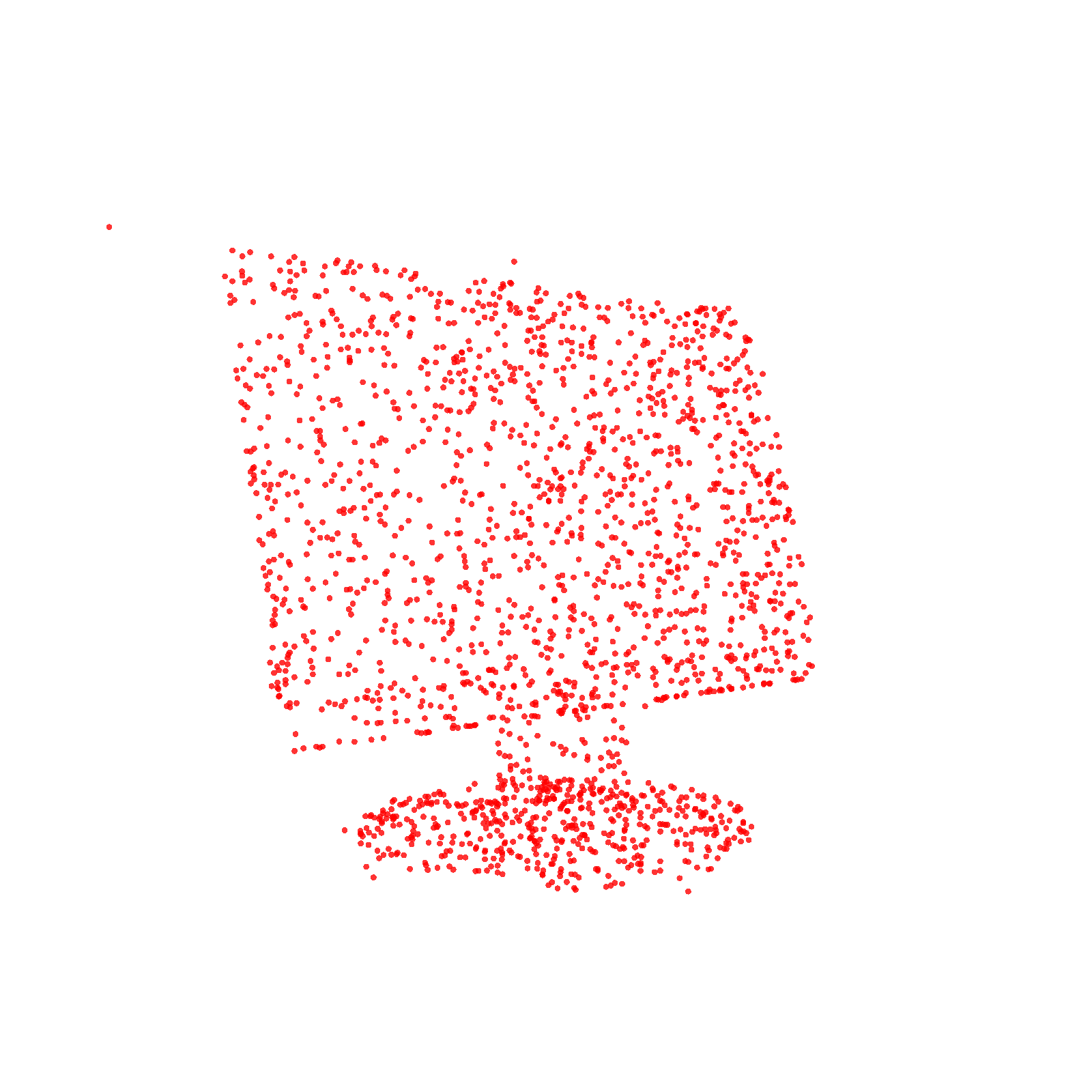} &
\includegraphics[width=0.2\linewidth]{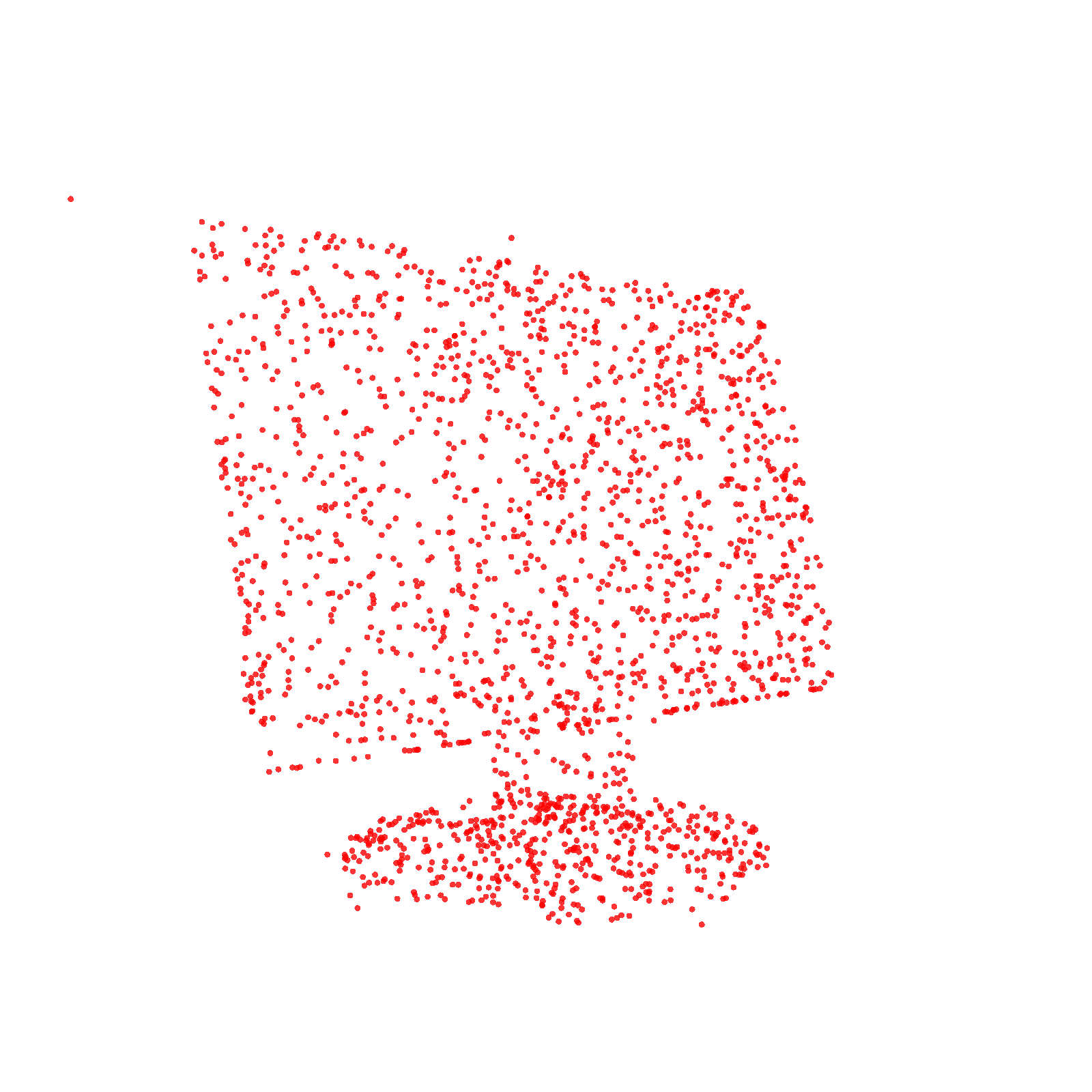} \\
    
\textbf{\scriptsize Monitor}&  \textbf{\textcolor{red}{\scriptsize Stairs}} & \textbf{\textcolor{red}{\scriptsize Stairs}} & \textbf{\textcolor{red}{\scriptsize Stairs}}\\
    
\end{tabular}
    
    \caption{Visualization of representative original and adversarial ModelNet40 point clouds generated by Topo-ADV, together with their misclassified labels, for varied values of the budget $\epsilon$. The victim model is PointNet.
    }
    \label{fig:adv_examplesMN40_PN}
\end{figure}

Adversarial examples on the ModelNet40 dataset, using PointNet as the victim, are presented in Fig.~\ref{fig:adv_examplesMN40_PN}. We note that with an increase in the budget $\epsilon$, the increase in perturbation is barely present. 
Consequently, the misclassified labels remained unchanged as $\epsilon$ was increased. 

\end{document}